\documentclass[journal]{IEEEtran}

\usepackage[utf8]{inputenc}

\usepackage{amsmath, amssymb, amsthm} 
\usepackage{bm} 
\usepackage{mathtools} 
\usepackage{stmaryrd} 

\usepackage{siunitx}

\usepackage{hyperref}
\usepackage{cleveref}

\usepackage{algorithm}
\usepackage{algpseudocode}

\usepackage{makecell}

\usepackage{subfiles}

\usepackage{subcaption}

\usepackage{graphicx}
\graphicspath{{images/}}

\usepackage{ifthen}
\usepackage{etoolbox} 

\usepackage{tikz, tikz-3dplot}
\usetikzlibrary{arrows.meta, calc, 3d}
\usepackage{pgfplots}

\usepackage{xcolor}

\usepackage[style=ieee]{biblatex}
\makeatletter
\tikzoption{canvas is plane}[]{\setOxy#1}
\def\setOxy O(#1)x(#2)y(#3)%
{\def\tikz@plane@origin{#1}%
  \def\tikz@plane@x{#2}%
    \def\tikz@plane@y{#3}%
    \tikz@canvas@is@plane
}
\makeatother

\newcommand\dodeca[1][]{%
  \coordinate (o) at ({0.0000}, {0.0000}, {0.0000});
  \coordinate (v_1) at ({-0.1618}, {-0.1648}, {-0.0816});
  \coordinate (v_2) at ({-0.0618}, {-0.1512}, {0.1826});
  \coordinate (v_3) at ({-0.2000}, {0.1155}, {-0.0816});
  \coordinate (v_4) at ({-0.1000}, {0.1291}, {0.1826});
  \coordinate (v_5) at ({0.1000}, {-0.1291}, {-0.1826});
  \coordinate (v_6) at ({0.2000}, {-0.1155}, {0.0816});
  \coordinate (v_7) at ({0.0618}, {0.1512}, {-0.1826});
  \coordinate (v_8) at ({0.1618}, {0.1648}, {0.0816});
  \coordinate (v_9) at ({0.0618}, {-0.2225}, {0.0816});
  \coordinate (v_10) at ({0.0000}, {-0.2309}, {-0.0816});
  \coordinate (v_11) at ({0.0000}, {0.2309}, {0.0816});
  \coordinate (v_12) at ({-0.0618}, {0.2225}, {-0.0816});
  \coordinate (v_13) at ({0.0000}, {0.0000}, {-0.2449});
  \coordinate (v_14) at ({0.1618}, {0.0221}, {0.1826});
  \coordinate (v_15) at ({-0.1618}, {-0.0221}, {-0.1826});
  \coordinate (v_16) at ({0.0000}, {0.0000}, {0.2449});
  \coordinate (v_17) at ({-0.2236}, {0.0577}, {0.0816});
  \coordinate (v_18) at ({-0.2000}, {-0.1155}, {0.0816});
  \coordinate (v_19) at ({0.2000}, {0.1155}, {-0.0816});
  \coordinate (v_20) at ({0.2236}, {-0.0577}, {-0.0816});

  \draw (v_1) -- (v_10);
  \draw (v_1) -- (v_15);
  \draw (v_1) -- (v_18);
  \draw (v_2) -- (v_9);
  \draw (v_2) -- (v_16);
  \draw (v_2) -- (v_18);
  \draw (v_3) -- (v_12);
  \draw (v_3) -- (v_17);
  \draw (v_3) -- (v_15);
  \draw (v_4) -- (v_11);
  \draw (v_4) -- (v_16);
  \draw (v_4) -- (v_17);
  \draw (v_5) -- (v_10);
  \draw (v_5) -- (v_13);
  \draw (v_5) -- (v_20);
  \draw (v_6) -- (v_9);
  \draw (v_6) -- (v_14);
  \draw (v_6) -- (v_20);
  \draw (v_7) -- (v_12);
  \draw (v_7) -- (v_13);
  \draw (v_7) -- (v_19);
  \draw (v_8) -- (v_11);
  \draw (v_8) -- (v_19);
  \draw (v_8) -- (v_14);
  \draw (v_9) -- (v_10);
  \draw (v_11) -- (v_12);
  \draw (v_13) -- (v_15);
  \draw (v_14) -- (v_16);
  \draw (v_17) -- (v_18);
  \draw (v_19) -- (v_20);
  \ifthenelse{\not\equal{#1}{}}{
    \draw[#1] (o) circle (0.1978);
  }{}
}

\newcommand\dodecaConnections{
     \coordinate (u_1) at ({0.2000}, {-0.1155}, {0.3266});
     \coordinate (u_2) at ({0.0000}, {0.2309}, {0.3266});
     \coordinate (u_3) at ({-0.2000}, {-0.1155}, {0.3266});
     \coordinate (u_4) at ({0.2000}, {-0.3464}, {0.0000});
     \coordinate (u_5) at ({0.2000}, {0.3464}, {0.0000});
     \coordinate (u_6) at ({-0.4000}, {0.0000}, {0.0000});
     \coordinate (u_7) at ({0.4000}, {0.0000}, {0.0000});
     \coordinate (u_8) at ({-0.2000}, {0.3464}, {0.0000});
     \coordinate (u_9) at ({-0.2000}, {-0.3464}, {0.0000});
     \coordinate (u_10) at ({-0.2000}, {0.1155}, {-0.3266});
     \coordinate (u_11) at ({0.0000}, {-0.2309}, {-0.3266});
     \coordinate (u_12) at ({0.2000}, {0.1155}, {-0.3266});
     
     \draw[fill, tdplot_screen_coords] (o) circle (0.002);
     
     \draw[fill, orange, tdplot_screen_coords] (v_3) circle (0.008);
     \draw[fill, orange, tdplot_screen_coords] (v_6) circle (0.008);
     \draw[fill, orange, tdplot_screen_coords] (v_10) circle (0.008);
     \draw[fill, orange, tdplot_screen_coords] (v_11) circle (0.008);
     \draw[fill, orange, tdplot_screen_coords] (v_13) circle (0.008);
     \draw[fill, orange, tdplot_screen_coords] (v_16) circle (0.008);
     \draw[fill, orange, tdplot_screen_coords] (v_18) circle (0.008);
     \draw[fill, orange, tdplot_screen_coords] (v_19) circle (0.008);
     
     \draw[dashed] (v_6) -- (v_16);
     \draw[dashed] (v_11) -- (v_16);
     \draw[dashed] (v_16) -- (v_18);
     \draw[dashed] (v_6) -- (v_10);
     \draw[dashed] (v_11) -- (v_19);
     \draw[dashed] (v_3) -- (v_18);
     \draw[dashed] (v_6) -- (v_19);
     \draw[dashed] (v_3) -- (v_11);
     \draw[dashed] (v_10) -- (v_18);
     \draw[dashed] (v_3) -- (v_13);
     \draw[dashed] (v_10) -- (v_13);
     \draw[dashed] (v_13) -- (v_19);
     
     \draw[dotted, -stealth, shorten >=2pt] (o) -- (u_1);
     \draw[dotted, -stealth, shorten >=2pt] (o) -- (u_2);
     \draw[dotted, -stealth, shorten >=2pt] (o) -- (u_3);
     \draw[dotted, -stealth, shorten >=2pt] (o) -- (u_4);
     \draw[dotted, -stealth, shorten >=2pt] (o) -- (u_5);
     \draw[dotted, -stealth, shorten >=2pt] (o) -- (u_6);
     \draw[dotted, -stealth, shorten >=2pt] (o) -- (u_7);
     \draw[dotted, -stealth, shorten >=2pt] (o) -- (u_8);
     \draw[dotted, -stealth, shorten >=2pt] (o) -- (u_9);
     \draw[dotted, -stealth, shorten >=2pt] (o) -- (u_10);
     \draw[dotted, -stealth, shorten >=2pt] (o) -- (u_11);
     \draw[dotted, -stealth, shorten >=2pt] (o) -- (u_12);
     }

\newcommand\dodecaOrientations[1][]{%
     \ifthenelse{\not\equal{#1}{}}
     {
      \draw[#1] (o) -- (v_16);
     }{
      \draw[dashed, -stealth, shorten <=2pt, shorten >=2pt] (o) -- (v_1);
      \draw[dashed, -stealth, shorten <=2pt, shorten >=2pt] (o) -- (v_2);
      \draw[dashed, -stealth, shorten <=2pt, shorten >=2pt] (o) -- (v_3);
      \draw[dashed, -stealth, shorten <=2pt, shorten >=2pt] (o) -- (v_4);
      \draw[dashed, -stealth, shorten <=2pt, shorten >=2pt] (o) -- (v_5);
      \draw[dashed, -stealth, shorten <=2pt, shorten >=2pt] (o) -- (v_6);
      \draw[dashed, -stealth, shorten <=2pt, shorten >=2pt] (o) -- (v_7);
      \draw[dashed, -stealth, shorten <=2pt, shorten >=2pt] (o) -- (v_8);
      \draw[dashed, -stealth, shorten <=2pt, shorten >=2pt] (o) -- (v_9);
      \draw[dashed, -stealth, shorten <=2pt, shorten >=2pt] (o) -- (v_10);
      \draw[dashed, -stealth, shorten <=2pt, shorten >=2pt] (o) -- (v_11);
      \draw[dashed, -stealth, shorten <=2pt, shorten >=2pt] (o) -- (v_12);
      \draw[dashed, -stealth, shorten <=2pt, shorten >=2pt] (o) -- (v_13);
      \draw[dashed, -stealth, shorten <=2pt, shorten >=2pt] (o) -- (v_14);
      \draw[dashed, -stealth, shorten <=2pt, shorten >=2pt] (o) -- (v_15);
      \draw[dashed, -stealth, shorten <=2pt, shorten >=2pt] (o) -- (v_16);
      \draw[dashed, -stealth, shorten <=2pt, shorten >=2pt] (o) -- (v_17);
      \draw[dashed, -stealth, shorten <=2pt, shorten >=2pt] (o) -- (v_18);
      \draw[dashed, -stealth, shorten <=2pt, shorten >=2pt] (o) -- (v_19);
      \draw[dashed, -stealth, shorten <=2pt, shorten >=2pt] (o) -- (v_20);
  }
}

\newcommand\quadconf{%
  \begin{scope}[shift={(-0.2000, 0.0000, 0.0000)}];
  \dodeca[green,thick,fill=green,fill opacity=0.2]
    \end{scope}
  \begin{scope}[shift={(0.0000, 0.3464, 0.0000)}];
  \dodeca[blue,thick,fill=blue,fill opacity=0.2]
    \end{scope}
  \begin{scope}[shift={(0.2000, 0.0000, 0.0000)}];
  \dodeca[green,thick,fill=green,fill opacity=0.2]
    \end{scope}
  \begin{scope}[shift={(0.0000, -0.3464, 0.0000)}];
  \dodeca[blue,thick,fill=blue,fill opacity=0.2]
    \end{scope}
}

\newcommand\hexconf{%
  \begin{scope}[shift={(-0.4000, 0.0000, 0.0000)}];
  \dodeca[green,thick,fill=green,fill opacity=0.2]
    \end{scope}
  \begin{scope}[shift={(-0.2000, 0.3464, 0.0000)}];
  \dodeca[blue,thick,fill=blue,fill opacity=0.2]
    \end{scope}
  \begin{scope}[shift={(-0.2000, -0.3464, 0.0000)}];
  \dodeca[blue,thick,fill=blue,fill opacity=0.2]
    \end{scope}
  \begin{scope}[shift={(0.2000, 0.3464, 0.0000)}];
  \dodeca[green,thick,fill=green,fill opacity=0.2]
    \end{scope}
  \begin{scope}[shift={(0.2000, -0.3464, 0.0000)}];
  \dodeca[green,thick,fill=green,fill opacity=0.2]
    \end{scope}
  \begin{scope}[shift={(0.4000, 0.0000, 0.0000)}];
  \dodeca[blue,thick,fill=blue,fill opacity=0.2]
    \end{scope}
}

\newcommand\tetraconf{%
  \begin{scope}[shift={(0.0000, 0.0000, 0.2449)}]
    \dodeca[black,thick,fill=green,fill opacity=0.2];
  \end{scope}
  \begin{scope}[shift={(0.0000, -0.2309, -0.0816)}]
    \dodeca[black,thick,fill=blue,fill opacity=0.2];
  \end{scope}
  \begin{scope}[shift={(-0.2000, 0.1155, -0.0816)}]
    \dodeca[black,thick,fill=green,fill opacity=0.2];
  \end{scope}
  \begin{scope}[shift={(0.2000, 0.1155, -0.0816)}]
    \dodeca[black,thick,fill=blue,fill opacity=0.2];
  \end{scope}
}

\newcommand\tetradconf{%
  \begin{scope}[shift={(0.0000, 0.0000, 0.7348)}]
    \dodeca[black,thick,fill=green,fill opacity=0.2];
  \end{scope}
  \begin{scope}[shift={(0.0000, -0.2309, 0.4082)}]
    \dodeca[black,thick,fill=blue,fill opacity=0.2];
  \end{scope}
  \begin{scope}[shift={(-0.2000, 0.1155, 0.4082)}]
    \dodeca[black,thick,fill=green,fill opacity=0.2];
  \end{scope}
  \begin{scope}[shift={(0.2000, 0.1155, 0.4082)}]
    \dodeca[black,thick,fill=blue,fill opacity=0.2];
  \end{scope}
  \begin{scope}[shift={(0.0000, -0.4619, 0.0816)}]
    \dodeca[black,thick,fill=green,fill opacity=0.2];
  \end{scope}
  \begin{scope}[shift={(0.0000, -0.6928, -0.2449)}]
    \dodeca[black,thick,fill=blue,fill opacity=0.2];
  \end{scope}
  \begin{scope}[shift={(-0.2000, -0.3464, -0.2449)}]
    \dodeca[black,thick,fill=green,fill opacity=0.2];
  \end{scope}
  \begin{scope}[shift={(0.2000, -0.3464, -0.2449)}]
    \dodeca[black,thick,fill=blue,fill opacity=0.2];
  \end{scope}
  \begin{scope}[shift={(-0.4000, 0.2309, 0.0816)}]
    \dodeca[black,thick,fill=green,fill opacity=0.2];
  \end{scope}
  \begin{scope}[shift={(-0.4000, 0.0000, -0.2449)}]
    \dodeca[black,thick,fill=blue,fill opacity=0.2];
  \end{scope}
  \begin{scope}[shift={(-0.6000, 0.3464, -0.2449)}]
    \dodeca[black,thick,fill=green,fill opacity=0.2];
  \end{scope}
  \begin{scope}[shift={(-0.2000, 0.3464, -0.2449)}]
    \dodeca[black,thick,fill=blue,fill opacity=0.2];
  \end{scope}
  \begin{scope}[shift={(0.4000, 0.2309, 0.0816)}]
    \dodeca[black,thick,fill=green,fill opacity=0.2];
  \end{scope}
  \begin{scope}[shift={(0.4000, 0.0000, -0.2449)}]
    \dodeca[black,thick,fill=blue,fill opacity=0.2];
  \end{scope}
  \begin{scope}[shift={(0.2000, 0.3464, -0.2449)}]
    \dodeca[black,thick,fill=green,fill opacity=0.2];
  \end{scope}
  \begin{scope}[shift={(0.6000, 0.3464, -0.2449)}]
    \dodeca[black,thick,fill=blue,fill opacity=0.2];
  \end{scope}
}

\newcommand\dodecaconf{%
\tdplotsetrotatedcoords{127.7612}{-41.8103}{-127.7612}
     \begin{scope}[shift={(-0.4000, -0.0000, 0.0000)},tdplot_rotated_coords];
     \dodeca[blue,thick,fill=blue,fill opacity=0.2]
     \end{scope}
     \tdplotsetrotatedcoords{-112.2388}{-41.8103}{112.2388}
     \begin{scope}[shift={(-0.6000, 0.3464, 0.0000)},tdplot_rotated_coords];
     \dodeca[green,thick,fill=green,fill opacity=0.2]
     \end{scope}
     \tdplotsetrotatedcoords{-112.2388}{-41.8103}{112.2388}
     \begin{scope}[shift={(-0.2000, 0.3464, 0.0000)},tdplot_rotated_coords];
     \dodeca[green,thick,fill=green,fill opacity=0.2]
     \end{scope}
     \tdplotsetrotatedcoords{7.7612}{-41.8103}{-7.7612}
     \begin{scope}[shift={(-0.0000, 0.6928, 0.0000)},tdplot_rotated_coords];
     \dodeca[blue,thick,fill=blue,fill opacity=0.2]
     \end{scope}
     \tdplotsetrotatedcoords{7.7612}{-41.8103}{-7.7612}
     \begin{scope}[shift={(0.2000, 0.3464, 0.0000)},tdplot_rotated_coords];
     \dodeca[blue,thick,fill=blue,fill opacity=0.2]
     \end{scope}
     \tdplotsetrotatedcoords{127.7612}{-41.8103}{-127.7612}
     \begin{scope}[shift={(0.6000, 0.3464, 0.0000)},tdplot_rotated_coords];
     \dodeca[green,thick,fill=green,fill opacity=0.2]
     \end{scope}
     \tdplotsetrotatedcoords{127.7612}{-41.8103}{-127.7612}
     \begin{scope}[shift={(0.4000, -0.0000, 0.0000)},tdplot_rotated_coords];
     \dodeca[green,thick,fill=green,fill opacity=0.2]
     \end{scope}
     \tdplotsetrotatedcoords{-112.2388}{-41.8103}{112.2388}
     \begin{scope}[shift={(0.6000, -0.3464, 0.0000)},tdplot_rotated_coords];
     \dodeca[blue,thick,fill=blue,fill opacity=0.2]
     \end{scope}
     \tdplotsetrotatedcoords{-112.2388}{-41.8103}{112.2388}
     \begin{scope}[shift={(0.2000, -0.3464, 0.0000)},tdplot_rotated_coords];
     \dodeca[blue,thick,fill=blue,fill opacity=0.2]
     \end{scope}
     \tdplotsetrotatedcoords{7.7612}{-41.8103}{-7.7612}
     \begin{scope}[shift={(-0.0000, -0.6928, 0.0000)},tdplot_rotated_coords];
     \dodeca[green,thick,fill=green,fill opacity=0.2]
     \end{scope}
     \tdplotsetrotatedcoords{7.7612}{-41.8103}{-7.7612}
     \begin{scope}[shift={(-0.2000, -0.3464, 0.0000)},tdplot_rotated_coords];
     \dodeca[green,thick,fill=green,fill opacity=0.2]
     \end{scope}
     \tdplotsetrotatedcoords{127.7612}{-41.8103}{-127.7612}
     \begin{scope}[shift={(-0.6000, -0.3464, 0.0000)},tdplot_rotated_coords];
     \dodeca[blue,thick,fill=blue,fill opacity=0.2]
     \end{scope}
}

\newcommand\getdodecaopm{%
\coordinate (o_A) at ({0}, {0}, {0});
     \coordinate (o_B) at ({(2^(1/2)*3^(1/2))/(12*(5^(1/2)/2 + 1/2)^2) - (2^(1/2)*3^(1/2))/(12*(5^(1/2)/2 + 1/2)) - (2^(1/2)*3^(1/2)*(5^(1/2)/2 + 1/2))/12}, {- 2^(1/2)/4 - (2^(1/2)*(5^(1/2)/2 + 1/2))/4}, {(2^(1/2)*3^(1/2))/12 - (2^(1/2)*3^(1/2)*(5^(1/2)/2 + 1/2))/12 + (2^(1/2)*3^(1/2))/(12*(5^(1/2)/2 + 1/2))});
     \coordinate (vA_1) at ({-30^(1/2)/12}, {-2^(1/2)/4}, {-6^(1/2)/12});
     \coordinate (vA_15) at ({-6^(1/2)/6}, {0}, {-30^(1/2)/12});
     \coordinate (vA_18) at ({-(6^(1/2)*(5^(1/2) + 1)^2)/48}, {2^(1/2)/8 - 10^(1/2)/8}, {6^(1/2)/12});
     \coordinate (vA_17) at ({-(6^(1/2)*(5^(1/2) + 1)^2)/48}, {10^(1/2)/8 - 2^(1/2)/8}, {6^(1/2)/12});
     \coordinate (vA_2) at ({-6^(1/2)/12}, {-2^(1/2)/4}, {30^(1/2)/12});
     \coordinate (vA_16) at ({0}, {0}, {6^(1/2)/4});
     \coordinate (vA_9) at ({6^(1/2)/8 - 30^(1/2)/24}, {-(2^(1/2)*(5^(1/2)/2 + 1/2))/4}, {6^(1/2)/12});
     \coordinate (vA_6) at ({30^(1/2)/12}, {-2^(1/2)/4}, {6^(1/2)/12});
     \coordinate (vA_10) at ({30^(1/2)/24 - 6^(1/2)/8}, {-(2^(1/2)*(5^(1/2)/2 + 1/2))/4}, {-6^(1/2)/12});
     \coordinate (vA_5) at ({6^(1/2)/12}, {-2^(1/2)/4}, {-30^(1/2)/12});
     \coordinate (vA_3) at ({-30^(1/2)/12}, {2^(1/2)/4}, {-6^(1/2)/12});
     \coordinate (vA_3) at ({-30^(1/2)/12}, {2^(1/2)/4}, {-6^(1/2)/12});
     \coordinate (vA_4) at ({-6^(1/2)/12}, {2^(1/2)/4}, {30^(1/2)/12});
     \coordinate (vA_4) at ({-6^(1/2)/12}, {2^(1/2)/4}, {30^(1/2)/12});
     \coordinate (vA_13) at ({0}, {0}, {-6^(1/2)/4});
     \coordinate (vA_20) at ({(6^(1/2)*(5^(1/2) + 1)^2)/48}, {2^(1/2)/8 - 10^(1/2)/8}, {-6^(1/2)/12});
     \coordinate (vA_14) at ({6^(1/2)/6}, {0}, {30^(1/2)/12});
     \coordinate (vA_20) at ({(6^(1/2)*(5^(1/2) + 1)^2)/48}, {2^(1/2)/8 - 10^(1/2)/8}, {-6^(1/2)/12});
     \coordinate (vA_13) at ({0}, {0}, {-6^(1/2)/4});
     \coordinate (vA_14) at ({6^(1/2)/6}, {0}, {30^(1/2)/12});
     \coordinate (vB_7) at ({(5*6^(1/2))/24 - 30^(1/2)/8}, {-(2^(1/2)*(5^(1/2)/2 + 1/2))/4}, {-30^(1/2)/12});
     \coordinate (vB_13) at ({6^(1/2)/8 - 30^(1/2)/8}, {- (3*2^(1/2))/8 - 10^(1/2)/8}, {-6^(1/2)/4});
     \coordinate (vB_12) at ({-30^(1/2)/12}, {-2^(1/2)/4}, {-6^(1/2)/12});
     \coordinate (vB_3) at ({6^(1/2)/8 - (5*30^(1/2))/24}, {-(2^(1/2)*(5^(1/2)/2 + 1/2))/4}, {-6^(1/2)/12});
     \coordinate (vB_11) at ({6^(1/2)/4 - 30^(1/2)/6}, {-2^(1/2)/4}, {6^(1/2)/12});
     \coordinate (vB_4) at ({6^(1/2)/24 - 30^(1/2)/8}, {-(2^(1/2)*(5^(1/2)/2 + 1/2))/4}, {30^(1/2)/12});
     \coordinate (vB_8) at ({6^(1/2)/8 - 30^(1/2)/24}, {-(2^(1/2)*(5^(1/2)/2 + 1/2))/4}, {6^(1/2)/12});
     \coordinate (vB_14) at ({(7*6^(1/2))/24 - 30^(1/2)/8}, {- (3*2^(1/2))/8 - 10^(1/2)/8}, {30^(1/2)/12});
     \coordinate (vB_19) at ({6^(1/2)/4 - 30^(1/2)/12}, {-2^(1/2)/2}, {-6^(1/2)/12});
     \coordinate (vB_20) at ({6^(1/2)/4 - 30^(1/2)/12}, {- 2^(1/2)/4 - 10^(1/2)/4}, {-6^(1/2)/12});
     \coordinate (vB_17) at ({-30^(1/2)/6}, {-2^(1/2)/2}, {6^(1/2)/12});
     \coordinate (vB_15) at ({- 6^(1/2)/24 - 30^(1/2)/8}, {- (3*2^(1/2))/8 - 10^(1/2)/8}, {-30^(1/2)/12});
     \coordinate (vB_16) at ({6^(1/2)/8 - 30^(1/2)/8}, {- (3*2^(1/2))/8 - 10^(1/2)/8}, {6^(1/2)/4});
     \coordinate (vB_17) at ({-30^(1/2)/6}, {-2^(1/2)/2}, {6^(1/2)/12});
     \coordinate (vB_5) at ({(5*6^(1/2))/24 - 30^(1/2)/8}, {- (5*2^(1/2))/8 - 10^(1/2)/8}, {-30^(1/2)/12});
     \coordinate (vB_5) at ({(5*6^(1/2))/24 - 30^(1/2)/8}, {- (5*2^(1/2))/8 - 10^(1/2)/8}, {-30^(1/2)/12});
     \coordinate (vB_6) at ({6^(1/2)/8 - 30^(1/2)/24}, {- (5*2^(1/2))/8 - 10^(1/2)/8}, {6^(1/2)/12});
     \coordinate (vB_6) at ({6^(1/2)/8 - 30^(1/2)/24}, {- (5*2^(1/2))/8 - 10^(1/2)/8}, {6^(1/2)/12});
     \coordinate (vB_15) at ({- 6^(1/2)/24 - 30^(1/2)/8}, {- (3*2^(1/2))/8 - 10^(1/2)/8}, {-30^(1/2)/12});
     \coordinate (vB_16) at ({6^(1/2)/8 - 30^(1/2)/8}, {- (3*2^(1/2))/8 - 10^(1/2)/8}, {6^(1/2)/4});
     \coordinate (vAi1_17_3) at ({- 30^(1/2)/40 - (7*6^(1/2)*(5^(1/2) + 1)^2)/480}, {(7*10^(1/2))/80 - 2^(1/2)/80}, {6^(1/2)/30});
     \coordinate (vAi2_17_3) at ({- (7*30^(1/2))/120 - (6^(1/2)*(5^(1/2) + 1)^2)/160}, {(11*2^(1/2))/80 + (3*10^(1/2))/80}, {-6^(1/2)/30});
     \coordinate (vAi1_15_3) at ({- (7*6^(1/2))/60 - 30^(1/2)/40}, {(3*2^(1/2))/40}, {- 6^(1/2)/40 - (7*30^(1/2))/120});
     \coordinate (vAi2_15_3) at ({- 6^(1/2)/20 - (7*30^(1/2))/120}, {(7*2^(1/2))/40}, {- (7*6^(1/2))/120 - 30^(1/2)/40});
     \coordinate (vAi1_16_4) at ({-6^(1/2)/40}, {(3*2^(1/2))/40}, {(7*6^(1/2))/40 + 30^(1/2)/40});
     \coordinate (vAi2_16_4) at ({-(7*6^(1/2))/120}, {(7*2^(1/2))/40}, {(3*6^(1/2))/40 + (7*30^(1/2))/120});
     \coordinate (vAi1_17_4) at ({- 6^(1/2)/40 - (7*6^(1/2)*(5^(1/2) + 1)^2)/480}, {(7*10^(1/2))/80 - 2^(1/2)/80}, {(7*6^(1/2))/120 + 30^(1/2)/40});
     \coordinate (vAi2_17_4) at ({- (7*6^(1/2))/120 - (6^(1/2)*(5^(1/2) + 1)^2)/160}, {(11*2^(1/2))/80 + (3*10^(1/2))/80}, {6^(1/2)/40 + (7*30^(1/2))/120});
     \coordinate (vAi1_5_13) at ({(7*6^(1/2))/120}, {-(7*2^(1/2))/40}, {- (3*6^(1/2))/40 - (7*30^(1/2))/120});
     \coordinate (vAi2_5_13) at ({6^(1/2)/40}, {-(3*2^(1/2))/40}, {- (7*6^(1/2))/40 - 30^(1/2)/40});
     \coordinate (vAi1_5_20) at ({(7*6^(1/2))/120 + (6^(1/2)*(5^(1/2) + 1)^2)/160}, {- (11*2^(1/2))/80 - (3*10^(1/2))/80}, {- 6^(1/2)/40 - (7*30^(1/2))/120});
     \coordinate (vAi2_5_20) at ({6^(1/2)/40 + (7*6^(1/2)*(5^(1/2) + 1)^2)/480}, {2^(1/2)/80 - (7*10^(1/2))/80}, {- (7*6^(1/2))/120 - 30^(1/2)/40});
     \coordinate (vAi1_6_14) at ({6^(1/2)/20 + (7*30^(1/2))/120}, {-(7*2^(1/2))/40}, {(7*6^(1/2))/120 + 30^(1/2)/40});
     \coordinate (vAi2_6_14) at ({(7*6^(1/2))/60 + 30^(1/2)/40}, {-(3*2^(1/2))/40}, {6^(1/2)/40 + (7*30^(1/2))/120});
     \coordinate (vAi1_6_20) at ({(7*30^(1/2))/120 + (6^(1/2)*(5^(1/2) + 1)^2)/160}, {- (11*2^(1/2))/80 - (3*10^(1/2))/80}, {6^(1/2)/30});
     \coordinate (vAi2_6_20) at ({30^(1/2)/40 + (7*6^(1/2)*(5^(1/2) + 1)^2)/480}, {2^(1/2)/80 - (7*10^(1/2))/80}, {-6^(1/2)/30});
     \coordinate (vAi1_15_13) at ({-(7*6^(1/2))/60}, {0}, {- (3*6^(1/2))/40 - (7*30^(1/2))/120});
     \coordinate (vAi2_15_13) at ({-6^(1/2)/20}, {0}, {- (7*6^(1/2))/40 - 30^(1/2)/40});
     \coordinate (vAi1_16_14) at ({6^(1/2)/20}, {0}, {(7*6^(1/2))/40 + 30^(1/2)/40});
     \coordinate (vAi2_16_14) at ({(7*6^(1/2))/60}, {0}, {(3*6^(1/2))/40 + (7*30^(1/2))/120});
     \coordinate (vBi1_3_17) at ({(7*6^(1/2))/80 - (47*30^(1/2))/240}, {- (3*2^(1/2))/20 - (7*2^(1/2)*(5^(1/2)/2 + 1/2))/40}, {-6^(1/2)/30});
     \coordinate (vBi2_3_17) at ({(3*6^(1/2))/80 - (43*30^(1/2))/240}, {- (7*2^(1/2))/20 - (3*2^(1/2)*(5^(1/2)/2 + 1/2))/40}, {6^(1/2)/30});
     \coordinate (vBi1_3_15) at ({(3*6^(1/2))/40 - (11*30^(1/2))/60}, {- (9*2^(1/2))/80 - (3*10^(1/2))/80 - (7*2^(1/2)*(5^(1/2)/2 + 1/2))/40}, {- (7*6^(1/2))/120 - 30^(1/2)/40});
     \coordinate (vBi2_3_15) at ({6^(1/2)/120 - (3*30^(1/2))/20}, {- (21*2^(1/2))/80 - (7*10^(1/2))/80 - (3*2^(1/2)*(5^(1/2)/2 + 1/2))/40}, {- 6^(1/2)/40 - (7*30^(1/2))/120});
     \coordinate (vBi1_4_16) at ({6^(1/2)/15 - 30^(1/2)/8}, {- (9*2^(1/2))/80 - (3*10^(1/2))/80 - (7*2^(1/2)*(5^(1/2)/2 + 1/2))/40}, {(3*6^(1/2))/40 + (7*30^(1/2))/120});
     \coordinate (vBi2_4_16) at ({6^(1/2)/10 - 30^(1/2)/8}, {- (21*2^(1/2))/80 - (7*10^(1/2))/80 - (3*2^(1/2)*(5^(1/2)/2 + 1/2))/40}, {(7*6^(1/2))/40 + 30^(1/2)/40});
     \coordinate (vBi1_4_17) at ({(7*6^(1/2))/240 - (11*30^(1/2))/80}, {- (3*2^(1/2))/20 - (7*2^(1/2)*(5^(1/2)/2 + 1/2))/40}, {6^(1/2)/40 + (7*30^(1/2))/120});
     \coordinate (vBi2_4_17) at ({6^(1/2)/80 - (37*30^(1/2))/240}, {- (7*2^(1/2))/20 - (3*2^(1/2)*(5^(1/2)/2 + 1/2))/40}, {(7*6^(1/2))/120 + 30^(1/2)/40});
     \coordinate (vBi1_13_5) at ({(3*6^(1/2))/20 - 30^(1/2)/8}, {- (9*2^(1/2))/20 - 10^(1/2)/8}, {- (7*6^(1/2))/40 - 30^(1/2)/40});
     \coordinate (vBi2_13_5) at ({(11*6^(1/2))/60 - 30^(1/2)/8}, {- (11*2^(1/2))/20 - 10^(1/2)/8}, {- (3*6^(1/2))/40 - (7*30^(1/2))/120});
     \coordinate (vBi1_20_5) at ({(19*6^(1/2))/80 - (23*30^(1/2))/240}, {- (29*2^(1/2))/80 - (17*10^(1/2))/80}, {- (7*6^(1/2))/120 - 30^(1/2)/40});
     \coordinate (vBi2_20_5) at ({(53*6^(1/2))/240 - (9*30^(1/2))/80}, {- (41*2^(1/2))/80 - (13*10^(1/2))/80}, {- 6^(1/2)/40 - (7*30^(1/2))/120});
     \coordinate (vBi1_14_6) at ({(29*6^(1/2))/120 - 30^(1/2)/10}, {- (9*2^(1/2))/20 - 10^(1/2)/8}, {6^(1/2)/40 + (7*30^(1/2))/120});
     \coordinate (vBi2_14_6) at ({(7*6^(1/2))/40 - 30^(1/2)/15}, {- (11*2^(1/2))/20 - 10^(1/2)/8}, {(7*6^(1/2))/120 + 30^(1/2)/40});
     \coordinate (vBi1_20_6) at ({(17*6^(1/2))/80 - (17*30^(1/2))/240}, {- (29*2^(1/2))/80 - (17*10^(1/2))/80}, {-6^(1/2)/30});
     \coordinate (vBi2_20_6) at ({(13*6^(1/2))/80 - (13*30^(1/2))/240}, {- (41*2^(1/2))/80 - (13*10^(1/2))/80}, {6^(1/2)/30});
     \coordinate (vBi1_13_15) at ({(3*6^(1/2))/40 - 30^(1/2)/8}, {- (3*2^(1/2))/8 - 10^(1/2)/8}, {- (7*6^(1/2))/40 - 30^(1/2)/40});
     \coordinate (vBi2_13_15) at ({6^(1/2)/120 - 30^(1/2)/8}, {- (3*2^(1/2))/8 - 10^(1/2)/8}, {- (3*6^(1/2))/40 - (7*30^(1/2))/120});
     \coordinate (vBi1_14_16) at ({(29*6^(1/2))/120 - 30^(1/2)/8}, {- (3*2^(1/2))/8 - 10^(1/2)/8}, {(3*6^(1/2))/40 + (7*30^(1/2))/120});
     \coordinate (vBi2_14_16) at ({(7*6^(1/2))/40 - 30^(1/2)/8}, {- (3*2^(1/2))/8 - 10^(1/2)/8}, {(7*6^(1/2))/40 + 30^(1/2)/40});
     \coordinate (x1) at ({3^(1/2)/12 - 15^(1/2)/12 - 30^(1/2)/12}, {5^(1/2)/4 - 2^(1/2)/4 - 1/4}, {3^(1/2)/6 - 6^(1/2)/12 + 15^(1/2)/6});
     \coordinate (x2) at ({(6^(1/2)*(25 - 11*5^(1/2))^(1/2))/4 + (2*30^(1/2)*(25 - 11*5^(1/2))^(1/2))/15 - 30^(1/2)/12}, {- (2^(1/2)*(25 - 11*5^(1/2))^(1/2))/4 - (10^(1/2)*(25 - 11*5^(1/2))^(1/2))/10 - 2^(1/2)/4}, {(6^(1/2)*(25 - 11*5^(1/2))^(1/2))/12 + (30^(1/2)*(25 - 11*5^(1/2))^(1/2))/20 - 6^(1/2)/12});
     
     \begin{scope}[very thick];
     \draw[teal] (vA_1) -- (vA_18) -- (vA_2) -- (vA_9) -- (vA_10) -- (vA_1);
     \draw[magenta] (vB_7) -- (vB_12) -- (vB_11) -- (vB_8) -- (vB_19) -- (vB_7);
     
     \end{scope}
     \draw[teal] (vA_1) -- (vA_15);
     \draw[teal] (vA_18) -- (vA_17);
     \draw[teal] (vA_2) -- (vA_16);
     \draw[teal] (vA_9) -- (vA_6);
     \draw[teal] (vA_10) -- (vA_5);
     \draw[teal] (vA_17) -- (vAi1_17_3);
     \draw[teal, dotted] (vAi1_17_3) -- (vAi2_17_3);
     \draw[teal] (vA_15) -- (vAi1_15_3);
     \draw[teal, dotted] (vAi1_15_3) -- (vAi2_15_3);
     \draw[teal] (vA_16) -- (vAi1_16_4);
     \draw[teal, dotted] (vAi1_16_4) -- (vAi2_16_4);
     \draw[teal] (vA_17) -- (vAi1_17_4);
     \draw[teal, dotted] (vAi1_17_4) -- (vAi2_17_4);
     \draw[teal] (vA_5) -- (vAi1_5_13);
     \draw[teal, dotted] (vAi1_5_13) -- (vAi2_5_13);
     \draw[teal] (vA_5) -- (vAi1_5_20);
     \draw[teal, dotted] (vAi1_5_20) -- (vAi2_5_20);
     \draw[teal] (vA_6) -- (vAi1_6_14);
     \draw[teal, dotted] (vAi1_6_14) -- (vAi2_6_14);
     \draw[teal] (vA_6) -- (vAi1_6_20);
     \draw[teal, dotted] (vAi1_6_20) -- (vAi2_6_20);
     \draw[teal] (vA_15) -- (vAi1_15_13);
     \draw[teal, dotted] (vAi1_15_13) -- (vAi2_15_13);
     \draw[teal] (vA_16) -- (vAi1_16_14);
     \draw[teal, dotted] (vAi1_16_14) -- (vAi2_16_14);
     \draw[magenta] (vB_7) -- (vB_13);
     \draw[magenta] (vB_12) -- (vB_3);
     \draw[magenta] (vB_11) -- (vB_4);
     \draw[magenta] (vB_8) -- (vB_14);
     \draw[magenta] (vB_19) -- (vB_20);
     \draw[magenta] (vB_3) -- (vBi1_3_17);
     \draw[magenta, dotted] (vBi1_3_17) -- (vBi2_3_17);
     \draw[magenta] (vB_3) -- (vBi1_3_15);
     \draw[magenta, dotted] (vBi1_3_15) -- (vBi2_3_15);
     \draw[magenta] (vB_4) -- (vBi1_4_16);
     \draw[magenta, dotted] (vBi1_4_16) -- (vBi2_4_16);
     \draw[magenta] (vB_4) -- (vBi1_4_17);
     \draw[magenta, dotted] (vBi1_4_17) -- (vBi2_4_17);
     \draw[magenta] (vB_13) -- (vBi1_13_5);
     \draw[magenta, dotted] (vBi1_13_5) -- (vBi2_13_5);
     \draw[magenta] (vB_20) -- (vBi1_20_5);
     \draw[magenta, dotted] (vBi1_20_5) -- (vBi2_20_5);
     \draw[magenta] (vB_14) -- (vBi1_14_6);
     \draw[magenta, dotted] (vBi1_14_6) -- (vBi2_14_6);
     \draw[magenta] (vB_20) -- (vBi1_20_6);
     \draw[magenta, dotted] (vBi1_20_6) -- (vBi2_20_6);
     \draw[magenta] (vB_13) -- (vBi1_13_15);
     \draw[magenta, dotted] (vBi1_13_15) -- (vBi2_13_15);
     \draw[magenta] (vB_14) -- (vBi1_14_16);
     \draw[magenta, dotted] (vBi1_14_16) -- (vBi2_14_16);
     
     \begin{scope}[red, thick, canvas is plane={O(\pgfpointanchor{vA_1}{center})x(\pgfpointanchor{x1}{center})y(\pgfpointanchor{x2}{center})}];
     \draw (vA_9) circle (0.03);
     \draw (vA_1) circle (0.03);
     
     \end{scope}
     \draw[dashed] (vA_9) -- (vA_1);
     \draw[fill, tdplot_screen_coords] (o_A) circle (0.01);
     \draw[fill, tdplot_screen_coords] (o_B) circle (0.01);
     \draw[dashed, -stealth, shorten >=2pt] (o_A) -- (o_B.west);
}

\tikzset{
  bodyframe/.pic={
    \draw (0, 0, 0) -- ({1/sqrt(2)}, {-1/sqrt(2)}, 0) node[below] {$e_x^\mathcal{B}$};
    \draw (0, 0, 0) -- ({1/sqrt(2)}, {1/sqrt(2)}, 0) node[above] {$e_y^\mathcal{B}$};
    \draw (0, 0, 0) -- (0, 0, 1) node[above] {$e_z^\mathcal{B}$};
  }
}

\tikzset{
  pics/filledtetra/.style n args={1}{ code={
#1f;
    \fill[gray!10] (af.center) -- (bf.center) -- (cf.center) -- cycle;
    \fill[gray!15] (af.center) -- (bf.center) -- (df.center) -- cycle;
    \fill[gray!20] (af.center) -- (cf.center) -- (df.center) -- cycle;
    \fill[gray!25] (bf.center) -- (cf.center) -- (df.center) -- cycle;
  }
  }}

\tikzset{
  pics/frametetra/.style n args={1}{ code={
#1f;
    \draw (af) -- (bf);
    \draw (af) -- (cf);
    \draw (af) -- (df);
    \draw (bf) -- (cf);
    \draw (bf) -- (df);
    \draw (cf) -- (df);
  }
  }}

\tikzset{
  pics/tetrarotor/.style n args={1}{ code={
#1r;
    \begin{scope}[canvas is plane={O(\pgfpointanchor{abcr}{center})x(\pgfpointanchor{abcur}{center})y(\pgfpointanchor{abcvr}{center})}]
      \draw[fill,fill opacity=0.2] (abcr) circle ({1 / 2 / sqrt(3)});
    \end{scope}
  }
  }}

\tikzset{
  thrust/.pic={
    \draw (abc.center) -- (barycentric cs:o=-1.5,abc=2.5);
    \draw (abd.center) -- (barycentric cs:o=-1.5,abd=2.5);
    \draw (acd.center) -- (barycentric cs:o=1.5,acd=-0.5);
    \draw (bcd.center) -- (barycentric cs:o=1.5,bcd=-0.5);
  }
}

\tikzset{
  torques/.pic={
    \draw (abc.center) -- (barycentric cs:o=-1.5,abc=2.5);
    \draw (abd.center) -- (barycentric cs:o=-1.5,abd=2.5);
    \draw (acd.center) -- (barycentric cs:o=-1.5,acd=2.5);
    \draw (bcd.center) -- (barycentric cs:o=-1.5,bcd=2.5);
  }
}

\tikzset{
  rotors/.pic={
    \begin{scope}[canvas is plane={O(\pgfpointanchor{abc}{center})x(\pgfpointanchor{abcu}{center})y(\pgfpointanchor{abcv}{center})}]
      \draw[teal] (abc) circle ({1 / 2 / sqrt(3)});
      \draw[fill] (abc) circle ({1 / 80});
    \end{scope}
    \begin{scope}[canvas is plane={O(\pgfpointanchor{abd}{center})x(\pgfpointanchor{abdu}{center})y(\pgfpointanchor{abdv}{center})}]
      \draw[teal] (abd) circle ({1 / 2 / sqrt(3)});
      \draw[fill] (abd) circle ({1 / 80});
    \end{scope}
    \begin{scope}[canvas is plane={O(\pgfpointanchor{acd}{center})x(\pgfpointanchor{acdu}{center})y(\pgfpointanchor{acdv}{center})}]
      \draw[magenta] (acd) circle ({1 / 2 / sqrt(3)});
      \draw[fill] (acd) circle ({1 / 80});
    \end{scope}
    \begin{scope}[canvas is plane={O(\pgfpointanchor{bcd}{center})x(\pgfpointanchor{bcdu}{center})y(\pgfpointanchor{bcdv}{center})}]
      \draw[magenta] (bcd) circle ({1 / 2 / sqrt(3)});
      \draw[fill] (bcd) circle ({1 / 80});
    \end{scope}
  }
}

              \newcommand\sierpinskitetrahedron[4][0]{
                \begin{scope}[every node/.append style={transform shape}]
                  \ifnum #4 = 0 \relax
#3;
                  \else
#2[\number\thebnumexpr2**(#4-1)\relax]{s#4};
                  \begin{scope}[shift={($(as#4)$))}]
                  \sierpinskitetrahedron{#2}{#3}{\number\numexpr#4-1\relax}
                \end{scope}
                \begin{scope}[shift={($(bs#4)$))}]
                  \sierpinskitetrahedron{#2}{#3}{\number\numexpr#4-1\relax}
                \end{scope}
                \begin{scope}[shift={($(cs#4)$))}]
                  \sierpinskitetrahedron{#2}{#3}{\number\numexpr#4-1\relax}
                \end{scope}
                \begin{scope}[shift={($(ds#4)$))}]
                  \sierpinskitetrahedron{#2}{#3}{\number\numexpr#4-1\relax}
                \end{scope}
                \fi
                  \end{scope}

                \ifnum #1 = 1 \relax

                \draw[->, thick] (0, 0, 0) node[above right] {$O$} -- (b.center) node[right] {$p^\decrement{#2}_2$};
                \draw[->, thick] (0, 0, 0) -- (c.center) node[left] {$p^\decrement{#2}_3$};
                \draw[->, thick] (0, 0, 0) -- (d.center) node[above right] {$p^\decrement{#2}_4$};
                \draw[->, thick] (0, 0, 0) -- (a.center) node[below] {$p^\decrement{#2}_1$};
                \fi
              }

\tikzset{
  pics/beam/.style n args={6}{ code={
    \pgfmathsetmacro{\innerradius}{0.15}
    \pgfmathsetmacro{\outerradius}{0.25}
    \pgfmathsetmacro{\xa}{#1}
    \pgfmathsetmacro{\ya}{#2}
    \pgfmathsetmacro{\ta}{#3}
    \pgfmathsetmacro{\xb}{#4}
    \pgfmathsetmacro{\yb}{#5}
    \pgfmathsetmacro{\tb}{#6}
    \pgfmathsetmacro{\stiffness}{2}

    \draw[gray!90, thick, line join=round]
      ([shift=({\ta+90}:\outerradius)] \xa,\ya) {[rotate=\ta] arc [start angle=90,end angle=270,x radius=0.5*\outerradius,y radius=\outerradius]}
    .. controls +(\ta:\stiffness)
      and +({\tb+180}:\stiffness)
      .. ([shift=({\tb+270}:\outerradius)] \xb,\yb) {[rotate=\tb] arc [start angle=-90,end angle=450,x radius=0.5*\outerradius,y radius=\outerradius]}
    .. controls +({\tb+180}:\stiffness)
      and +(\ta:\stiffness)
      .. cycle;
    \draw[gray!90, thick, rotate around={\tb:(\xb,\yb)}]
      (\xb,\yb) circle [x radius=0.5*\innerradius,y radius=\innerradius];
    \draw[thick,dashed]
      ([shift=(\ta+180:2*\outerradius)] \xa,\ya)
      -- (\xa,\ya)
      .. controls +(\ta:\stiffness)
      and +({\tb+180}:\stiffness)
      .. (\xb,\yb)
      -- ([shift=(\tb:2*\outerradius)] \xb,\yb);
    \begin{scope}[rotate around={\ta:(\xa,\ya)}]
      \draw[fill] (\xa,\ya) circle (0.03);
    \draw[-stealth] (\xa,\ya) -- ([shift=(0:0.5)] \xa,\ya);
    \draw[-stealth] (\xa,\ya) -- ([shift=(40:0.4)] \xa,\ya);
    \draw[-stealth] (\xa,\ya) -- ([shift=(90:0.5)] \xa,\ya);
    \end{scope}
    \begin{scope}[rotate around={\tb:(\xb,\yb)}]
      \draw[fill] (\xb,\yb) circle (0.03);
    \draw[-stealth] (\xb,\yb) -- ([shift=(0:0.5)] \xb,\yb);
    \draw[-stealth] (\xb,\yb) -- ([shift=(40:0.4)] \xb,\yb);
    \draw[-stealth] (\xb,\yb) -- ([shift=(90:0.5)] \xb,\yb);
    \end{scope}
  }}
}

\tikzset{
  pics/frame/.style n args={3}{ code={
    \ifdefmacro{#3}{
      \foreach\x\y\t\v in #1{
#3{\t}{\v}
        \ifthenelse{\equal{#2}{}}
        {
          \draw[\opte] (v\x) -- (v\y);
        }{
          \draw[\opte] ($(v\x)+#2*(d\x)$) -- ($(v\y)+#2*(d\y)$);
        }
      };
    }{
      \foreach\x\y in #1{
        \ifthenelse{\equal{#2}{}}
        {
          \draw (v\x) -- (v\y);
        }{
          \draw ($(v\x)+#2*(d\x)$) -- ($(v\y)+#2*(d\y)$);
        }
      };
    }
  }}
}

\tikzset{
  pics/frameForces/.style n args={3}{ code={
    \foreach\x in {1,...,#1}{
      \draw[-latex] ($(v\x)+#2*(d\x)-#3*(f\x)$) -- ($(v\x)+#2*(d\x)$);
      \draw[fill,black,tdplot_screen_coords] ($(v\x)+#2*(d\x)$) circle (0.003);
    };
  }}
}

\newcommand\quadVertices{%
  \coordinate (v1) at ({-0.3618}, {-0.1648}, {-0.0816});
  \coordinate (v2) at ({-0.2618}, {-0.1512}, {0.1826});
  \coordinate (v3) at ({-0.4000}, {0.1155}, {-0.0816});
  \coordinate (v4) at ({-0.3000}, {0.1291}, {0.1826});
  \coordinate (v5) at ({-0.1000}, {-0.1291}, {-0.1826});
  \coordinate (v6) at ({0.0000}, {-0.1155}, {0.0816});
  \coordinate (v7) at ({-0.1382}, {0.1512}, {-0.1826});
  \coordinate (v8) at ({-0.0382}, {0.1648}, {0.0816});
  \coordinate (v9) at ({-0.1382}, {-0.2225}, {0.0816});
  \coordinate (v10) at ({-0.2000}, {-0.2309}, {-0.0816});
  \coordinate (v11) at ({-0.2000}, {0.2309}, {0.0816});
  \coordinate (v12) at ({-0.2618}, {0.2225}, {-0.0816});
  \coordinate (v13) at ({-0.2000}, {0.0000}, {-0.2449});
  \coordinate (v14) at ({-0.0382}, {0.0221}, {0.1826});
  \coordinate (v15) at ({-0.3618}, {-0.0221}, {-0.1826});
  \coordinate (v16) at ({-0.2000}, {0.0000}, {0.2449});
  \coordinate (v17) at ({-0.4236}, {0.0577}, {0.0816});
  \coordinate (v18) at ({-0.4000}, {-0.1155}, {0.0816});
  \coordinate (v19) at ({0.0000}, {0.1155}, {-0.0816});
  \coordinate (v20) at ({0.0236}, {-0.0577}, {-0.0816});
  \coordinate (v21) at ({-0.1618}, {-0.5112}, {-0.0816});
  \coordinate (v22) at ({-0.0618}, {-0.4976}, {0.1826});
  \coordinate (v23) at ({-0.1000}, {-0.2173}, {0.1826});
  \coordinate (v24) at ({0.1000}, {-0.4755}, {-0.1826});
  \coordinate (v25) at ({0.2000}, {-0.4619}, {0.0816});
  \coordinate (v26) at ({0.0618}, {-0.1953}, {-0.1826});
  \coordinate (v27) at ({0.1618}, {-0.1816}, {0.0816});
  \coordinate (v28) at ({0.0618}, {-0.5689}, {0.0816});
  \coordinate (v29) at ({0.0000}, {-0.5774}, {-0.0816});
  \coordinate (v30) at ({-0.0618}, {-0.1239}, {-0.0816});
  \coordinate (v31) at ({0.0000}, {-0.3464}, {-0.2449});
  \coordinate (v32) at ({0.1618}, {-0.3244}, {0.1826});
  \coordinate (v33) at ({-0.1618}, {-0.3685}, {-0.1826});
  \coordinate (v34) at ({0.0000}, {-0.3464}, {0.2449});
  \coordinate (v35) at ({-0.2236}, {-0.2887}, {0.0816});
  \coordinate (v36) at ({-0.2000}, {-0.4619}, {0.0816});
  \coordinate (v37) at ({0.2000}, {-0.2309}, {-0.0816});
  \coordinate (v38) at ({0.2236}, {-0.4041}, {-0.0816});
  \coordinate (v39) at ({-0.1618}, {0.1816}, {-0.0816});
  \coordinate (v40) at ({-0.0618}, {0.1953}, {0.1826});
  \coordinate (v41) at ({-0.2000}, {0.4619}, {-0.0816});
  \coordinate (v42) at ({-0.1000}, {0.4755}, {0.1826});
  \coordinate (v43) at ({0.1000}, {0.2173}, {-0.1826});
  \coordinate (v44) at ({0.2000}, {0.2309}, {0.0816});
  \coordinate (v45) at ({0.0618}, {0.4976}, {-0.1826});
  \coordinate (v46) at ({0.1618}, {0.5112}, {0.0816});
  \coordinate (v47) at ({0.0618}, {0.1239}, {0.0816});
  \coordinate (v48) at ({0.0000}, {0.5774}, {0.0816});
  \coordinate (v49) at ({-0.0618}, {0.5689}, {-0.0816});
  \coordinate (v50) at ({0.0000}, {0.3464}, {-0.2449});
  \coordinate (v51) at ({0.1618}, {0.3685}, {0.1826});
  \coordinate (v52) at ({-0.1618}, {0.3244}, {-0.1826});
  \coordinate (v53) at ({0.0000}, {0.3464}, {0.2449});
  \coordinate (v54) at ({-0.2236}, {0.4041}, {0.0816});
  \coordinate (v55) at ({0.2000}, {0.4619}, {-0.0816});
  \coordinate (v56) at ({0.2236}, {0.2887}, {-0.0816});
  \coordinate (v57) at ({0.0382}, {-0.1648}, {-0.0816});
  \coordinate (v58) at ({0.1382}, {-0.1512}, {0.1826});
  \coordinate (v59) at ({0.1000}, {0.1291}, {0.1826});
  \coordinate (v60) at ({0.3000}, {-0.1291}, {-0.1826});
  \coordinate (v61) at ({0.4000}, {-0.1155}, {0.0816});
  \coordinate (v62) at ({0.2618}, {0.1512}, {-0.1826});
  \coordinate (v63) at ({0.3618}, {0.1648}, {0.0816});
  \coordinate (v64) at ({0.2618}, {-0.2225}, {0.0816});
  \coordinate (v65) at ({0.1382}, {0.2225}, {-0.0816});
  \coordinate (v66) at ({0.2000}, {0.0000}, {-0.2449});
  \coordinate (v67) at ({0.3618}, {0.0221}, {0.1826});
  \coordinate (v68) at ({0.0382}, {-0.0221}, {-0.1826});
  \coordinate (v69) at ({0.2000}, {0.0000}, {0.2449});
  \coordinate (v70) at ({-0.0236}, {0.0577}, {0.0816});
  \coordinate (v71) at ({0.4000}, {0.1155}, {-0.0816});
  \coordinate (v72) at ({0.4236}, {-0.0577}, {-0.0816});
}
\newcommand\quadEdges{%
  {1/10},%
  {1/15},%
  {1/18},%
  {2/9},%
  {2/16},%
  {2/18},%
  {3/12},%
  {3/17},%
  {3/15},%
  {4/11},%
  {4/16},%
  {4/17},%
  {5/10},%
  {5/13},%
  {5/20},%
  {6/9},%
  {6/14},%
  {6/20},%
  {7/12},%
  {7/13},%
  {7/19},%
  {8/11},%
  {8/19},%
  {8/14},%
  {9/10},%
  {11/12},%
  {13/15},%
  {14/16},%
  {17/18},%
  {19/20},%
  {21/29},%
  {21/33},%
  {21/36},%
  {22/28},%
  {22/34},%
  {22/36},%
  {10/30},%
  {10/35},%
  {10/33},%
  {23/6},%
  {23/34},%
  {23/35},%
  {24/29},%
  {24/31},%
  {24/38},%
  {25/28},%
  {25/32},%
  {25/38},%
  {26/30},%
  {26/31},%
  {26/37},%
  {27/6},%
  {27/37},%
  {27/32},%
  {28/29},%
  {6/30},%
  {31/33},%
  {32/34},%
  {35/36},%
  {37/38},%
  {39/19},%
  {39/52},%
  {39/11},%
  {40/47},%
  {40/53},%
  {40/11},%
  {41/49},%
  {41/54},%
  {41/52},%
  {42/48},%
  {42/53},%
  {42/54},%
  {43/19},%
  {43/50},%
  {43/56},%
  {44/47},%
  {44/51},%
  {44/56},%
  {45/49},%
  {45/50},%
  {45/55},%
  {46/48},%
  {46/55},%
  {46/51},%
  {47/19},%
  {48/49},%
  {50/52},%
  {51/53},%
  {54/11},%
  {55/56},%
  {57/37},%
  {57/68},%
  {57/6},%
  {58/64},%
  {58/69},%
  {58/6},%
  {19/65},%
  {19/70},%
  {19/68},%
  {59/44},%
  {59/69},%
  {59/70},%
  {60/37},%
  {60/66},%
  {60/72},%
  {61/64},%
  {61/67},%
  {61/72},%
  {62/65},%
  {62/66},%
  {62/71},%
  {63/44},%
  {63/71},%
  {63/67},%
  {64/37},%
  {44/65},%
  {66/68},%
  {67/69},%
  {70/6},%
  {71/72}%
}
\newcommand\quadMaxDisDisplacements{%
  \coordinate (d1) at ({0.0029}, {-0.0156}, {-0.0289});
  \coordinate (d2) at ({0.0084}, {-0.0076}, {-0.0264});
  \coordinate (d3) at ({-0.0017}, {-0.0135}, {-0.0257});
  \coordinate (d4) at ({-0.0031}, {-0.0162}, {-0.0303});
  \coordinate (d5) at ({-0.0081}, {-0.0117}, {-0.0317});
  \coordinate (d6) at ({0.0013}, {0.0136}, {-0.0505});
  \coordinate (d7) at ({-0.0118}, {-0.0023}, {-0.0381});
  \coordinate (d8) at ({0.0057}, {0.0040}, {-0.0466});
  \coordinate (d9) at ({0.0111}, {0.0009}, {-0.0291});
  \coordinate (d10) at ({0.0028}, {-0.0158}, {-0.0251});
  \coordinate (d11) at ({-0.0017}, {-0.0142}, {-0.0270});
  \coordinate (d12) at ({-0.0055}, {-0.0085}, {-0.0259});
  \coordinate (d13) at ({-0.0060}, {-0.0086}, {-0.0287});
  \coordinate (d14) at ({0.0045}, {0.0078}, {-0.0414});
  \coordinate (d15) at ({-0.0043}, {-0.0137}, {-0.0263});
  \coordinate (d16) at ({0.0103}, {-0.0063}, {-0.0314});
  \coordinate (d17) at ({-0.0031}, {-0.0195}, {-0.0280});
  \coordinate (d18) at ({0.0064}, {-0.0183}, {-0.0273});
  \coordinate (d19) at ({0.0013}, {0.0136}, {-0.0505});
  \coordinate (d20) at ({-0.0057}, {0.0126}, {-0.0519});
  \coordinate (d21) at ({0.0019}, {-0.0195}, {0.0855});
  \coordinate (d22) at ({-0.0017}, {0.0693}, {0.0854});
  \coordinate (d23) at ({-0.0104}, {0.0553}, {-0.0200});
  \coordinate (d24) at ({0.0002}, {-0.0592}, {0.0763});
  \coordinate (d25) at ({0.0007}, {0.0408}, {0.0661});
  \coordinate (d26) at ({-0.0054}, {-0.0523}, {-0.0355});
  \coordinate (d27) at ({0.0069}, {0.0272}, {-0.0375});
  \coordinate (d28) at ({0.0022}, {0.0388}, {0.1117});
  \coordinate (d29) at ({0.0011}, {-0.0214}, {0.1153});
  \coordinate (d30) at ({0.0035}, {-0.0168}, {-0.0498});
  \coordinate (d31) at ({0.0029}, {-0.0812}, {0.0263});
  \coordinate (d32) at ({-0.0006}, {0.0710}, {0.0245});
  \coordinate (d33) at ({0.0017}, {-0.0576}, {0.0315});
  \coordinate (d34) at ({0.0005}, {0.0897}, {0.0338});
  \coordinate (d35) at ({-0.0049}, {0.0311}, {-0.0096});
  \coordinate (d36) at ({0.0002}, {0.0318}, {0.0696});
  \coordinate (d37) at ({-0.0017}, {-0.0142}, {-0.0270});
  \coordinate (d38) at ({-0.0013}, {-0.0142}, {0.0464});
  \coordinate (d39) at ({0.0069}, {0.0272}, {-0.0375});
  \coordinate (d40) at ({-0.0054}, {-0.0523}, {-0.0355});
  \coordinate (d41) at ({0.0007}, {0.0408}, {0.0661});
  \coordinate (d42) at ({0.0002}, {-0.0592}, {0.0763});
  \coordinate (d43) at ({-0.0104}, {0.0553}, {-0.0200});
  \coordinate (d44) at ({0.0028}, {-0.0158}, {-0.0251});
  \coordinate (d45) at ({-0.0017}, {0.0693}, {0.0854});
  \coordinate (d46) at ({0.0019}, {-0.0195}, {0.0855});
  \coordinate (d47) at ({0.0035}, {-0.0168}, {-0.0498});
  \coordinate (d48) at ({0.0011}, {-0.0214}, {0.1153});
  \coordinate (d49) at ({0.0022}, {0.0388}, {0.1117});
  \coordinate (d50) at ({0.0005}, {0.0897}, {0.0338});
  \coordinate (d51) at ({0.0017}, {-0.0576}, {0.0315});
  \coordinate (d52) at ({-0.0006}, {0.0710}, {0.0245});
  \coordinate (d53) at ({0.0029}, {-0.0812}, {0.0263});
  \coordinate (d54) at ({-0.0013}, {-0.0142}, {0.0464});
  \coordinate (d55) at ({0.0002}, {0.0318}, {0.0696});
  \coordinate (d56) at ({-0.0049}, {0.0311}, {-0.0096});
  \coordinate (d57) at ({0.0057}, {0.0040}, {-0.0466});
  \coordinate (d58) at ({-0.0118}, {-0.0023}, {-0.0381});
  \coordinate (d59) at ({-0.0081}, {-0.0117}, {-0.0317});
  \coordinate (d60) at ({-0.0031}, {-0.0162}, {-0.0303});
  \coordinate (d61) at ({-0.0017}, {-0.0135}, {-0.0257});
  \coordinate (d62) at ({0.0084}, {-0.0076}, {-0.0264});
  \coordinate (d63) at ({0.0029}, {-0.0156}, {-0.0289});
  \coordinate (d64) at ({-0.0055}, {-0.0085}, {-0.0259});
  \coordinate (d65) at ({0.0111}, {0.0009}, {-0.0291});
  \coordinate (d66) at ({0.0103}, {-0.0063}, {-0.0314});
  \coordinate (d67) at ({-0.0043}, {-0.0137}, {-0.0263});
  \coordinate (d68) at ({0.0045}, {0.0078}, {-0.0414});
  \coordinate (d69) at ({-0.0060}, {-0.0086}, {-0.0287});
  \coordinate (d70) at ({-0.0057}, {0.0126}, {-0.0519});
  \coordinate (d71) at ({0.0064}, {-0.0183}, {-0.0273});
  \coordinate (d72) at ({-0.0031}, {-0.0195}, {-0.0280});
}
\newcommand\quadMaxDisForces{%
  \coordinate (f1) at ({0.0055}, {-0.0299}, {-0.0554});
  \coordinate (f2) at ({0.0161}, {-0.0146}, {-0.0505});
  \coordinate (f3) at ({-0.0033}, {-0.0258}, {-0.0491});
  \coordinate (f4) at ({-0.0060}, {-0.0310}, {-0.0581});
  \coordinate (f5) at ({-0.0155}, {-0.0224}, {-0.0608});
  \coordinate (f6) at ({0.0025}, {0.0260}, {-0.0967});
  \coordinate (f7) at ({-0.0227}, {-0.0044}, {-0.0729});
  \coordinate (f8) at ({0.0109}, {0.0077}, {-0.0892});
  \coordinate (f9) at ({0.0213}, {0.0018}, {-0.0557});
  \coordinate (f10) at ({0.0054}, {-0.0303}, {-0.0480});
  \coordinate (f11) at ({-0.0033}, {-0.0272}, {-0.0516});
  \coordinate (f12) at ({-0.0105}, {-0.0164}, {-0.0495});
  \coordinate (f13) at ({-0.0114}, {-0.0164}, {-0.0550});
  \coordinate (f14) at ({0.0087}, {0.0148}, {-0.0792});
  \coordinate (f15) at ({-0.0082}, {-0.0263}, {-0.0503});
  \coordinate (f16) at ({0.0197}, {-0.0121}, {-0.0602});
  \coordinate (f17) at ({-0.0059}, {-0.0374}, {-0.0536});
  \coordinate (f18) at ({0.0122}, {-0.0350}, {-0.0523});
  \coordinate (f19) at ({0.0025}, {0.0260}, {-0.0967});
  \coordinate (f20) at ({-0.0109}, {0.0241}, {-0.0993});
  \coordinate (f21) at ({0.0036}, {-0.0373}, {0.1637});
  \coordinate (f22) at ({-0.0032}, {0.1327}, {0.1636});
  \coordinate (f23) at ({-0.0199}, {0.1059}, {-0.0383});
  \coordinate (f24) at ({0.0004}, {-0.1133}, {0.1461});
  \coordinate (f25) at ({0.0014}, {0.0781}, {0.1266});
  \coordinate (f26) at ({-0.0103}, {-0.1002}, {-0.0680});
  \coordinate (f27) at ({0.0131}, {0.0521}, {-0.0717});
  \coordinate (f28) at ({0.0042}, {0.0744}, {0.2139});
  \coordinate (f29) at ({0.0021}, {-0.0410}, {0.2207});
  \coordinate (f30) at ({0.0067}, {-0.0321}, {-0.0953});
  \coordinate (f31) at ({0.0056}, {-0.1555}, {0.0504});
  \coordinate (f32) at ({-0.0011}, {0.1359}, {0.0468});
  \coordinate (f33) at ({0.0033}, {-0.1104}, {0.0603});
  \coordinate (f34) at ({0.0009}, {0.1718}, {0.0648});
  \coordinate (f35) at ({-0.0095}, {0.0596}, {-0.0183});
  \coordinate (f36) at ({0.0004}, {0.0609}, {0.1333});
  \coordinate (f37) at ({-0.0033}, {-0.0272}, {-0.0516});
  \coordinate (f38) at ({-0.0025}, {-0.0271}, {0.0888});
  \coordinate (f39) at ({0.0131}, {0.0521}, {-0.0717});
  \coordinate (f40) at ({-0.0103}, {-0.1002}, {-0.0680});
  \coordinate (f41) at ({0.0014}, {0.0781}, {0.1266});
  \coordinate (f42) at ({0.0004}, {-0.1133}, {0.1461});
  \coordinate (f43) at ({-0.0199}, {0.1059}, {-0.0383});
  \coordinate (f44) at ({0.0054}, {-0.0303}, {-0.0480});
  \coordinate (f45) at ({-0.0032}, {0.1327}, {0.1636});
  \coordinate (f46) at ({0.0036}, {-0.0373}, {0.1637});
  \coordinate (f47) at ({0.0067}, {-0.0321}, {-0.0953});
  \coordinate (f48) at ({0.0021}, {-0.0410}, {0.2207});
  \coordinate (f49) at ({0.0042}, {0.0744}, {0.2139});
  \coordinate (f50) at ({0.0009}, {0.1718}, {0.0648});
  \coordinate (f51) at ({0.0033}, {-0.1104}, {0.0603});
  \coordinate (f52) at ({-0.0011}, {0.1359}, {0.0468});
  \coordinate (f53) at ({0.0056}, {-0.1555}, {0.0504});
  \coordinate (f54) at ({-0.0025}, {-0.0271}, {0.0888});
  \coordinate (f55) at ({0.0004}, {0.0609}, {0.1333});
  \coordinate (f56) at ({-0.0095}, {0.0596}, {-0.0183});
  \coordinate (f57) at ({0.0109}, {0.0077}, {-0.0892});
  \coordinate (f58) at ({-0.0227}, {-0.0044}, {-0.0729});
  \coordinate (f59) at ({-0.0155}, {-0.0224}, {-0.0608});
  \coordinate (f60) at ({-0.0060}, {-0.0310}, {-0.0581});
  \coordinate (f61) at ({-0.0033}, {-0.0258}, {-0.0491});
  \coordinate (f62) at ({0.0161}, {-0.0146}, {-0.0505});
  \coordinate (f63) at ({0.0055}, {-0.0299}, {-0.0554});
  \coordinate (f64) at ({-0.0105}, {-0.0164}, {-0.0495});
  \coordinate (f65) at ({0.0213}, {0.0018}, {-0.0557});
  \coordinate (f66) at ({0.0197}, {-0.0121}, {-0.0602});
  \coordinate (f67) at ({-0.0082}, {-0.0263}, {-0.0503});
  \coordinate (f68) at ({0.0087}, {0.0148}, {-0.0792});
  \coordinate (f69) at ({-0.0114}, {-0.0164}, {-0.0550});
  \coordinate (f70) at ({-0.0109}, {0.0241}, {-0.0993});
  \coordinate (f71) at ({0.0122}, {-0.0350}, {-0.0523});
  \coordinate (f72) at ({-0.0059}, {-0.0374}, {-0.0536});
}

\newcommand\quadMaxAxLDisplacements{%
  \coordinate (d1) at ({0.0023}, {-0.0067}, {-0.0260});
  \coordinate (d2) at ({-0.0004}, {-0.0107}, {-0.0232});
  \coordinate (d3) at ({-0.0032}, {-0.0066}, {-0.0175});
  \coordinate (d4) at ({-0.0080}, {-0.0157}, {-0.0165});
  \coordinate (d5) at ({-0.0046}, {-0.0034}, {-0.0241});
  \coordinate (d6) at ({0.0045}, {0.0073}, {-0.0327});
  \coordinate (d7) at ({-0.0060}, {0.0010}, {-0.0174});
  \coordinate (d8) at ({-0.0035}, {-0.0010}, {-0.0184});
  \coordinate (d9) at ({0.0090}, {0.0015}, {-0.0203});
  \coordinate (d10) at ({0.0031}, {-0.0048}, {-0.0178});
  \coordinate (d11) at ({-0.0075}, {-0.0108}, {-0.0109});
  \coordinate (d12) at ({-0.0049}, {-0.0044}, {-0.0122});
  \coordinate (d13) at ({-0.0022}, {0.0005}, {-0.0199});
  \coordinate (d14) at ({-0.0026}, {-0.0031}, {-0.0213});
  \coordinate (d15) at ({-0.0022}, {-0.0035}, {-0.0214});
  \coordinate (d16) at ({-0.0009}, {-0.0119}, {-0.0200});
  \coordinate (d17) at ({-0.0058}, {-0.0139}, {-0.0205});
  \coordinate (d18) at ({-0.0000}, {-0.0131}, {-0.0246});
  \coordinate (d19) at ({-0.0024}, {0.0064}, {-0.0204});
  \coordinate (d20) at ({-0.0022}, {0.0064}, {-0.0340});
  \coordinate (d21) at ({0.0061}, {-0.0040}, {0.0574});
  \coordinate (d22) at ({0.0156}, {0.0545}, {0.0531});
  \coordinate (d23) at ({0.0010}, {0.0358}, {-0.0075});
  \coordinate (d24) at ({0.0012}, {-0.0311}, {0.0398});
  \coordinate (d25) at ({0.0147}, {0.0334}, {0.0299});
  \coordinate (d26) at ({-0.0031}, {-0.0277}, {-0.0277});
  \coordinate (d27) at ({0.0113}, {0.0241}, {-0.0322});
  \coordinate (d28) at ({0.0140}, {0.0343}, {0.0653});
  \coordinate (d29) at ({0.0057}, {-0.0051}, {0.0705});
  \coordinate (d30) at ({0.0050}, {-0.0072}, {-0.0322});
  \coordinate (d31) at ({0.0009}, {-0.0453}, {0.0110});
  \coordinate (d32) at ({0.0186}, {0.0515}, {0.0066});
  \coordinate (d33) at ({0.0019}, {-0.0316}, {0.0183});
  \coordinate (d34) at ({0.0219}, {0.0655}, {0.0203});
  \coordinate (d35) at ({0.0037}, {0.0287}, {-0.0058});
  \coordinate (d36) at ({0.0125}, {0.0299}, {0.0486});
  \coordinate (d37) at ({0.0017}, {-0.0038}, {-0.0260});
  \coordinate (d38) at ({0.0051}, {-0.0034}, {0.0155});
  \coordinate (d39) at ({-0.0016}, {0.0084}, {-0.0153});
  \coordinate (d40) at ({-0.0112}, {-0.0293}, {-0.0124});
  \coordinate (d41) at ({-0.0024}, {0.0128}, {0.0304});
  \coordinate (d42) at ({-0.0085}, {-0.0321}, {0.0374});
  \coordinate (d43) at ({-0.0028}, {0.0221}, {-0.0049});
  \coordinate (d44) at ({-0.0042}, {-0.0137}, {-0.0008});
  \coordinate (d45) at ({0.0000}, {0.0242}, {0.0445});
  \coordinate (d46) at ({-0.0049}, {-0.0174}, {0.0478});
  \coordinate (d47) at ({-0.0057}, {-0.0117}, {-0.0182});
  \coordinate (d48) at ({-0.0047}, {-0.0168}, {0.0565});
  \coordinate (d49) at ({-0.0005}, {0.0104}, {0.0535});
  \coordinate (d50) at ({-0.0002}, {0.0352}, {0.0180});
  \coordinate (d51) at ({-0.0083}, {-0.0336}, {0.0248});
  \coordinate (d52) at ({-0.0017}, {0.0271}, {0.0112});
  \coordinate (d53) at ({-0.0098}, {-0.0424}, {0.0180});
  \coordinate (d54) at ({-0.0076}, {-0.0108}, {0.0213});
  \coordinate (d55) at ({-0.0019}, {0.0070}, {0.0411});
  \coordinate (d56) at ({-0.0034}, {0.0068}, {0.0066});
  \coordinate (d57) at ({0.0046}, {0.0031}, {-0.0314});
  \coordinate (d58) at ({-0.0043}, {-0.0089}, {-0.0265});
  \coordinate (d59) at ({-0.0073}, {-0.0151}, {-0.0053});
  \coordinate (d60) at ({-0.0006}, {0.0011}, {-0.0233});
  \coordinate (d61) at ({-0.0004}, {-0.0092}, {-0.0174});
  \coordinate (d62) at ({0.0016}, {0.0043}, {-0.0051});
  \coordinate (d63) at ({-0.0033}, {-0.0114}, {-0.0043});
  \coordinate (d64) at ({-0.0018}, {-0.0073}, {-0.0245});
  \coordinate (d65) at ({0.0013}, {0.0016}, {-0.0036});
  \coordinate (d66) at ({0.0041}, {0.0080}, {-0.0165});
  \coordinate (d67) at ({-0.0032}, {-0.0154}, {-0.0100});
  \coordinate (d68) at ({0.0015}, {0.0093}, {-0.0228});
  \coordinate (d69) at ({-0.0038}, {-0.0153}, {-0.0114});
  \coordinate (d70) at ({-0.0064}, {0.0058}, {-0.0212});
  \coordinate (d71) at ({-0.0002}, {-0.0044}, {-0.0057});
  \coordinate (d72) at ({-0.0029}, {-0.0048}, {-0.0162});
}
\newcommand\quadMaxAxLForces{%
  \coordinate (f1) at ({-0.0126}, {0.0034}, {-0.1023});
  \coordinate (f2) at ({-0.0451}, {-0.0722}, {-0.0831});
  \coordinate (f3) at ({-0.0132}, {-0.0037}, {-0.0358});
  \coordinate (f4) at ({-0.0389}, {-0.0741}, {-0.0179});
  \coordinate (f5) at ({0.0037}, {0.0194}, {-0.0809});
  \coordinate (f6) at ({-0.0196}, {-0.0636}, {-0.0437});
  \coordinate (f7) at ({-0.0026}, {0.0142}, {0.0000});
  \coordinate (f8) at ({-0.0302}, {-0.0597}, {0.0154});
  \coordinate (f9) at ({-0.0322}, {-0.0473}, {-0.0849});
  \coordinate (f10) at ({-0.0080}, {0.0147}, {-0.0972});
  \coordinate (f11) at ({-0.0273}, {-0.0526}, {0.0153});
  \coordinate (f12) at ({-0.0101}, {-0.0077}, {0.0064});
  \coordinate (f13) at ({0.0030}, {0.0332}, {-0.0513});
  \coordinate (f14) at ({-0.0402}, {-0.0857}, {-0.0214});
  \coordinate (f15) at ({-0.0037}, {0.0251}, {-0.0715});
  \coordinate (f16) at ({-0.0463}, {-0.0900}, {-0.0388});
  \coordinate (f17) at ({-0.0294}, {-0.0425}, {-0.0519});
  \coordinate (f18) at ({-0.0305}, {-0.0427}, {-0.0926});
  \coordinate (f19) at ({-0.0170}, {-0.0216}, {0.0070});
  \coordinate (f20) at ({-0.0157}, {-0.0214}, {-0.0282});
  \coordinate (f21) at ({0.0161}, {0.0489}, {0.1355});
  \coordinate (f22) at ({0.0933}, {0.2404}, {0.1054});
  \coordinate (f23) at ({0.2810}, {0.3036}, {-0.2800});
  \coordinate (f24) at ({0.0015}, {-0.0415}, {0.0658});
  \coordinate (f25) at ({0.0936}, {0.0706}, {0.0260});
  \coordinate (f26) at ({-0.0002}, {-0.0385}, {-0.0604});
  \coordinate (f27) at ({0.0342}, {0.0680}, {-0.0809});
  \coordinate (f28) at ({0.0504}, {0.1263}, {0.1334});
  \coordinate (f29) at ({0.0108}, {0.0359}, {0.1531});
  \coordinate (f30) at ({0.0147}, {-0.0145}, {-0.0592});
  \coordinate (f31) at ({0.0085}, {-0.0674}, {0.0010});
  \coordinate (f32) at ({0.1673}, {0.1178}, {-0.0105});
  \coordinate (f33) at ({0.0041}, {-0.0500}, {-0.0045});
  \coordinate (f34) at ({0.1265}, {0.2928}, {-0.0545});
  \coordinate (f35) at ({0.1332}, {0.1864}, {-0.0162});
  \coordinate (f36) at ({0.0739}, {0.1783}, {0.1100});
  \coordinate (f37) at ({0.0143}, {0.0026}, {-0.0658});
  \coordinate (f38) at ({0.0331}, {0.0051}, {-0.0059});
  \coordinate (f39) at ({-0.0173}, {-0.0224}, {0.0085});
  \coordinate (f40) at ({-0.0348}, {-0.0705}, {0.0192});
  \coordinate (f41) at ({-0.0261}, {-0.0231}, {0.0590});
  \coordinate (f42) at ({-0.0425}, {-0.0686}, {0.0690});
  \coordinate (f43) at ({-0.0149}, {0.0021}, {0.0330});
  \coordinate (f44) at ({-0.0342}, {-0.0419}, {0.0392});
  \coordinate (f45) at ({-0.0217}, {0.0028}, {0.0814});
  \coordinate (f46) at ({-0.0381}, {-0.0440}, {0.0909});
  \coordinate (f47) at ({-0.0281}, {-0.0497}, {0.0127});
  \coordinate (f48) at ({-0.0396}, {-0.0476}, {0.0931});
  \coordinate (f49) at ({-0.0292}, {-0.0191}, {0.0877});
  \coordinate (f50) at ({-0.0127}, {0.0121}, {0.0501});
  \coordinate (f51) at ({-0.0422}, {-0.0626}, {0.0645});
  \coordinate (f52) at ({-0.0162}, {-0.0032}, {0.0356});
  \coordinate (f53) at ({-0.0452}, {-0.0794}, {0.0509});
  \coordinate (f54) at ({-0.0334}, {-0.0535}, {0.0472});
  \coordinate (f55) at ({-0.0268}, {-0.0110}, {0.0835});
  \coordinate (f56) at ({-0.0231}, {-0.0105}, {0.0519});
  \coordinate (f57) at ({0.0049}, {-0.0205}, {-0.0510});
  \coordinate (f58) at ({-0.0139}, {-0.0614}, {-0.0507});
  \coordinate (f59) at ({-0.0358}, {-0.0719}, {0.0074});
  \coordinate (f60) at ({0.0081}, {0.0258}, {-0.0485});
  \coordinate (f61) at ({-0.0005}, {-0.0153}, {-0.0311});
  \coordinate (f62) at ({-0.0186}, {0.0216}, {0.0202});
  \coordinate (f63) at ({-0.0275}, {-0.0256}, {0.0277});
  \coordinate (f64) at ({0.0026}, {-0.0193}, {-0.0602});
  \coordinate (f65) at ({-0.0256}, {-0.0105}, {0.0344});
  \coordinate (f66) at ({-0.0063}, {0.0310}, {-0.0147});
  \coordinate (f67) at ({-0.0215}, {-0.0446}, {0.0009});
  \coordinate (f68) at ({-0.0043}, {0.0016}, {-0.0198});
  \coordinate (f69) at ({-0.0234}, {-0.0726}, {-0.0140});
  \coordinate (f70) at ({-0.0263}, {-0.0645}, {-0.0095});
  \coordinate (f71) at ({-0.0195}, {0.0121}, {0.0182});
  \coordinate (f72) at ({-0.0069}, {0.0138}, {-0.0217});
}

\newcommand\quadEdgesWithMaxAxLoads{%
  {1/10/t/12},%
  {1/15/t/17},%
  {1/18/c/4},%
  {2/9/c/9},%
  {2/16/c/10},%
  {2/18/t/26},%
  {3/12/t/13},%
  {3/17/t/1},%
  {3/15/t/4},%
  {4/11/c/11},%
  {4/16/c/8},%
  {4/17/c/3},%
  {5/10/t/77},%
  {5/13/t/15},%
  {5/20/t/19},%
  {6/9/t/2},%
  {6/14/c/18},%
  {6/20/c/17},%
  {7/12/t/19},%
  {7/13/t/40},%
  {7/19/c/1},%
  {8/11/c/14},%
  {8/19/t/12},%
  {8/14/c/18},%
  {9/10/c/32},%
  {11/12/t/28},%
  {13/15/c/6},%
  {14/16/t/33},%
  {17/18/c/2},%
  {19/20/t/35},%
  {21/29/c/4},%
  {21/33/t/42},%
  {21/36/c/9},%
  {22/28/c/2},%
  {22/34/c/20},%
  {22/36/t/52},%
  {10/30/c/2},%
  {10/35/t/100},%
  {10/33/t/2},%
  {23/6/c/100},%
  {23/34/c/17},%
  {23/35/c/14},%
  {24/29/c/2},%
  {24/31/t/7},%
  {24/38/c/6},%
  {25/28/t/5},%
  {25/32/c/6},%
  {25/38/t/46},%
  {26/30/t/18},%
  {26/31/t/33},%
  {26/37/c/7},%
  {27/6/c/24},%
  {27/37/c/8},%
  {27/32/c/20},%
  {28/29/c/3},%
  {6/30/t/45},%
  {31/33/c/4},%
  {32/34/t/54},%
  {35/36/c/19},%
  {37/38/t/24},%
  {39/19/t/45},%
  {39/52/t/4},%
  {39/11/c/2},%
  {40/47/t/5},%
  {40/53/c/8},%
  {40/11/t/10},%
  {41/49/c/3},%
  {41/54/c/8},%
  {41/52/t/20},%
  {42/48/c/0},%
  {42/53/t/2},%
  {42/54/t/17},%
  {43/19/t/7},%
  {43/50/t/7},%
  {43/56/t/8},%
  {44/47/t/33},%
  {44/51/c/3},%
  {44/56/c/14},%
  {45/49/c/2},%
  {45/50/t/6},%
  {45/55/c/5},%
  {46/48/c/0},%
  {46/55/t/15},%
  {46/51/c/10},%
  {47/19/t/9},%
  {48/49/t/2},%
  {50/52/c/4},%
  {51/53/t/15},%
  {54/11/c/9},%
  {55/56/c/1},%
  {57/37/t/55},%
  {57/68/t/25},%
  {57/6/c/6},%
  {58/64/c/8},%
  {58/69/c/21},%
  {58/6/c/12},%
  {19/65/t/16},%
  {19/70/c/16},%
  {19/68/t/56},%
  {59/44/c/32},%
  {59/69/t/8},%
  {59/70/c/10},%
  {60/37/t/21},%
  {60/66/t/32},%
  {60/72/t/22},%
  {61/64/c/13},%
  {61/67/t/8},%
  {61/72/t/7},%
  {62/65/t/20},%
  {62/66/t/6},%
  {62/71/c/5},%
  {63/44/t/7},%
  {63/71/c/0},%
  {63/67/c/6},%
  {64/37/c/11},%
  {44/65/t/41},%
  {66/68/c/5},%
  {67/69/t/11},%
  {70/6/c/52},%
  {71/72/t/1}%
}

\newcommand\ctrlAuthPlot{%
\pgfplotsset{%
  set layers,
      scale only axis,
      height=0.5\linewidth,
}
\begin{axis}[%
xmode=log,
  xlabel=$s$,
  xtick pos=left,
  axis y line*=left,
  ylabel=Thrust (\SI{}{\newton})
]
\addplot[magenta,thick,dash pattern=on 3mm off 0.5mm] coordinates {%
  (0.100,89.649) (0.112,89.479) (0.126,89.268) (0.141,89.008) (0.158,88.690) (0.178,88.303) (0.200,87.831) (0.224,87.241) (0.251,86.535) (0.282,85.776) (0.316,84.860) (0.355,83.754) (0.398,82.749) (0.447,81.558) (0.501,80.353) (0.562,79.113) (0.631,77.914) (0.708,76.794) (0.794,75.717) (0.891,74.840) (1.000,73.967) (1.122,73.134) (1.259,72.241) (1.413,72.141) (1.585,71.566) (1.778,70.911) (1.995,70.542) (2.239,70.221) (2.512,70.208) (2.818,70.063) (3.162,70.086) (3.548,70.322) (3.981,70.280) (4.467,70.216) (5.012,70.150) (5.623,70.160) (6.310,70.136) (7.079,70.040) (7.943,70.728) (8.913,69.540) (10.000,69.547)
};\label{fig:control:ctrl_authority:th}
\addplot[magenta,thick,densely dashed] coordinates {%
  (0.100,99.113) (0.112,98.925) (0.126,98.692) (0.141,98.405) (0.158,98.053) (0.178,97.625) (0.200,97.104) (0.224,96.451) (0.251,95.671) (0.282,94.832) (0.316,93.818) (0.355,92.597) (0.398,91.485) (0.447,90.168) (0.501,88.837) (0.562,87.465) (0.631,86.140) (0.708,84.901) (0.794,83.710) (0.891,82.741) (1.000,81.776) (1.122,80.855) (1.259,79.867) (1.413,79.758) (1.585,79.122) (1.778,78.398) (1.995,77.990) (2.239,77.635) (2.512,77.620) (2.818,77.460) (3.162,77.485) (3.548,77.746) (3.981,77.700) (4.467,77.629) (5.012,77.556) (5.623,77.567) (6.310,77.541) (7.079,77.434) (7.943,78.195) (8.913,76.882) (10.000,76.890)
};\label{fig:control:ctrl_authority:thx}
\addplot[magenta,thick,dashed] coordinates {%
  (0.100,106.097) (0.112,105.896) (0.126,105.646) (0.141,105.339) (0.158,104.963) (0.178,104.505) (0.200,103.947) (0.224,103.247) (0.251,102.413) (0.282,101.514) (0.316,100.430) (0.355,99.121) (0.398,97.932) (0.447,96.522) (0.501,95.097) (0.562,93.629) (0.631,92.209) (0.708,90.884) (0.794,89.609) (0.891,88.572) (1.000,87.538) (1.122,86.552) (1.259,85.495) (1.413,85.378) (1.585,84.697) (1.778,83.922) (1.995,83.485) (2.239,83.106) (2.512,83.090) (2.818,82.919) (3.162,82.945) (3.548,83.225) (3.981,83.175) (4.467,83.099) (5.012,83.021) (5.623,83.033) (6.310,83.005) (7.079,82.891) (7.943,83.705) (8.913,82.300) (10.000,82.308)
};\label{fig:control:ctrl_authority:thy}
\addplot[magenta,thick,loosely dashed] coordinates {%
  (0.100,219.593) (0.112,219.177) (0.126,218.660) (0.141,218.024) (0.158,217.245) (0.178,216.297) (0.200,215.142) (0.224,213.695) (0.251,211.968) (0.282,210.108) (0.316,207.863) (0.355,205.156) (0.398,202.693) (0.447,199.776) (0.501,196.825) (0.562,193.787) (0.631,190.849) (0.708,188.106) (0.794,185.467) (0.891,183.320) (1.000,181.181) (1.122,179.140) (1.259,176.953) (1.413,176.710) (1.585,175.301) (1.778,173.696) (1.995,172.793) (2.239,172.007) (2.512,171.974) (2.818,171.620) (3.162,171.675) (3.548,172.253) (3.981,172.151) (4.467,171.993) (5.012,171.831) (5.623,171.857) (6.310,171.798) (7.079,171.562) (7.943,173.247) (8.913,170.338) (10.000,170.355)
};\label{fig:control:ctrl_authority:thz}

    \end{axis}

    \begin{axis}[
      axis y line*=right,
      axis x line=none,
      ylabel=Torque (\SI{}{\newton\meter}),
      legend columns=4,
      legend style={%
        font=\mystrut,
        legend cell align=left,
        at={(0.5,1.03)},anchor=south
      },
    ]

    \addlegendimage{legend image with text=$xyz$}
    \addlegendentry{};
\addlegendimage{legend image with text=$x$}
\addlegendentry{};
\addlegendimage{legend image with text=$y$}
\addlegendentry{};
\addlegendimage{legend image with text=$z$}
\addlegendentry{};

\addlegendimage{/pgfplots/refstyle=fig:control:ctrl_authority:th}\addlegendentry{};
\addlegendimage{/pgfplots/refstyle=fig:control:ctrl_authority:thx}\addlegendentry{};
\addlegendimage{/pgfplots/refstyle=fig:control:ctrl_authority:thy}\addlegendentry{};
\addlegendimage{/pgfplots/refstyle=fig:control:ctrl_authority:thz}\addlegendentry{$T$};
\addplot[teal,thick] coordinates {%
  (0.100,42.961) (0.112,43.005) (0.126,43.047) (0.141,43.110) (0.158,43.189) (0.178,43.288) (0.200,43.399) (0.224,43.475) (0.251,43.592) (0.282,43.847) (0.316,44.085) (0.355,44.292) (0.398,44.747) (0.447,45.139) (0.501,45.592) (0.562,46.063) (0.631,46.559) (0.708,47.068) (0.794,47.552) (0.891,48.036) (1.000,48.461) (1.122,48.836) (1.259,49.152) (1.413,49.502) (1.585,49.747) (1.778,49.946) (1.995,50.128) (2.239,50.278) (2.512,50.411) (2.818,50.513) (3.162,50.599) (3.548,50.671) (3.981,50.723) (4.467,50.765) (5.012,50.798) (5.623,50.825) (6.310,50.846) (7.079,50.863) (7.943,50.878) (8.913,50.886) (10.000,50.895)
};\label{fig:control:ctrl_authority:to}
\addplot[teal,thick,densely dotted] coordinates {%
  (0.100,49.324) (0.112,49.377) (0.126,49.429) (0.141,49.505) (0.158,49.601) (0.178,49.721) (0.200,49.857) (0.224,49.952) (0.251,50.099) (0.282,50.412) (0.316,50.696) (0.355,50.957) (0.398,51.504) (0.447,51.977) (0.501,52.529) (0.562,53.098) (0.631,53.704) (0.708,54.318) (0.794,54.908) (0.891,55.485) (1.000,55.992) (1.122,56.488) (1.259,56.882) (1.413,57.294) (1.585,57.597) (1.778,57.849) (1.995,58.081) (2.239,58.266) (2.512,58.420) (2.818,58.548) (3.162,58.650) (3.548,58.723) (3.981,58.788) (4.467,58.837) (5.012,58.877) (5.623,58.907) (6.310,58.932) (7.079,58.958) (7.943,58.969) (8.913,59.016) (10.000,59.004)
};\label{fig:control:ctrl_authority:tox}
\addplot[teal,thick,dotted] coordinates {%
  (0.100,56.804) (0.112,56.865) (0.126,56.925) (0.141,57.013) (0.158,57.123) (0.178,57.261) (0.200,57.418) (0.224,57.527) (0.251,57.697) (0.282,58.057) (0.316,58.384) (0.355,58.685) (0.398,59.315) (0.447,59.859) (0.501,60.495) (0.562,61.150) (0.631,61.849) (0.708,62.555) (0.794,63.235) (0.891,63.775) (1.000,62.633) (1.122,61.573) (1.259,60.769) (1.413,59.961) (1.585,59.389) (1.778,58.924) (1.995,58.508) (2.239,58.181) (2.512,57.914) (2.818,57.695) (3.162,57.522) (3.548,57.400) (3.981,57.292) (4.467,57.211) (5.012,57.143) (5.623,57.095) (6.310,57.053) (7.079,57.011) (7.943,56.992) (8.913,56.916) (10.000,56.935)
};\label{fig:control:ctrl_authority:toy}
\addplot[teal,thick,loosely dotted] coordinates {%
  (0.100,87.446) (0.112,87.520) (0.126,87.589) (0.141,87.694) (0.158,87.826) (0.178,87.992) (0.200,88.173) (0.224,88.284) (0.251,88.451) (0.282,88.862) (0.316,89.286) (0.355,89.581) (0.398,90.376) (0.447,91.045) (0.501,91.801) (0.562,92.608) (0.631,93.424) (0.708,94.299) (0.794,95.108) (0.891,95.984) (1.000,96.749) (1.122,97.180) (1.259,97.662) (1.413,98.321) (1.585,98.714) (1.778,98.995) (1.995,99.250) (2.239,99.492) (2.512,99.751) (2.818,99.908) (3.162,100.064) (3.548,100.260) (3.981,100.342) (4.467,100.421) (5.012,100.475) (5.623,100.537) (6.310,100.573) (7.079,100.576) (7.943,100.643) (8.913,100.465) (10.000,100.594)
};\label{fig:control:ctrl_authority:toz}

\addlegendimage{/pgfplots/refstyle=fig:control:ctrl_authority:to}\addlegendentry{};
\addlegendimage{/pgfplots/refstyle=fig:control:ctrl_authority:tox}\addlegendentry{};
\addlegendimage{/pgfplots/refstyle=fig:control:ctrl_authority:toy}\addlegendentry{};
\addlegendimage{/pgfplots/refstyle=fig:control:ctrl_authority:toz}\addlegendentry{$\tau$};

\end{axis}
}

\newcommand\prAthVertices{%
\coordinate (V1) at (49.5809,121.2598,73.0175);
\coordinate (V2) at (80.2236,-103.5682,73.0175);
\coordinate (V3) at (-0.0000,-0.0000,219.4851);
\coordinate (V4) at (0.0000,0.0000,-219.9177);
\coordinate (V5) at (-80.2236,103.5682,-73.4501);
\coordinate (V6) at (-49.5809,-121.2598,-73.4501);
\coordinate (V7) at (-129.8045,-17.6916,73.0175);
\coordinate (V8) at (129.8045,17.6916,-73.4501);
}
\newcommand\prAthFaces{%
1/{6,2,8,4},%
2/{3,7,5,1},%
3/{4,8,1,5},%
4/{2,6,7,3},%
5/{7,6,4,5},%
6/{2,3,1,8}%
}

\newcommand\prBthVertices{%
\coordinate (V1) at (66.1907,-85.4518,60.2451);
\coordinate (V2) at (40.9081,100.0487,60.2451);
\coordinate (V3) at (107.0988,14.5969,-60.6020);
\coordinate (V4) at (-107.0988,-14.5969,60.2451);
\coordinate (V5) at (-40.9081,-100.0487,-60.6020);
\coordinate (V6) at (-66.1907,85.4518,-60.6020);
\coordinate (V7) at (-0.0000,0.0000,-181.4492);
\coordinate (V8) at (-0.0000,0.0000,181.0922);
}
\newcommand\prBthFaces{%
1/{5,1,3,7},%
2/{8,4,6,2},%
3/{7,3,2,6},%
4/{1,5,4,8},%
5/{4,5,7,6},%
6/{1,8,2,3}%
}

\newcommand\prCthVertices{%
\coordinate (V1) at (100.6893,13.7233,-56.9752);
\coordinate (V2) at (-0.0000,-0.0000,170.2546);
\coordinate (V3) at (62.2294,-80.3379,56.6397);
\coordinate (V4) at (-62.2294,80.3379,-56.9752);
\coordinate (V5) at (-0.0000,-0.0000,-170.5901);
\coordinate (V6) at (-100.6894,-13.7234,56.6397);
\coordinate (V7) at (-38.4599,-94.0612,-56.9752);
\coordinate (V8) at (38.4599,94.0612,56.6397);
}
\newcommand\prCthFaces{%
1/{7,3,1,5},%
2/{2,6,4,8},%
3/{5,1,8,4},%
4/{3,7,6,2},%
5/{6,7,5,4},%
6/{3,2,8,1}%
}

\newcommand\prAtoVertices{%
\coordinate (V1) at (-65.7645,0.3017,-29.1487);
\coordinate (V2) at (-32.6210,57.1046,29.1487);
\coordinate (V3) at (0.0000,0.0000,-87.4462);
\coordinate (V4) at (33.1435,56.8029,-29.1487);
\coordinate (V5) at (-0.0000,0.0000,87.4462);
\coordinate (V6) at (32.6210,-57.1046,-29.1487);
\coordinate (V7) at (65.7645,-0.3017,29.1487);
\coordinate (V8) at (-33.1435,-56.8029,29.1487);
}
\newcommand\prAtoFaces{%
1/{7,5,2,4},%
2/{6,7,4,3},%
3/{6,8,5,7},%
4/{8,6,3,1},%
5/{5,8,1,2},%
6/{1,3,4,2}%
}

\newcommand\prCtoVertices{%
\coordinate (V1) at (-43.8431,-50.0286,10.7090);
\coordinate (V2) at (-49.1855,-46.6786,17.0998);
\coordinate (V3) at (-7.9842,-13.6837,86.9804);
\coordinate (V4) at (-0.0000,0.0000,100.4654);
\coordinate (V5) at (-65.2476,12.9550,-10.7090);
\coordinate (V6) at (-7.8583,13.7564,86.9804);
\coordinate (V7) at (-15.8321,65.9352,17.0998);
\coordinate (V8) at (7.8583,-13.7564,-86.9804);
\coordinate (V9) at (0.0000,-0.0000,-100.4654);
\coordinate (V10) at (-15.8425,0.0727,-86.9804);
\coordinate (V11) at (-65.0176,19.2566,-17.0998);
\coordinate (V12) at (-21.4044,62.9836,10.7090);
\coordinate (V13) at (7.9842,13.6837,-86.9804);
\coordinate (V14) at (65.2476,-12.9550,10.7090);
\coordinate (V15) at (49.1855,46.6786,-17.0998);
\coordinate (V16) at (21.4044,-62.9836,-10.7090);
\coordinate (V17) at (65.0176,-19.2566,17.0998);
\coordinate (V18) at (15.8425,-0.0727,86.9804);
\coordinate (V19) at (43.8431,50.0286,-10.7090);
\coordinate (V20) at (15.8321,-65.9352,-17.0998);
}
\newcommand\prCtoFaces{%
1/{13,15,19,7,12},%
2/{18,4,6,7,19},%
3/{14,17,18,19,15},%
4/{9,8,14,15,13},%
5/{20,16,17,14,8},%
6/{16,3,4,18,17},%
7/{16,20,1,2,3},%
8/{1,20,8,9,10},%
9/{2,1,10,11,5},%
10/{3,2,5,6,4},%
11/{6,5,11,12,7},%
12/{11,10,9,13,12}%
}

\newcommand\prBtoVertices{%
\coordinate (V1) at (53.9419,35.9153,-3.8267);
\coordinate (V2) at (67.0986,-13.2914,19.2971);
\coordinate (V3) at (58.0745,-28.7574,3.8267);
\coordinate (V4) at (32.6510,-0.1498,68.5392);
\coordinate (V5) at (16.4552,28.2017,-68.5392);
\coordinate (V6) at (0.0000,0.0000,-96.7487);
\coordinate (V7) at (16.1958,-28.3515,-68.5392);
\coordinate (V8) at (22.0387,-64.7548,-19.2971);
\coordinate (V9) at (4.1326,-64.6727,-3.8267);
\coordinate (V10) at (-32.6510,0.1498,-68.5392);
\coordinate (V11) at (-4.1326,64.6727,3.8267);
\coordinate (V12) at (-22.0387,64.7548,19.2971);
\coordinate (V13) at (-58.0745,28.7574,-3.8267);
\coordinate (V14) at (-67.0986,13.2914,-19.2971);
\coordinate (V15) at (-53.9419,-35.9153,3.8267);
\coordinate (V16) at (-45.0600,-51.4634,19.2971);
\coordinate (V17) at (-16.1958,28.3515,68.5392);
\coordinate (V18) at (0.0000,-0.0000,96.7487);
\coordinate (V19) at (-16.4552,-28.2017,68.5392);
\coordinate (V20) at (45.0600,51.4634,-19.2971);
}
\newcommand\prBtoFaces{%
1/{5,20,11},%
2/{20,1,4,18,17,12,11},%
3/{2,4,1},%
4/{6,7,3,2,1,20,5},%
5/{8,3,7},%
6/{8,9,19,18,4,2,3},%
7/{9,16,19},%
8/{9,8,7,6,10,15,16},%
9/{15,10,14},%
10/{19,16,15,14,13,17,18},%
11/{17,13,12},%
12/{13,14,10,6,5,11,12}%
}

\newcommand\pMaxthVertices{%
\coordinate (V1) at (0.0000,0.0000,-221.4797);
\coordinate (V2) at (130.7908,17.8260,-73.8992);
\coordinate (V3) at (-49.9576,-122.1812,-73.8992);
\coordinate (V4) at (80.8332,-104.3552,73.6813);
\coordinate (V5) at (-80.8332,104.3552,-73.8992);
\coordinate (V6) at (49.9576,122.1812,73.6813);
\coordinate (V7) at (-130.7908,-17.8260,73.6813);
\coordinate (V8) at (0.0000,0.0000,221.2618);
}
\newcommand\pMaxthFaces{%
1/{8,7,5,6},%
2/{1,2,6,5},%
3/{7,3,1,5},%
4/{4,8,6,2},%
5/{3,4,2,1},%
6/{4,3,7,8}%
}

\newcommand\pMaxtoVertices{%
\coordinate (V1) at (-74.3846,-12.9761,-12.2181);
\coordinate (V2) at (-49.8517,-55.9221,34.9674);
\coordinate (V3) at (-50.4876,-27.9611,-34.9674);
\coordinate (V4) at (-25.9547,-70.9071,12.2181);
\coordinate (V5) at (-1.0288,-28.1879,-82.1530);
\coordinate (V6) at (23.5041,-71.1339,-34.9674);
\coordinate (V7) at (-24.9259,-42.7191,57.7168);
\coordinate (V8) at (-1.0288,-57.7040,34.9674);
\coordinate (V9) at (48.4300,-57.9309,-12.2181);
\coordinate (V10) at (-49.4588,29.7430,34.9674);
\coordinate (V11) at (-24.9259,-13.2030,82.1530);
\coordinate (V12) at (-24.5329,42.9460,57.7168);
\coordinate (V13) at (0.0000,0.0000,104.9023);
\coordinate (V14) at (23.8971,-14.9849,82.1530);
\coordinate (V15) at (49.4588,-0.2269,57.7168);
\coordinate (V16) at (73.3558,-15.2118,34.9674);
\coordinate (V17) at (-73.3558,15.2118,-34.9674);
\coordinate (V18) at (-49.4588,0.2269,-57.7168);
\coordinate (V19) at (-23.8971,14.9849,-82.1530);
\coordinate (V20) at (0.0000,0.0000,-104.9023);
\coordinate (V21) at (24.5329,-42.9460,-57.7168);
\coordinate (V22) at (24.9259,13.2030,-82.1530);
\coordinate (V23) at (49.4588,-29.7430,-34.9674);
\coordinate (V24) at (-48.4300,57.9309,12.2181);
\coordinate (V25) at (1.0288,57.7040,-34.9674);
\coordinate (V26) at (24.9259,42.7191,-57.7168);
\coordinate (V27) at (-23.5041,71.1339,34.9674);
\coordinate (V28) at (1.0288,28.1879,82.1530);
\coordinate (V29) at (25.9547,70.9071,-12.2181);
\coordinate (V30) at (50.4876,27.9611,34.9674);
\coordinate (V31) at (49.8517,55.9221,-34.9674);
\coordinate (V32) at (74.3846,12.9761,12.2181);
}
\newcommand\pMaxtoFaces{%
1/{12,10,24,27},%
2/{1,17,24,10},%
3/{28,13,12,27},%
4/{15,13,28,30},%
5/{30,28,27,29},%
6/{32,30,29,31},%
7/{27,24,25,29},%
8/{1,3,18,17},%
9/{18,20,19,17},%
10/{17,19,25,24},%
11/{20,22,31,26},%
12/{19,20,26,25},%
13/{26,31,29,25},%
14/{16,15,30,32},%
15/{13,11,10,12},%
16/{11,2,1,10},%
17/{6,21,20,5},%
18/{3,5,20,18},%
19/{23,32,31,22},%
20/{21,23,22,20},%
21/{4,6,5,3},%
22/{2,4,3,1},%
23/{7,2,11,13},%
24/{8,4,2,7},%
25/{16,14,13,15},%
26/{8,7,13,14},%
27/{4,8,9,6},%
28/{9,8,14,16},%
29/{6,9,23,21},%
30/{23,9,16,32}%
}

\newcommand\plotCtrlSet[3][100]{%
  #2;
  \foreach \n\f in #3{%
    \draw[fill,opacity=0.5] \foreach\v[count=\i] in \f {%
      \ifnum\i=1\relax\else--\fi%
        (V\v)
    } -- cycle;
  }
  \draw[->] (xyz cs:x=0) -- (xyz cs:x=#1) node[above] {$x$};
  \draw[->] (xyz cs:y=0) -- (xyz cs:y=#1) node[right] {$y$};
  \draw[->] (xyz cs:z=0) -- (xyz cs:z=#1) node[above] {$z$};
}

\usepackage{tikz, tikz-3dplot}

\theoremstyle{definition}
\newtheorem{definition}{Definition}[section]
\newtheorem{program}{Program}[section]
\theoremstyle{plain}
\newtheorem{fact}{Fact}[section]
\theoremstyle{remark}
\newtheorem*{remark}{Remark}

\def\mystrut{\vphantom{hg}}

\renewcommand\d{\mathop{}\!\mathrm{d}}
\newcommand\e{\mathop{}\!\mathrm{e}}

\let\norm\undefined 
\DeclarePairedDelimiter\abs{\lvert}{\rvert}%
\DeclarePairedDelimiter\norm{\lVert}{\rVert}%

\newcommand{\nset}[2]{\llbracket#1;#2\rrbracket}%

\newcommand{\nulls}[1]{\mathrm{null}\left(#1\right)}%

\newcommand{\range}[1]{\ensuremath{\mathrm{range}\left(#1\right)}}%

\makeatletter
\newcommand{\minimize}[2]{%
  \begin{array}{ll}
    \text{minimize} & #1 \\
    \text{subject to} & #2 \checknextarg%
  }

  \newcommand{\maximize}[2]{%
    \begin{array}{ll}
      \text{maximize} & #1 \\
      \text{subject to} & #2 \checknextarg%
    }

    \newcommand{\checknextarg}{%
    \@ifnextchar\bgroup{\gobblenextarg}{\end{array}}%
  }
\newcommand{\gobblenextarg}[1]{\\ & #1\kernel@ifnextchar\bgroup{\gobblenextarg}{\end{array}}}
\makeatother

\crefname{definition}{definition}{definitions}%
\Crefname{paragraph}{Paragraph}{Paragraphs}%
\crefname{program}{program}{programs}%
\crefname{fact}{fact}{facts}%
\crefformat{enumi}{#2\textup{(#1)}#3}%

\def\Mtt{\ensuremath{{\bm{M}^{\mathrm{tt}}}}} 
\def\uh{\ensuremath{\bm{u}^{\mathrm{h}}}} 
\def\th{\ensuremath{\bm{t}^{\mathrm{h}}}} 

\def\njt{\ensuremath{n^{\mathrm{jt}}}}%
\def\nmb{\ensuremath{n^{\mathrm{mb}}}}%

\NewDocumentCommand{\bp}{o}{%
  \ensuremath{%
    \chi_{%
      \IfNoValueTF{#1}{\bm{x}}{{#1}}%
    }%
    ^{\mathrm{p}}
  }
}%

\NewDocumentCommand{\bpo}{d<>o}{%
  \ensuremath{%
    \chi_{%
      \IfNoValueTF{#1}{\bm{x}}{{#1}}%
      \IfNoValueTF{#2}{\eta}{{#2}}%
    }
    ^{\mathrm{po}}
  }
}%

\NewDocumentCommand{\bpd}{d<>o}{%
  \ensuremath{%
    \chi_{%
      \IfNoValueTF{#1}{\bm{x}}{{#1}}%
      \IfNoValueTF{#2}{\epsilon}{{#2}}%
    }
    ^{\mathrm{pd}}
  }
}%

\NewDocumentCommand{\bj}{o}{%
  \ensuremath{%
    \chi_{%
      \IfNoValueTF{#1}{j}{{#1}}%
    }
    ^{\mathrm{jt}}
  }
}%

\pgfplotsset{%
  compat=1.18,
  width=0.7\linewidth,
  legend image with text/.style={%
    legend image code/.code={%
      \node[anchor=center] at (0.3cm,0cm) {#1};
    }
  },
}

\title{The Dodecacopter: a Versatile Multirotor System of Dodecahedron-Shaped Modules}
\author{K\'evin Garanger$^{1}$, Thanakorn Khamvilai$^{2}$, Jeremy Epps$^{3}$, Eric Feron$^{4}$%
  \thanks{$^{1}$K\'{e}vin Garanger is affiliated with the Department of Mechanical and Aerospace Engineering at the University of California, Irvine.
  $^{2}$Thanakorn Khamvilai is affiliated with the Department of Mechanical Engineering at Texas Tech University. $^{3}$Jeremy Epps is affiliated with optimAero LLC.
    $^{4}$Eric Feron is affiliated with the Computer, Electrical, and Mathematical Sciences and Engineering division at King Abdullah University of Science and Technology (KAUST).
    Support for this project comes from the KAUST baseline faculty fund and a gift from General Atomics.
}%
}
\date{}

\begin{document}

\maketitle

\begin{abstract}
  With the promise of greater safety and adaptability, modular reconfigurable uncrewed air vehicles have been proposed as unique, versatile platforms holding the potential to replace multiple types of monolithic vehicles at once.
  State-of-the-art rigidly assembled modular vehicles are generally two-dimensional configurations in which the rotors are coplanar and assume the shape of a ``flight array''.
  We introduce the Dodecacopter, a new type of modular rotorcraft where all modules take the shape of a regular dodecahedron, allowing the creation of richer sets of configurations beyond flight arrays.
  In particular, we show how the chosen module design can be used to create three-dimensional and fully actuated configurations.
  We justify the relevance of these types of configurations in terms of their structural and actuation properties with various performance indicators.
  Given the broad range of configurations and capabilities that can be achieved with our proposed design, we formulate tractable optimization programs to find optimal configurations given structural and actuation constraints.
  Finally, a prototype of such a vehicle is presented along with results of performed flights in multiple configurations.
\end{abstract}

\section{Introduction}

Modular robotic systems employ a group of compatible autonomous components, which, through interactions between components, hardware redundancy, and reconfiguration capabilities, promise to offer increased versatility and robustness over other robotic systems~\cite{seo2019modular}.
These promises and the rapid development of uncrewed air system (UAS) technologies have triggered investigations regarding the feasibility of autonomous modular aerial vehicles~\cite{oung2011distributed,naldi2011class,duffy2015lift,saldana2018modquad,mu2019universal,gandhi2020self,garanger2020modeling,kulkarni2020reconfigurable,schiano2022reconfigurable,davis2024modular,kobayashi2024modular,ren2024modeling}.
Such vehicles would benefit from the advantages of modularity for robotic systems and from the ability of aerial vehicles to perform missions in otherwise hard-to-reach places.

For rotorcraft systems, the most straightforward advantage of modularity is the ability to scale the number of rotors to the weight of a payload~\cite{mu2019universal,schiano2022reconfigurable,davis2024modular,kobayashi2024modular}.
Modular systems could allow UAS operators to use a unified system that can be configured to adapt to specific payload and mission requirements, instead of a large fleet of different vehicles, simplifying their operations.
Another advantage of systems made of self-contained modules is redundancy of hardware, which, especially when combined with overactuation, improves robustness and tolerance to failures such as those of a rotor or flight computer~\cite{ren2024modeling}.

A potential yet so far unexplored benefit of rigid reconfigurable modular UAS is their ability to be configured in fully actuated or omnidirectional aerial vehicles by changing their module orientations~\cite{michieletto2018fundamental}.
Fully actuated (or six degrees of freedom) aerial vehicles can independently change their thrust and torque around hovering conditions in all directions~\cite{rajappa2015modeling} and omnidirectional aerial vehicles are fully actuated aerial vehicles that are able to hover in any attitude~\cite{brescianini2016design}.
Fully actuated aerial vehicles can perform translational motion and position stabilization without the need for tilting their airframe, only by changing the direction of their thrust vector.
This feature makes them suitable for carrying payloads that need to remain at a constant orientation~\cite{mehmood2016maneuverability} and for achieving a more reactive and efficient position tracking in adverse weather conditions than their underactuated counterparts~\cite{tadokoro2017maneuverability}.
The ability of fully actuated aerial vehicles to remain at a fixed position while compensating for external torque and force disturbances also makes them good candidates for operating robotic manipulators~\cite{rajappa2015modeling,ollero2021past}.
As for omnidirectional aerial vehicles, they can track arbitrary pose trajectories,
which presents an advantage for three-dimensional mapping, sensing, and object manipulation~\cite{brescianini2016design,park2016design,kamel2018voliro,nguyen2018novel,allenspach2020design}.
Vehicles with tilted rotors may also produce arbitrary moments via differential thrust to control their attitude.
In contrast, vehicles with parallel rotors rely on counteracting torque form propeller drag to create yawing moments, which leads to vanishing control authority for many-rotor vehicles~\cite{gabrich2020modquad}.


Almost all prior works on rigid modular rotorcraft UAS rely on vehicles that form coplanar configurations, that is with vehicles assembling in a horizontal plane~\cite{oung2011distributed,duffy2015lift,saldana2018modquad,gandhi2020self,kulkarni2020reconfigurable}.
To the authors' knowledge, three-dimensional configurations of modular rotorcraft have seldom been explored, except for the work in~\cite{naldi2011class}, which presents a class of ducted fan modules able to assemble horizontally but also vertically, thereby creating coaxial rotorcraft systems.
Despite the fact that three-dimensional configurations may incur more pronounced rotor wake interactions, reducing their overall efficiency, they also benefit from some advantages.
These advantages are increased rigidity, more compact designs, and, if modules exhibit the appropriate symmetries, the ability to adjust their relative orientation to create fully actuated or omnidirectional vehicles.
With a careful positioning of the rotors in a three-dimensional configuration, effects of wake interactions can also be mitigated~\cite{garanger2020modeling,epps2020wake}.

In the work presented here, a rotorcraft module concept that exploits all the aforementioned advantages, named Dodecacopter, is introduced.
The essential components of this module are its frame, in the shape of a regular dodecahedron, and its propulsion system, consisting of a fixed-pitch propeller whose rotation axis is fixed with respect to the frame.
The choice of the regular dodecahedron shape for the module frame is made for two reasons.
The first reason is the high number of symmetries of this shape, and the second one is the fact that it enables connections between modules that facilitate the creation of diverse vehicle configurations.
Both these points are detailed in the next section.
Based on the Dodecacopter concept, three main contributions to the field of modular aerial vehicles are provided.
The first contribution is the module design itself with the methodology for connecting modules together, which result in a wide variety of feasible vehicle configurations spanning standard multirotor configurations, three-dimensional vehicles, and fully-actuated vehicles.
The second contribution is a set of methods for determining optimal configurations given various constraints encompassing structural and control properties.
These methods depend directly on the introduced connection methodology between modules, as it results in the modules configurations being representable with discrete variables, allowing the use of mixed-integer programming (MIP).
Convex programming also plays a significant part in these methods.
Finally, the third contribution is an experimental demonstration of the viability of the Dodecacopter concept, based on a prototype and various flight tests.


A description of the Dodecacopter concept and its specific connection methodology is introduced in \cref{sec:modular_vehicles}.
This section also explicitly describes the feasible configurations in mathematical terms.
An emphasis is placed on the versatility offered by the regular dodecahedron shape.
The problem of allocating control inputs for overactuated systems of many modules is then studied in \cref{sec:control}.
Different objectives for optimal control allocation are introduced and compared, such as power consumption or control authority.
Since the Dodecacopter is intended as a module for creating three-dimensional vehicles with enhanced stiffness, a structural characterization of modular vehicles is formulated in \cref{sec:structures}.
Building upon the different metrics introduced for implementing control allocation strategies and structurally characterizing modular vehicles, MIP programs are formulated to describe feasible configurations and some of their properties in \cref{sec:configuration_optimization}.
Properties of modular configurations are shown to be well represented by linear and second-order conic constraints, thereby allowing the use of specialized solvers to find optimal configurations.
Finally, in \cref{sec:prototype}, a prototype of the Dodecacopter is presented and results of flight tests in different configurations are given.

\section*{Notations}

$\mathbb{R}$, $\mathbb{Z}$, and $\mathbb{N}$ are respectively the real numbers, the integers, and the nonnegative integers.
For $i$ and $j$ some integers with $i \leq j$, $\nset{i}{j}$ is the set of integers from $i$ to $j$.
Vectors are denoted by lowercase boldface letters and matrices by uppercase boldface letters.
For $m$ and $n$ some integers, $\bm{0}_m$ is used to denote the zero vector of size $m$, $\bm{0}_{m \times n}$ the zero matrix of size $m \times n$, and $\bm{I}_m$ the identity matrix of size $m\times m$.
Size subscripts are ignored in the absence of ambiguity.
For a matrix $\bm{M}$, the notation $\range{\bm{M}}$ corresponds to its column space and $\nulls{\bm{M}}$ to its null space.
For $F$ a subspace of an Euclidean space $E$, the orthogonal complement of $F$ in $E$ is written $F^\perp$.

\section{Dodecacopter-based modular vehicles}\label{sec:modular_vehicles}

In order to understand the choice of the dodecahedron shape, one can consider an arbitrary module frame represented by a polyhedron whose edges correspond to structural members and whose vertices are points where rigid connections between modules can be made.
The orientation-preserving symmetries of this polyhedron are then rotations that permute the positions of the vertices but leave the shape of the frame unchanged.
Each of these symmetries is therefore a possible choice of orientation for a given module that does not affect how other modules can connect to it.
A large symmetry group is consequently desirable to achieve diverse module orientations while retaining the capability to rigidly connect to adjacent modules.
The finite subgroups of $SO(3)$, the rotations in three dimensions, are well characterized~\cite{bui2020classifying}.
They are either a cyclic group, a dihedral group, the tetrahedral group, the octahedral group, or the icosahedral group.
The cyclic group and dihedral group would result in a frame shape that is a prism (a polyhedron formed by extending a polygon along the direction perpendicular to it).
These shapes have little practical purpose for three-dimensional and fully actuated modular vehicles and their corresponding groups can therefore be eliminated.
Among the three remaining possibilities, the icosahedral group is the largest one with \num{60} rotations, and hence the natural one to choose.
There is an infinite number of polyhedra with icosahedral symmetry, but practical considerations for manufacturing modules dictate to look at all but the simplest cases, for example regular polyhedra.
Among the five regular polyhedra, the two having icosahedral symmetry are the one that gave this symmetry group its name, that is the icosahedron, and its dual, the dodecahedron.
The dodecahedron is selected from these two options since the relative positions between connected modules, induced from the connection mechanism introduced in the rest of this section, results in a rich and easy to characterize set of vehicle configurations.
This connection mechanism is specific to the dodecahedron shape and could not be achieved as easily with icosahedron-shaped modules.

\subsection{Dodecacopter description}

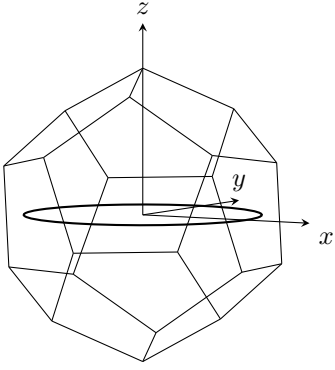
\begin{figure}
  \centering
  \tdplotsetmaincoords{85}{30}
  \begin{tikzpicture}[tdplot_main_coords, scale=8]
    \dodeca[thick];
    \draw [-stealth] (0,0,0) -- (0.32,0,0) node[below right] {$x$};
    \draw [-stealth] (0,0,0) -- (0,0.32,0) node[above] {$y$};
    \draw [-stealth] (0,0,0) -- (0,0,0.32) node[above] {$z$};
  \end{tikzpicture}
  \caption{Representation of a module frame and its propeller.\label{fig:modular:module_plot}}
\end{figure}

Each module of the Dodecacopter system is a self-contained unit with a structural frame in the shape of a regular dodecahedron.
A module uses at its center a fixed-pitch propeller for propulsion which rotates either clockwise (CW) or counter-clockwise (CCW).
The propeller axis intersects two opposed vertices of the dodecahedron frame.
A frame of reference is associated with each module and placed at its center, as represented in \cref{fig:modular:module_plot}.

\subsubsection{The dodecahedron frame}

The geometry of the structural frame of a module is described fully by the vertex coordinates and topology of the regular dodecahedron, one of the five convex regular polyhedra~\cite{coxeter1973regular}.

A regular dodecahedron centered in $\bm{0}_3$, of inradius $1$, and with two opposed vertices on the vertical axis in a frame of reference, is assumed given and denoted by $\mathcal{D}$.
The following sets describe the geometry and topology of $\mathcal{D}$:
\begin{itemize}
  \item $V^\mathcal{D} \subset \mathbb{R}^3$, the vertices of $\mathcal{D}$;
  \item $E^\mathcal{D}$, the edges of $\mathcal{D}$, a set of unordered pairs of $V^\mathcal{D}$;
  \item $F^\mathcal{D}$, the faces of $\mathcal{D}$, a set of unordered quintuples of $V^\mathcal{D}$.
\end{itemize}
For the mathematical description of modular configurations, the Dodecacopter module can be assimilated without loss of generality to $\mathcal{D}$, regardless of its actual size.

\begin{definition}\label{def:modular:dodecahedron_rot_group}
  The rotation group of $\mathcal{D}$ is
  \begin{equation*}
    S^\mathcal{D} := \left\{\bm{R} \in SO(3)\ |\ \bm{R}V^\mathcal{D} = V^\mathcal{D}\right\}.
  \end{equation*}
\end{definition}

\begin{fact}\label{fact:modular:dodecahedron_rot_group_order}
  The order (or cardinality) of $S^\mathcal{D}$ is \num{60}~\cite{coxeter1973regular}.
\end{fact}

\subsubsection{Connection between modules}

To form a rotorcraft vehicle capable of performing stable flight, modules can be combined together in different configurations.
Possible configurations are determined by the method for joining two modules together, which are referred to as ``connections''.
Every possible connection between modules is formally described as a rotation followed by a translation (i.e.\ an orientation-preserving isometry of $\mathbb{R}^3$, or an element of the special euclidean group $SE(3)$), such that this isometry gives the transformation from the frame of reference of a module to the frame of a reference of the connected one.

\begin{definition}
  For $\bm{R} \in SO(3)$ and $\bm{x} \in \mathbb{R}^3$, $f_{\bm{R},\bm{x}} \in SE(3)$ is defined by
  \begin{equation*}
    \forall \bm{y} \in \mathbb{R}^3,\ f_{\bm{R},\bm{x}}(\bm{y}) :=\bm{R}\bm{y} + \bm{x}.
  \end{equation*}
\end{definition}

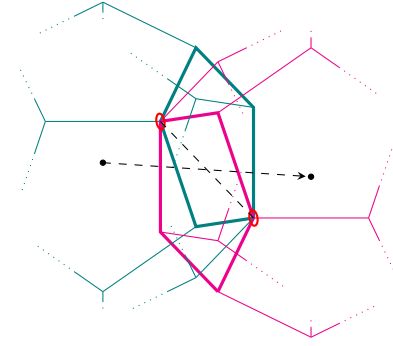
\begin{figure}
  \centering
  \tdplotsetmaincoords{95}{210}
  \begin{tikzpicture}[tdplot_main_coords, scale=3.5, transform shape]
    \getdodecaopm;
  \end{tikzpicture}
  \caption{Connection between modules. Connecting faces are represented with thicker lines.
    Connecting vertices are circled.
  Vertices from one module to the other can be obtained via a translation along the dashed arrow or a rotation of $\pi$ radians around the dashed line.\label{fig:modular:module_connections}}
\end{figure}

The method used for connecting two modules consists of first selecting two vertices per module with each pair of vertices being on a same face but not adjacent.
Then, these pairs of vertices are joined together such that the faces they belong to are in the same plane, but the other vertices of these faces do not coincide.
This somewhat unintuitive connection method is better represented visually, as done in \cref{fig:modular:module_connections}.
Each possible connection to a module can be characterized by a translation equal to the sum of the two chosen vertices for that connection and by a rotation from $S^\mathcal{D}$.

\begin{definition}\label{def:modular:connections}
  For $\bm{R} \in S^\mathcal{D}$, $\bm{g} \in V^\mathcal{D}$, and $\bm{h} \in V^\mathcal{D}$ with $\{\bm{g},\bm{h}\} \notin E^\mathcal{D}$ and $\{\bm{g},\bm{h}\} \subset e$ for some $e$ in $F^\mathcal{D}$, the connection $c_{\bm{R},\{\bm{g},\bm{h}\}}$ is given by
  \begin{equation*}
    c_{\bm{R},\{\bm{g},\bm{h}\}} := f_{\bm{R},\bm{g}+\bm{h}}.
  \end{equation*}
  These connections are grouped in equivalence classes:
  \begin{equation*}
    \mathcal{C}_{\bm{g},\bm{h}} := \left\{c_{\bm{R},\{\bm{g},\bm{h}\}}\ |\ \bm{R} \in S^\mathcal{D}\right\},
  \end{equation*}
  and $\mathcal{C}$ is used to denote the set of all these classes.
\end{definition}


The number of possible connections defined by \cref{def:modular:connections} is equal to \num{3600}, since there are \num{60} rotations in $S^\mathcal{D}$, \num{12} faces in $F^\mathcal{D}$, and \num{5} choices of non-adjacent pairs of vertices per face.
For practical purposes, this number of connections can be reduced by selecting a specific subset of vertices from which they are made, as described in the fact below.
This subset has the advantage of resulting in fewer translation components, which generate a lattice of $\mathbb{R}^3$ under the additive operation.
With this lattice structure, all module positions achievable via successive connections are discrete and hence representable with integer coordinates, enabling mixed-integer programming for optimizing modular configurations.

\begin{figure}
  \centering
  \tdplotsetmaincoords{85}{225}
  \begin{tikzpicture}[tdplot_main_coords, scale=8, transform shape]
    \centering
    \dodeca;
    \dodecaConnections;
  \end{tikzpicture}
  \caption{Restricted set of module connections. Vertices used for connections are circled. Dashed lines join the pairs of vertices used to make a connection at every face and form a cube. Dotted arrows represent the translation vectors induced by each connection.\label{fig:modular:restricted_connections}}
\end{figure}
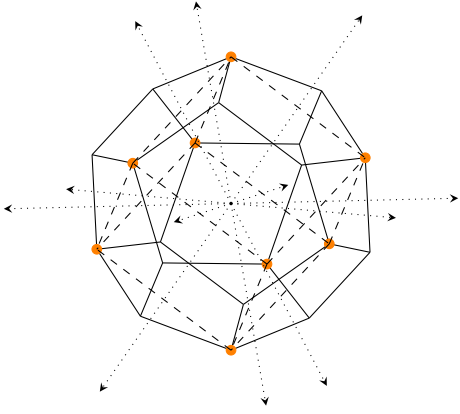

\begin{fact}\label{fact:modular:cube_connections}
  \begin{itemize}
    \item[]
    \item There exists a subset $V^{\mathrm{cube}}$ of $V^\mathcal{D}$ that defines a cube. The set of unordered pairs of $V^{\mathrm{cube}}$ corresponding to the edges of this cube are denoted $E^{\mathrm{cube}}$.
    \item Any subset of $V^\mathcal{D}$ verifying the same condition can be obtained from $V^{\mathrm{cube}}$ with a rotation of $S^\mathcal{D}$.
    \item $\{\bm{g},\bm{h}\} \in E^{\mathrm{cube}}$ implies $\{\bm{g},\bm{h}\} \notin E^\mathcal{D}$ and there is a face $e \in F^\mathcal{D}$ such that $\{\bm{g},\bm{h}\} \subset e$, implying that $\mathcal{C}_{\bm{g},\bm{h}} \in \mathcal{C}$.
  \end{itemize}
\end{fact}
\begin{proof}
  Immediate from looking at the regular dodecahedron vertices (\cref{fig:modular:restricted_connections}).
\end{proof}

A cube as in \cref{fact:modular:cube_connections} is now assumed fixed.

\begin{definition}\label{def:modular:cube_connections}
  The restricted set of connections based on the vertices of $V^{\mathrm{cube}}$ is
  \begin{equation*}
    \bar{\mathcal{C}} := \bigcup_{\{\bm{g},\bm{h}\} \in E^{\mathrm{cube}}} \mathcal{C}_{\bm{g},\bm{h}}
  \end{equation*}
  and the set of translation components from these connections is
  \begin{equation*}
    \bar{\mathcal{C}}^{\mathrm{tr}} := \left\{\bm{g}+\bm{h}\ |\ \{\bm{g},\bm{h}\} \in E^{\mathrm{cube}}\right\}
  \end{equation*}
\end{definition}

The restricted set of connections, as defined in \cref{def:modular:cube_connections} and shown in \cref{fig:modular:restricted_connections}, allows only one possible connection per face for a total of \num{12} different induced relative positions between two connected dodecahedra.
Only \num{8} vertices of the dodecahedron are used for these connections, with each of these vertices involved in \num{3} different possible connections.
These vertices can be chosen arbitrarily as long as they form a cube, but one of the vertices must be the top or bottom one to recreate conventional configurations.

Note that in practice, since modules may be rotated according to all symmetries of the regular dodecahedron, all vertices of their structural frame need to be able to accommodate connections with other modules.
Therefore, the choice of a dodecahedron shape still affords more possible orientation than, for example, a cube that would offer the same possible relative connections.
However, once the orientation of a module is fixed within a vehicle, only a subset of its vertices can be used to form connections.

\subsection{Geometry of Achievable Configurations}

\begin{figure}
  \centering
  \begin{subfigure}[b]{0.49\linewidth}
    \centering
    \tdplotsetmaincoords{125}{230}
    \begin{tikzpicture}[tdplot_main_coords, scale=3.4, transform shape]
      \quadconf;
    \end{tikzpicture}
    \caption{Quadrotor.\label{fig:modular:configuration_examples:quad}}
  \end{subfigure}
  \begin{subfigure}[b]{0.49\linewidth}
    \centering
    \tdplotsetmaincoords{125}{230}
    \begin{tikzpicture}[tdplot_main_coords, scale=3.4, transform shape]
      \hexconf;
    \end{tikzpicture}
    \caption{Hexarotor.\label{fig:modular:configuration_examples:hex}}
  \end{subfigure}
  \caption{%
    Examples of conventional configurations recreated with Dodecacopter modules.%
  \label{fig:modular:configuration_examples}}
\end{figure}
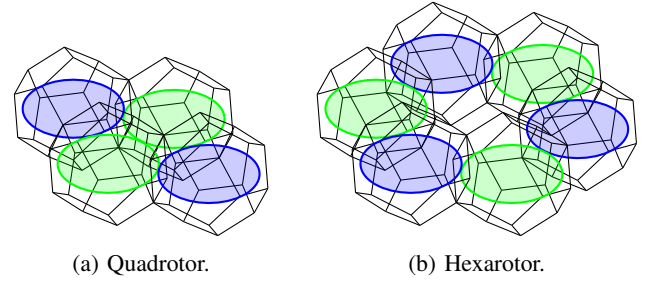

Though perhaps not intuitive, the introduced connection method is particularly useful because it allows a configuration designer to form triangles and tetrahedra with respectively three and four modules, where each pair of modules is connected.
These shapes can be used to form a variety of structurally efficient configurations.
In addition, several common rotor configurations can be recreated by using this connection method, as shown in \cref{fig:modular:configuration_examples}.

\subsubsection{Feasible module positions and orientations}

Modular configurations consist of modules assembled together via the foregoing connection method.
A configuration is fully characterized by the positions of the modules, their orientations, and the spinning direction of their propeller.
Given an initial module, all module positions and orientations are constrained by successive connections.
These successions of connections are characterized by the following isometries, allowing the problems of determining compatible module positions and orientations to be decoupled.

\begin{definition}\label{def:modular:succ_connections}
  Given a finite sequence of $n$ connections $\left(c_{\bm{p}_0,\bm{R}_0},\ldots,c_{\bm{p}_{n-1},\bm{R}_{n-1}}\right)$, the succession of these connections is given by
  \begin{equation*}
    c_{\bm{p}_0,\bm{R}_0} \circ \cdots \circ c_{\bm{p}_{n-1},\bm{R}_{n-1}} := c_{\bm{p}_0+\cdots+\bm{p}_{n-1},\bm{R}_{n-1}\cdots \bm{R}_0}.
  \end{equation*}
  Note that this definition is different from the standard composition of isometries.
  Here, the translations are always done in the original coordinate frame.
\end{definition}

\begin{definition}\label{def:modular:compatible_module}
  A module's position $\bm{p}_1$ and orientation $\bm{R}_1$ are said compatible with respect to another module's position $\bm{p}_2$ and orientation $\bm{R}_2$ if there exists a finite sequence of connections $c_i \in \bar{\mathcal{C}},\ 0 \leq i \leq n$, such that 
  \begin{equation*}
    c_{\bm{p}_1,\bm{R}_1} = c_0 \circ \cdots \circ c_n \circ c_{\bm{p}_2,\bm{R}_2}.
  \end{equation*}
\end{definition}

\begin{definition}\label{def:modular:positions_group}
  The set $\mathcal{P}$ is defined as the additive subgroup of $\mathbb{R}^3$ generated by the translation vectors of all possible connections of $\bar{\mathcal{C}}$:
  \begin{equation*}
    \mathcal{P} := \sum_{\bm{x} \in \bar{\mathcal{C}}^{\mathrm{tr}}} \bm{x} \mathbb{Z}.
  \end{equation*}
\end{definition}

\begin{fact}\label{fact:modular:compatible}
  A module's position $\bm{p}_1$ and orientation $\bm{R}_1$ are compatible with respect to another module's position $\bm{p}_2$ and orientation $\bm{R}_2$ if and only if
  \begin{align*}
    \bm{p}_2 - \bm{p}_1 &\in \mathcal{P} \\
    \bm{R}_2 \bm{R}_1^\top &\in S^\mathcal{D}.
  \end{align*}
\end{fact}
\begin{proof}
  Immediate given \cref{def:modular:connections,def:modular:cube_connections}, and from the group structure of $\mathcal{P}$ and $S^\mathcal{D}$.
\end{proof}

\Cref{fact:modular:compatible} offers a convenient characterization of compatible module positions and orientations.
It also shows the intuitive result that the compatibility condition is transitive, in the sense that one module is compatible with another if and only if the converse is true.
Given that $\mathcal{P}$ is an additive subgroup of $\mathbb{R}^3$, it is either discrete (i.e.\ a lattice) or is dense in a non-trivial subspace of $\mathbb{R}^3$.
The following fact shows that the former applies and this lattice is three-dimensional.

\begin{fact}\label{fact:modular:cube_group_lattice}
  $\mathcal{P}$ is a three-dimensional lattice of $\mathbb{R}^3$ and a basis of it can be found among the vectors of $\bar{\mathcal{C}}^{\mathrm{tr}}$.
\end{fact}
\begin{proof}
  By choosing an appropriate basis of $\mathbb{R}^3$, the coordinates of the vertices of $V^{\mathrm{cube}}$ are given by $\left(\pm \frac12, \pm \frac12, \pm \frac12\right)$.
  In this basis, the vectors in $\left\{\bm{q}+\bm{h}, \{\bm{q},\bm{h}\} \in E^{\mathrm{cube}}\right\}$ have coordinates
  \begin{align*}
    &(\pm 1, \pm 1, 0) \\
    &(\pm 1, 0, \pm 1) \\
    &(0, \pm 1, \pm 1),
  \end{align*}
  From here, it can be seen that $\mathcal{P}$ is generated by $(1,1,0)$, $(1,0,1)$, and $(0,1,1)$, which are three linearly independent vectors, thus forming the basis of a lattice.
\end{proof}

\begin{remark}
  It is easily verified with the basis of \cref{fact:modular:cube_group_lattice} that no two distinct lattice points result in module frames intersecting.
  Therefore, new modules to unoccupied lattice points can always be added without having to worry about potential conflicts between modules.
\end{remark}

\subsubsection{Orientation of module's rotors}

For defining modular configurations, instead of characterizing a module's orientation from elements of $S^\mathcal{D}$, the orientation of its rotors as a unit-norm vector is used.
This representation is more compact since each feasible orientation of a module's rotor can be equivalently achieved by three different orientations of the module.

\begin{definition}\label{def:modular:rotor_orientation}
  Feasible rotor orientations are given by
  \begin{equation*}
    \mathcal{H}
    :=
    \left\{
      \frac{\bm{u}}{\norm{\bm{u}}_2}
      \ |\ \bm{u} \in V^{\mathcal{D}}
    \right\}.
  \end{equation*}
\end{definition}

It is obvious from \cref{def:modular:dodecahedron_rot_group,def:modular:rotor_orientation} that
\begin{equation*}
  \mathcal{H}
  =
  \left\{
    \bm{R}
    \begin{bmatrix}
      0 & 0 & 1
    \end{bmatrix}^\top
    \ |\ \bm{R} S^\mathcal{D}
  \right\},
\end{equation*}
which shows that $\mathcal{H}$ does indeed represent feasible rotor orientations.

\begin{figure}
  \centering
  \begin{subfigure}[t]{0.32\linewidth}
    \centering
    \tdplotsetmaincoords{95}{100}
    \begin{tikzpicture}[tdplot_main_coords, scale=5, transform shape]
      \tdplotsetrotatedcoords{0}{0}{0}
      \begin{scope}[tdplot_rotated_coords]
        \dodeca[thick];
        \dodecaOrientations[dashed, -stealth, shorten <=2pt, shorten >=2pt];
      \end{scope}
    \end{tikzpicture}
    \caption{$0^\circ$}
  \end{subfigure}
  \begin{subfigure}[t]{0.32\linewidth}
    \centering
    \tdplotsetmaincoords{95}{160}
    \begin{tikzpicture}[tdplot_main_coords, scale=5, transform shape]
      \tdplotsetrotatedcoords{0}{-41.81}{0}
      \begin{scope}[tdplot_rotated_coords]
        \dodeca[thick];
        \dodecaOrientations[dashed, -stealth, shorten <=2pt, shorten >=2pt];
      \end{scope}
    \end{tikzpicture}
    \caption{\ang{41.81}}
  \end{subfigure}
  \begin{subfigure}[t]{0.32\linewidth}
    \centering
    \tdplotsetmaincoords{95}{160}
    \begin{tikzpicture}[tdplot_main_coords, scale=5, transform shape]
      \tdplotsetrotatedcoords{-37.76}{-70.53}{-37.76}
      \begin{scope}[tdplot_rotated_coords]
        \dodeca[thick];
        \dodecaOrientations[dashed, -stealth, shorten <=2pt, shorten >=2pt];
      \end{scope}
    \end{tikzpicture}
    \caption{\ang{70.53}}
  \end{subfigure}
  \caption{Tilt angles inducing a positive thrust allowed by the different possible module orientations.\label{fig:modular:positive_thrust_orientations}}
\end{figure}
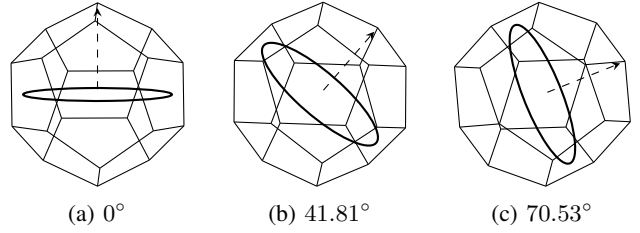

By further grouping feasible rotor orientations (the direction of the thrust vector) by tilt angle (that is the angle they form with the vertical axis of the reference dodecahedron), six groups are obtained.
The three groups with angles inferior to $\pi/2$ radians, corresponding to a positive vertical thrust, are represented in \cref{fig:modular:positive_thrust_orientations}.
The first of these three angles is equal to \ang{0} and corresponds to a vertical orientation while the two others are approximately equal to \ang{41.81} and \ang{70.53}.

\subsubsection{Definition of modular configurations}

With the characterization provided by \cref{fact:modular:cube_group_lattice}, feasible modular configurations can finally be defined directly from the positions and orientations of their modules' rotors.

\begin{definition}\label{def:modular:configuration}
  A configuration $\kappa$ of size $n \in \mathbb{N}$ is a triple containing $n$ positions, $n$ orientations, and $n$ rotor spinning directions:
  \begin{equation*}
    \kappa = \{(\bm{p}_i, \bm{\eta}_i, \epsilon_i),\ i \in \nset{0}{n-1}\}.
  \end{equation*}
  \begin{itemize}
    \item $\bm{p}_i \in \mathbb{R}^3$ is the position of module $i$ in the vehicle coordinate frame, with $\sum_{i=0}^{n-1} \bm{p}_i = \bm{0}$ and such that there exists an offset vector $\bm{z} \in \mathbb{R}^3$ with $\forall i \in \nset{0}{n-1},\ \bm{p}_i - \bm{z} \in \mathcal{P}$.
    \item $\bm{\eta}_i \in \mathcal{H}$ is the rotor orientation of module $i$.
    \item $\epsilon_i \in \{-1,1\}$ indicates the spinning direction of the rotor of module $i$, with $\epsilon_i = 1$ for a CW rotor and $\epsilon_i = -1$ for a CCW rotor.
  \end{itemize}
\end{definition}

The fact that the positions of the modules are chosen to belong to $\mathcal{P}$ up to a constant is to accommodate the $\sum_{i=0}^{n-1} \bm{p}_i = \bm{0}$ constraint, which ensures that the center of mass of the vehicle is at $\bm{0}$.
Configurations where all modules have distinct positions and form a connected component are said to be \emph{valid}.

\begin{definition}\label{def:modular:valid_configuration}
  A configuration $\kappa = \{(\bm{p}_i, \bm{\eta}_i, \epsilon_i),\ i \in \nset{0}{n-1}\}$ is valid if, for every $i \in \nset{0}{n-1},\ j\in \nset{0}{n-1}$, such that $i \neq j$,
  \begin{itemize}
    \item $\bm{p}_i \neq \bm{p}_j$,
    \item there is a sequence of translations $\bm{c}_k \in \bar{\mathcal{C}}^{\mathrm{tr}}$ with $k \in \nset{0}{m-1}$ and $m > 0$ such that
      \begin{align*}
        & \forall k \in \nset{0}{m-1},\ \bm{p}_i + \sum_{l=0}^{k} \bm{c}_l \in \{\bm{p}_j\ |\ j \in \nset{0}{n-1}\}, \\
        & \bm{p}_i + \sum_{l=0}^{m} \bm{c}_l = \bm{p}_j.
      \end{align*}
  \end{itemize}
\end{definition}

\subsection{Dynamics}

\subsubsection{Equations of motion}

The dynamics of a configuration $\kappa$ are formulated from rigid body kinematics and forces and moments induced by each module's rotor.
The body-fixed frame $\mathcal{B}$, which is the frame of the configuration already used to describe feasible positions and orientations, is attached to the assembled vehicle to describe its motion with respect to an inertial frame $\mathcal{I}$.
The position and orientation of the vehicle in $\mathcal{I}$ are given by the vector $\bm{x} \in \mathbb{R}^3$ and the attitude matrix $\bm{R}\in SO(3)$.
The angular velocity of the vehicle in $\mathcal{B}$ is given by the vector $\bm{\Omega} \in \mathbb{R}^3$.

Each module is assumed to generate a thrust and torque both proportional to the square of the rotation speed of its rotor.
The total thrust and torque induced on the vehicle in $\mathcal{B}$ are respectively written $\bm{M}_{\kappa}^{\mathrm{th}} \bm{u}$ and $\bm{M}_{\kappa}^{\mathrm{to}} \bm{u}$, where $\bm{M}_{\kappa}^{\mathrm{th}} \in \mathbb{R}^{3\times n}$ and $\bm{M}_{\kappa}^{\mathrm{to}} \in \mathbb{R}^{3\times n}$ are the thrust and torque matrices.
$\bm{u} \in \mathcal{U}$ is the control vector, which corresponds to the rotors' squared rotation speeds.
The admissible control set is defined by
\begin{equation}\label{eq:modular:admissible_ctrl_set}
  \mathcal{U}_n := \{\bm{u} \in \mathbb{R}^n\ |\ \forall i \in \nset{0}{n-1}\ u_{\min} \leq \bm{u}_i\leq u_{\max}\},
\end{equation}
where $u_{\min}$ and $u_{\max}$ are positive real values.
Alternatively, the matrix $\bm{A}_n \in\mathbb{R}^{2n\times n}$ and vector $\bm{b}_n \in\mathbb{R}^{2n}$ are defined to represent $\mathcal{U}_n$ with
\begin{equation}\label{eq:modular:matrix_ctrl_bounds}
  \mathcal{U}_n = \{\bm{u} \in \mathbb{R}^n\ |\ \bm{A}_n \bm{u} \leq \bm{b}_n\}.
\end{equation}

The only properties of a modular vehicle that depend on its configuration are its mass, inertia tensor, thrust matrix, and torque matrix, which are all provided below.
\begin{fact}\label{fact:modular:mass_inertia}
The mass $m_\kappa$ and inertia tensor $\bm{J}_\kappa$ of $\kappa$  are determined from the mass $m^{\mathrm{mdl}}$ and inertia tensor $\bm{J}^{\mathrm{mdl}}$ of a module by adding the modules' masses and by using the parallel axis theorem:
\begin{align}
  m_\kappa &= n m^{\mathrm{mdl}}, \label{eq:modular:conf_mass} \\
  \bm{J}_{\kappa} &= \sum_{i=0}^{n-1} \bm{J}^{\mathrm{mdl}} + m^{\mathrm{mdl}} \left(\bm{p}_i^\top \bm{p}_i \bm{I}_3 - \bm{p}_i \bm{p}_i^\top \right)\label{eq:modular:conf_inertia}.
\end{align}
\end{fact}
\begin{fact}\label{fact:modular:thrust_torque_matrices}
The thrust and torques matrices of $\kappa$ are given by
\begin{align}
  \bm{M}_{\kappa}^\mathrm{th} =\ &k^{\mathrm{th}}
  \begin{bmatrix}
    \bm{\eta}_0 & \cdots & \bm{\eta}_{n-1}
  \end{bmatrix}, \label{eq:modular:thrust_mat} \\
  \bm{M}_{\kappa}^{\mathrm{to}} =\ &k^{\mathrm{to}}
  \begin{bmatrix}
    \epsilon_0\bm{\eta}_0 & \cdots & \epsilon_{n-1}\bm{\eta}_{n-1}
  \end{bmatrix} + \nonumber \\
                      &+ k^{\mathrm{th}}
                      \begin{bmatrix}
                        \bm{p}_0 \times \bm{\eta}_0 & \cdots & \bm{p}_{n-1} \times \bm{\eta}_{n-1}
                      \end{bmatrix}. \label{eq:modular:torque_mat}
\end{align}
$k^{\mathrm{th}}$ and $k^{\mathrm{to}}$ are respectively a thrust and a torque coefficient.
\end{fact}

The thrust generated by the rotors leads to the following translational dynamics:
\begin{equation}\label{eq:modular:trans_dyn}
  m_\kappa \ddot{\bm{x}} = \bm{R} \bm{M}_{\kappa}^{\mathrm{th}} \bm{u} + m_\kappa \bm{G}.
\end{equation}
$\bm{G} = \begin{bmatrix}0 & 0 & -g\end{bmatrix}^\top$ is the gravity vector with $g$ the gravity constant.
As to rotational dynamics, they are given by
\begin{equation}\label{eq:modular:rot_dyn}
  \bm{J}_\kappa \dot{\bm{\Omega}} = -\bm{\Omega} \times \bm{J}_\kappa\bm{\Omega} + \bm{M}_{\kappa}^{\mathrm{to}} \bm{u}.
\end{equation}
The kinematic equation relating the derivative of the attitude matrix to the angular velocities is
\begin{equation}
  \dot{\bm{R}} = \bm{R} {\left[\bm{\Omega}\right]}_\times.
\end{equation}
where ${[.]}_\times\ :\ \mathbb{R}^3 \mapsto \mathfrak{so}(3)$ denotes the cross-product from the left.

\subsubsection{Scaling of unmodeled forces and moments}

Multiple aerodynamics effects have not been included in the equations of motion of the vehicle.
Typically, they are considered as disturbances and are handled by the control system, for instance with the use of the integral terms of PID control loops~\cite{luukkonen2011modelling,rajappa2015modeling,brescianini2016design} or by directly measuring force disturbances and using them in control algorithms~\cite{tal2020accurate}.

Although no formal guarantee is usually provided regarding the stability of the control system of UAS with respect to unmodeled dynamics, it is important to estimate how unmodeled effects scale as the number of modules in a configuration increases.
If disturbance terms have a larger marginal increase as modular vehicles grow than modeled terms, then typical control systems may not be able to handle them properly.

Effects of no concern are the ones whose marginal impact per module is independent of the number of modules in a configuration.
They include for example the transitional regime which is needed to achieve a commanded rotor's angular velocity, and which is purely dependent on rotor inertia.
Drag coefficients that affect forward flight also scale proportionally with the number of modules.

A potential effect of concern has to do with rotations of a whole vehicle and the induced relative velocities of modules far from the rotation axis.
Larger velocities could result in adverse consequences by affecting the thrust efficiency of affected modules and increasing drag.
However, a dimensional analysis of \cref{eq:modular:torque_mat,eq:modular:conf_inertia,eq:modular:rot_dyn} shows that as the number of modules increases, the maximum velocity induced by the rotation of a vehicle does not in fact increase.

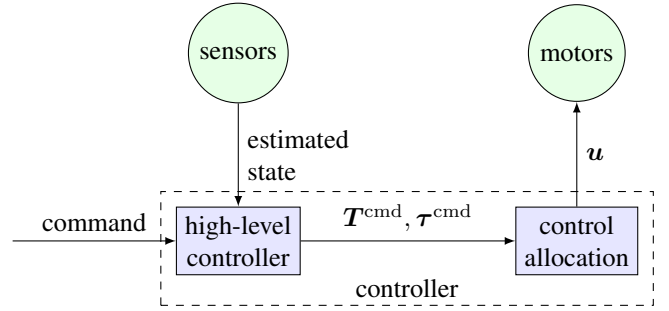
\begin{figure}
  \centering
  \begin{tikzpicture}
    \begin{scope}[
      align=center
      ]
      \node[fill=blue!10,rectangle,draw] (a) at (0,0) {high-level\\controller};
      \node[fill=blue!10,rectangle,draw] (b) at (4.5,0) {control\\allocation};
    \end{scope}
    \node[fill=green!10,circle,draw] (c) at (4.5,2.5) {motors};
    \node[fill=green!10,circle,draw] (d) at (0,2.5) {sensors};
    \draw[-latex] +(-3,0) -- node[above] {command} (a.west);
    \draw[-latex] (a.east) -- node[above] {$\bm{T}^{\mathrm{cmd}},\bm{\tau}^{\mathrm{cmd}}$} (b.west);
    \draw[-latex] (b.north) -- node[right] {$\bm{u}$} (c.south);
    \draw[-latex] (d.south) -- node[right,align=left] {estimated\\state} (a.north);
    \draw[dashed] ($(a.south west)-(0.2,0.4)$) rectangle node[below=8pt] {controller} ($(b.north east)+(0.2,0.2)$);
  \end{tikzpicture}
  \caption{Simplified block diagram of a controller for multirotor vehicles.\label{fig:control:diagram}}
\end{figure}

\section{Control allocation}\label{sec:control}

It is common for multirotor vehicles controller to output a command vector containing a collective thrust and three torque values.
The control allocation problem then consists of obtaining valid motor inputs that result in these commanded thrust and torque, as depicted in \cref{fig:control:diagram}.
Given the linear relationship between motor inputs and vehicle thrust and torque, a standard approach to the control allocation problem is to solve for the inverse of this linear relationship, via a control allocation matrix~\cite{luukkonen2011modelling,faessler2017differential,tal2020accurate}.
For fully actuated multirotor vehicles, the collective thrust is replaced by a three-dimensional thrust, resulting in a six-dimensional control vector~\cite{rajappa2015modeling,brescianini2016design}.

When adapting the control allocation matrix method to large modular vehicles, careful consideration must be taken in the choice of this matrix.
Indeed, overactuation results in non-unique actuation matrix choices that may impact vehicle performance differently.
In this section, several methods are proposed to determine control allocation matrices based on different optimization criteria.
This determination of control allocation matrices relies heavily on tools of convex programming~\cite{boyd2004convex}.
It is assumed throughout the rest of this section that an arbitrary configuration $\kappa$ of $n$ modules is considered.
Some notations of \cref{sec:modular_vehicles} are reused and expanded, although the use of the $\kappa$ subscript for newly introduced quantities is omitted for clarity, even if they may depend on the studied configuration.


\subsection{The control allocation matrix method}

To simplify the definition of the control problem for arbitrary modular configurations, thrusts and torques are grouped together into a six-dimensional \emph{actuation vector}, resulting in a single actuation matrix.
\begin{definition}\label{def:control:thrust_torque_matrix}
  Let $\Mtt \in \mathbb{R}^{6\times n}$, a matrix that bundles the thrust and torque matrices of \cref{fact:modular:thrust_torque_matrices}, be defined as
  \begin{equation*}
    \Mtt
    :=
    \begin{bmatrix}
      \bm{M}_{\kappa}^{\mathrm{th}} \\
      \bm{M}_{\kappa}^{\mathrm{to}}
    \end{bmatrix}.
  \end{equation*}
  This matrix is referred to as the \emph{actuation matrix}.
  In addition, the number of degrees of freedom (DOF) of a vehicle is defined as the rank of its actuation matrix.
  For all practical purposes, it is always between \num{4} and \num{6}.
  A vehicle is therefore said 4DOF, 5DOF, or 6DOF.
\end{definition}

The matrix $\Mtt$ can then be used to define the \emph{control allocation problem}.
\begin{definition}\label{def:control:allocation_pb}
  Let $\bm{t} \in \mathbb{R}^6$.
  A vector $\bm{u} \in \mathcal{U}_n$ is said to solve the control allocation problem for the command vector $\bm{t}$ if
  \begin{equation*}
    \Mtt \bm{u} = \bm{t}.
  \end{equation*}
\end{definition}

Clearly, not all vectors of $\mathbb{R}^6$ result in a feasible control problem, given that $\mathcal{U}_n$ is bounded and $\Mtt$ may not even be full-rank.
Vectors that result in a feasible control allocation problem are called \emph{reachable actuation vectors}.
\begin{definition}\label{def:control:reachable_thrust_torque}
  Let $\mathcal{T} \subset \mathbb{R}^6$, the global set of reachable actuation vectors, be defined as
  \begin{equation*}
    \mathcal{T} := \left\{\Mtt \bm{u}\ |\ \bm{u} \in \mathcal{U}_{n}\right\}.
  \end{equation*}
\end{definition}

In this work, the control allocation problem is solved via a control allocation matrix method.
\begin{definition}\label{def:control:allocation_matrix}
  A matrix $\bm{C} \in \mathbb{R}^{n \times 6}$ is a control allocation matrix if
  \begin{equation*}
    \forall \bm{t} \in \range{\Mtt},\ \Mtt \bm{C} \bm{t} = \bm{t}.
  \end{equation*}
\end{definition}
With that definition, it is clear that for $\bm{t} \in \range{\Mtt}$, if $\bm{C} \bm{t}$ is a feasible control vector, then it solves the control allocation problem.

To eliminate redundant control allocation matrices, which are matrices that are distinct but equal when restricted to inputs in $\range{\Mtt}$, \emph{proper} control matrices are defined.
\begin{definition}\label{def:control:proper_allocation_matrix}
  A control allocation matrix $\bm{C}$ is proper if
  \begin{equation*}
    \forall \bm{t} \in \range{\Mtt}^\perp,\ \bm{C} \bm{t} = \bm{0}.
  \end{equation*}
\end{definition}

\begin{fact}\label{fact:control:allocation_matrix}
  Let $\bm{H} \in \mathbb{R}^{6\times 6}$ be the orthogonal projector on $\range{\Mtt}$ (which can be obtained from a singular value decomposition of $\Mtt$).
  Then
  \begin{enumerate}
    \item \cref{def:control:allocation_matrix,def:control:proper_allocation_matrix} are respectively equivalent to
      \begin{align*}
        \Mtt \bm{C} &= \bm{H}\ \text{and} \\
        \bm{C} \bm{H} &= \bm{C},
      \end{align*} \label{fact:control:allocation_matrix:1}
    \item if $\bm{C}_1$ and $\bm{C}_2$ are two proper control allocation matrices such that $\forall \bm{t} \in \range{\Mtt},\ \bm{C}_1 \bm{t} = \bm{C}_2 \bm{t}$, then $\bm{C}_1 = \bm{C}_2$. \label{fact:control:allocation_matrix:2}
  \end{enumerate}
\end{fact}
\begin{proof}
  The first result comes from the definition of $\bm{H}$ as the orthogonal projector on $\range{\Mtt}$, which also implies that $\bm{I}_6 - \bm{H}$ is the orthogonal projector on $\range{\Mtt}^\perp$.
  The second result is immediate from the orthogonal decomposition theorem:
  \[
    \mathbb{R}^6 = \range{\Mtt} \oplus \range{\Mtt}^\perp.
  \]
\end{proof}

From now on, it is assumed that all allocation matrices are proper.

With the control allocation matrix method, no guarantee is provided regarding the feasibility of the obtained control vector.
In general, feasibility is checked \emph{a posteriori} and if a control vector is not feasible, it is scaled back to bring it in the feasible control set.
The set of actuation vectors that yields a feasible control vector via an allocation matrix is defined in the following.
\begin{definition}\label{def:control:reachable_matrix_thrust_torque}
  Let $\mathcal{T}_{\bm{C}} \subset \mathbb{R}^6$, the set of reachable actuation vectors under a control allocation matrix $\bm{C}$, be defined as
  \begin{equation*}
    \mathcal{T}_{\bm{C}} := \left\{\bm{t} \in \mathbb{R}^6\ |\ \bm{C}\bm{t} \in \mathcal{U}_{n}\right\}.
  \end{equation*}
\end{definition}

Different allocation matrices can be compared via the computation of this set and, for example, the size of the largest sphere contained in this set.
For a specific control matrix, this set can also be compared to the global set of reachable actuation as defined in \cref{def:control:reachable_thrust_torque}.
However, the latter set being a six-dimensional projection of a higher-dimensional polytope, it can be notoriously hard to compute for large $n$~\cite{zhen2018computing,oh1999solving}.

\subsection{Control allocation matrix choice}

For multirotor vehicles with as many degrees of freedom as rotors, like 4DOF quad-rotorcraft or 6DOF hexa-rotorcraft, the choice of the allocation matrix is always unique and given by the Moore-Penrose inverse of $\Mtt$.
However, for vehicles with more rotors than degrees of freedom, there are an infinite number of allocation matrices.
Although the default choice of flight controller software has often still been the Moore-Penrose inverse for these vehicles, it is not always the best choice, particularly for large modular configurations.
These shortcomings are explained in the following.

\subsubsection{Moore-Penrose inverse}


Given $\Mtt$, its Moore-Pensore inverse is denoted by $\Mtt^+$.
\begin{fact}\label{fact:control:moore_penrose_allocation}
  $\Mtt^+$ is a proper allocation matrix.
\end{fact}
\begin{proof}
  The definition of the Moore-Pensore inverse ensures that $\Mtt\Mtt^+$ is the orthogonal projection matrix on $\range{\Mtt}$, from which \cref{fact:control:allocation_matrix} with $\bm{C} = \Mtt$ concludes immediately.
\end{proof}

$\Mtt^+$ is known to be the choice of allocation matrix that provides the control input of minimal euclidean norm, that is $\forall \bm{t} \in \range{\Mtt},\ \forall \bm{u} \in \mathbb{R}^n,\ \Mtt \bm{u} = \bm{t} \implies \norm{\Mtt^+ \bm{t}}_2 \leq \norm{\bm{u}}_2$.
This property has the advantage of minimizing overall control effort, but can be inadequate for large modular configurations.
Indeed, for the purpose of generating a moment via different thrust (such as a pitching or rolling moment on a standard configuration), the further a rotor is from the moment axis, the greater lever arm and therefore actuation efficiency it has.
Achieving a target moment while minimizing the euclidean norm of the control vector would therefore result in the rotors far from the moment axis to be allocated a higher control input.
This fact can be seen easily by observing that the solution of quadratic minimization problem under a linear constraint is a vector proportional to the coefficients of the vector of constraints.
As a result, the uneven use of the different motors can lead to actuator saturation and thus the impossibility to achieve a desired moment target.

A different allocation matrix can be chosen with the goal of mitigating this effect.
For highly symmetric rotor configurations, it is sometimes possible to choose such an allocation matrix manually~\cite{saldana2018modquad}, but systematic methods need to be developed for arbitrary vehicle configurations.
In this section, it is shown that the determination of optimal control allocations matrices with respect to a couple of different objectives can be done with convex programming.

\subsubsection{Reachable actuation set maximization}

In the following, the characterization of the admissible control set with linear inequalities as given in \cref{eq:modular:matrix_ctrl_bounds} is used, but the $n$ subscript in $\bm{A}_n$ and $\bm{b}_n$ is dropped for brevity.
The vector in the actuation space corresponding to hover conditions is given by
\begin{equation}\label{eq:control:hover_tt}
  \th := \begin{bmatrix}0 & 0 & m_\kappa g & 0 & 0 & 0 \end{bmatrix}^\top.
\end{equation}
It is assumed that $\th \in \mathcal{T}$, which is a necessary condition for a configuration to be able to hover.
A fixed control vector $\uh$ such that $\Mtt \uh = \th$ is also assumed provided.
Indeed, regardless of the chosen control matrix, it may be desirable to fix the control vector corresponding to nominal hover conditions, for example with $\uh = \Mtt^+ \th$ to ensure that this vector is of minimum norm.
For configurations where modules have equal tilt angle, this choice of $\uh$ results in a vector with equal entries.

To reduce the potential for actuation vectors to lead to unfeasible control vectors with a control allocation matrix, the following program is formulated.
Its goal is to maximize the radius $r$ of a sphere included in $\mathcal{T}_{\bm{C}}$ and centered in $\th$, where $r$ and $\bm{C}$ are the variables.

{%
  \everymath{\displaystyle}
  \begin{program}\label{prog:control:alloc_max_auth_1}
    \begin{equation*}
      \maximize{r}
      {\Mtt\bm{C}=\bm{H}}
      {\bm{C}\bm{H}=\bm{C}}
      {\norm{\bm{t}}_2 \leq r \implies \bm{A} \left( \uh + \bm{C} \bm{t} \right) \leq \bm{b}.}
    \end{equation*}
  \end{program}
}
In a manner inspired by the problem of the Chebyshev center of a polyhedral set~\cite{boyd2004convex}, with the change of variable $s = \frac{1}{r}$ and by noticing that
\begin{align*}
    &\sup \left\{\bm{A}_{i,:} \bm{C} \bm{t}\ |\ \bm{t} \in \mathbb{R}^6, \norm{\bm{t}}_2 \leq r \right\} \\
  = &r\norm{\bm{C}^\top \bm{A}_{i,:}^\top}_2,
\end{align*}
where $\bm{A}_{i,:}$ denotes the $i$-th row of $\bm{A}$ and $\bm{b}_i$ the $i$-th value of $\bm{b}$,
\cref{prog:control:alloc_max_auth_1} can be reformulated to:
{%
  \everymath{\displaystyle}
  \begin{program}\label{prog:control:alloc_max_auth_2}
    \begin{equation*}
      \minimize{s}
      {%
        \Mtt \bm{C} = \bm{H}
      }
      {%
        \bm{C} \bm{H} = \bm{C}
      }
      {%
        \forall i \in \nset{0}{2n-1}\ \norm{\bm{C}^\top \bm{A}_{i,:}^\top}_2 \leq s \bm{b}_i - s \bm{A}_{i,:} \uh.
      }
  \end{equation*}
\end{program}
}

\Cref{prog:control:alloc_max_auth_2} is a second-order cone program, a type of optimization program that can be solved with algorithms having deterministic convergence properties~\cite{boyd2004convex}.
Instead of maximizing the radius of a sphere, it is also possible to maximize the size of an ellipse of fixed proportions in a subspace of $\mathbb{R}^6$ by replacing
\begin{equation}\label{eq:control:ellipse}
  \norm{\bm{t}}_2 \leq r \implies \bm{A} \left( \uh + \bm{C} \bm{t} \right) \leq \bm{b}
\end{equation}
with 
\[
  \sqrt{\bm{t}^\top \bm{Q} \bm{t}} \leq r \implies \bm{A} \left( \uh + \bm{C} \bm{S} \bm{t} \right) \leq \bm{b}
\]
in \cref{prog:control:alloc_max_auth_2}, where $\bm{Q} \in \mathbb{R}^{6 \time 6}$ is the nonnegative-definite matrix representation of such an ellipse and $\bm{S}$ the orthogonal projector matrix on $\range{\bm{Q}}$, that is the subspace on which this ellipse has non zero dimensions.
In the inequalities of \cref{prog:control:alloc_max_auth_2} $\norm{\bm{C}^\top \bm{A}_{i,:}^\top}_2$ becomes $\sqrt{\bm{A}_{i,:} \bm{C} \bm{Q}^+ \bm{C}^\top \bm{A}_{i,:}^\top}$, which leads to
{%
  \everymath{\displaystyle}
  \begin{program}\label{prog:control:alloc_max_auth_3}
    \begin{equation*}
      \minimize{s}
      {\Mtt \bm{C} = \bm{H}}
      {\bm{C} \bm{H} = \bm{C}}
      {%
        \begin{array}{l}
          \forall i \in \nset{0}{2n-1}\dots \\
          \kern1em \sqrt{\bm{A}_{i,:} \bm{C} \bm{Q}^+ \bm{C}^\top \bm{A}_{i,:}^\top} \leq s \bm{b}_i - s \bm{A}_{i,:} \uh
        .\end{array}
      }
    \end{equation*}
  \end{program}
}
The reason for using an ellipse instead of a sphere is to prioritize certain dimensions of the actuation space along which actuator saturation should not occur, or possibly not include them at all in the maximization of the control authority.
In fact, using an ellipse always makes more sense than a sphere when mixing thrust and torque since they have different units and thus need to be properly scaled.

\subsubsection{Power consumption optimization}

The two choices of allocation matrices introduced so far, that is the Moore-Penrose inverse of the actuation matrix and the one found by solving \cref{prog:control:alloc_max_auth_3}, are based on a minimization of the sum of the squared control inputs and on the maximization of the reachable actuation set.
None of these choices addresses directly the problem of optimizing power consumption despite its obvious importance.
For this reason, an other method for computing an allocation matrix based on power consumption considerations is introduced. 

For the purpose of this work, it is assumed that the power consumed by the rotor of module $i$ is proportional to $\bm{u}_i^{\frac{3}{2}}$.
This relationship is consistent with the modeling a rotor based on momentum theory~\cite{leishman2006principles} or blade element momentum theory~\cite{mccrink2015blade}.
Minimizing power consumption under the constraint $\Mtt \bm{u} = \bm{t}$ for an arbitrary $\bm{t} \in \mathbb{R}^6$ does not result in a closed-form solution, unlike the case of minimizing the sum of the squared control inputs, for which the optimal vector is $\Mtt^+ \bm{t}$.
There is therefore no allocation matrix that minimizes power consumption for every possible actuation vector.

A possible solution to this limitation is to assume that the actuation command vectors issued by the high-level controller are drawn from a multivariate normal distribution:
\begin{equation}
  \bm{t} \sim \mathcal{N}\left(\bm{t}^{\mathrm{h}}, \bm{\Sigma}\right),
\end{equation}
where $\bm{\Sigma} \in \mathbb{R}^{6 \times 6}$ is a covariance matrix.
Under this assumption, the control input for module $i$ is a random variable given by
\begin{equation}
  \bm{u}_i \sim \mathcal{N}(\bm{C}_{i,:} \th, \bm{C}_{i,:} \Sigma \bm{C}_{i,:}^\top).
\end{equation}
Since the power consumption of that module is proportional to $\abs{\bm{u}_i}^{\frac{3}{2}}$, where the absolute value is to ensure the validity of the expression, the average power consumption $P_i$ of module $i$ is the absolute moment of order \num[parse-numbers=false]{\frac{3}{2}} of $\bm{u}_i$:
\begin{equation}
  P_i \propto \mathbb{E}\left(\abs{\bm{u}_i}^{\frac{3}{2}}\right).
\end{equation}
This absolute moment is given by the confluent hypergeometric function of the first kind, denoted $\prescript{}{1}F_{1}(a, b, z)$, with $a=-\frac{3}{4}$, $b=\frac{1}{2}$, and $z=-\frac{(\bm{C}_{i,:}\th)^2}{2\bm{C}_{i,:} \bm{\Sigma} \bm{C}_{i,:}^\top}$~\cite{winkelbauer2012moments}.
Under the assumption $\bm{C} \th = \uh$, the average power consumption of module $i$ is then
\begin{equation}\label{eq:control:avg_pow}
  P_i \propto {\left(\bm{C}_{i,:} \Sigma \bm{C}_{i,:}^\top\right)}^{\frac{3}{4}} \prescript{}{1}F_{1}\left(-\frac{3}{4}, \frac{1}{2}, -\frac{(\uh_i)^2}{2\bm{C}_{i,:} \Sigma \bm{C}_{i,:}^\top}\right).
\end{equation}

In Appendix~\labelcref{appendix:conf_hyper_conv}, a proof of the convexity of
\begin{equation}
  x \mapsto x^{-\frac{3}{4}} \prescript{}{1}F_{1}\left(-\frac{3}{4}, \frac{1}{2}, x\right)
\end{equation}
over $(0, +\infty)$ is given, thereby proving the convexity of \cref{eq:control:avg_pow} with respect to $\bm{C}$.
From this convexity property, a convex optimization problem is formulated, in which the maximum average power consumption per module is minimized:
{%
  \everymath{\displaystyle}
  \begin{program}\label{prog:control:max_avg_power_min}
    \begin{equation*}
      \minimize{\lambda}
      {}
      {%
        \kern-2em%
          \forall i \in \nset{0}{n-1},\ \dots
        }
        {%
        \kern-1em%
          {\left(\bm{C}_{i,:} \Sigma \bm{C}_{i,:}^\top\right)}^{\frac{3}{4}} \prescript{}{1}F_{1}\left(-\frac{3}{4}, \frac{1}{2}, -\frac{{({\uh}_i)}^2}{2\bm{C}_{i,:} \bm{\Sigma} \bm{C}_{i,:}^\top}\right) \leq \lambda
      }
      {%
        \kern-2em%
        \Mtt \bm{C} = \bm{H}
      }
      {%
        \kern-2em%
        \bm{C}\bm{H} = \bm{C}.
      }
    \end{equation*}
  \end{program}
}

\subsection{Control allocation matrices comparison}

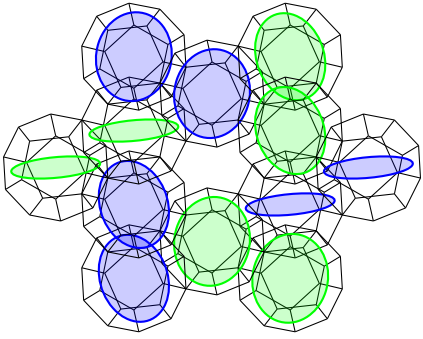
\begin{figure}
  \centering
    \tdplotsetmaincoords{35}{210}
    \begin{tikzpicture}[tdplot_main_coords,scale=3,transform shape]
      \dodecaconf;
    \end{tikzpicture}
    \caption{6DOF 12-rotor configuration.\label{fig:control:optimized:1}}
\end{figure}

Three different methods to compute a control allocation matrix for a given vehicle have been discussed.
A comparison of the matrices resulting from these methods is given for an overactuated, 6DOF, 12-rotor vehicle configuration shown in \cref{fig:control:optimized:1}.
That vehicle was configured specifically for its high yaw authority.

Multiple matrices $\bm{C}_s$ were computed as solutions of \cref{prog:control:alloc_max_auth_3}, each from a different ellipse formulation given by 
\begin{equation*}
  \bm{Q}_s
  =
  \begin{bmatrix}
    \frac{1}{s} \bm{I}_3 & \bm{0} \\
    \bm{0} & s\bm{I}_3
  \end{bmatrix}.
\end{equation*}
It can be seen from the definition of $\bm{Q}_s$ and \cref{eq:control:ellipse} that it corresponds to prioritizing torques versus thrusts with a ratio of $s$.
For each matrix $\bm{C}_s$, the radii of the largest spheres included respectively in the thrust space and torque space that result in a feasible control vector were computed.
They are shown in \cref{fig:control:ctrl_authority}.
In addition, it was observed that when $s$ tends to $0$, the resulting set of reachable actuation vectors becomes close to the set obtained with the allocation matrix $\Mtt^+$.

\begin{figure}
  \centering
  \begin{tikzpicture}
    \begin{axis}[%
      scale only axis,
      height=0.3\linewidth,
      xmode=log,
      xlabel=$s$,
      xtick pos=left,
      axis y line*=left,
      ylabel=Normalized power
      ]
      \addplot[thick] coordinates {%
          (0.100,0.966) (0.112,0.966) (0.126,0.966) (0.141,0.965) (0.158,0.965) (0.178,0.964) (0.200,0.963) (0.224,0.963) (0.251,0.962) (0.282,0.960) (0.316,0.959) (0.355,0.959) (0.398,0.961) (0.447,0.964) (0.501,0.967) (0.562,0.970) (0.631,0.973) (0.708,0.976) (0.794,0.979) (0.891,0.982) (1.000,0.984) (1.122,0.987) (1.259,0.990) (1.413,0.991) (1.585,0.992) (1.778,0.995) (1.995,0.996) (2.239,0.997) (2.512,0.997) (2.818,0.998) (3.162,0.998) (3.548,0.997) (3.981,0.998) (4.467,0.998) (5.012,0.998) (5.623,0.998) (6.310,0.998) (7.079,0.998) (7.943,0.997) (8.913,1.000) (10.000,1.000)
        };
      \addplot[dashed] coordinates {(0.1,0.9584) (10,0.9584)};
      \addplot[dotted] coordinates {(0.1,0.9675) (10,0.9675)};
    \end{axis}
  \end{tikzpicture}
  \caption{%
    Maximum module average power under allocation $\bm{C}_s$ as a function of $s$.
    The dashed line corresponds to the allocation matrix solution of \cref{prog:control:max_avg_power_min} and the dotted one to $\Mtt^+$.\label{fig:control:power_consumption}
  }
\end{figure}
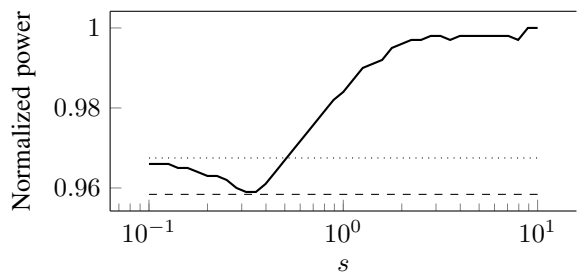

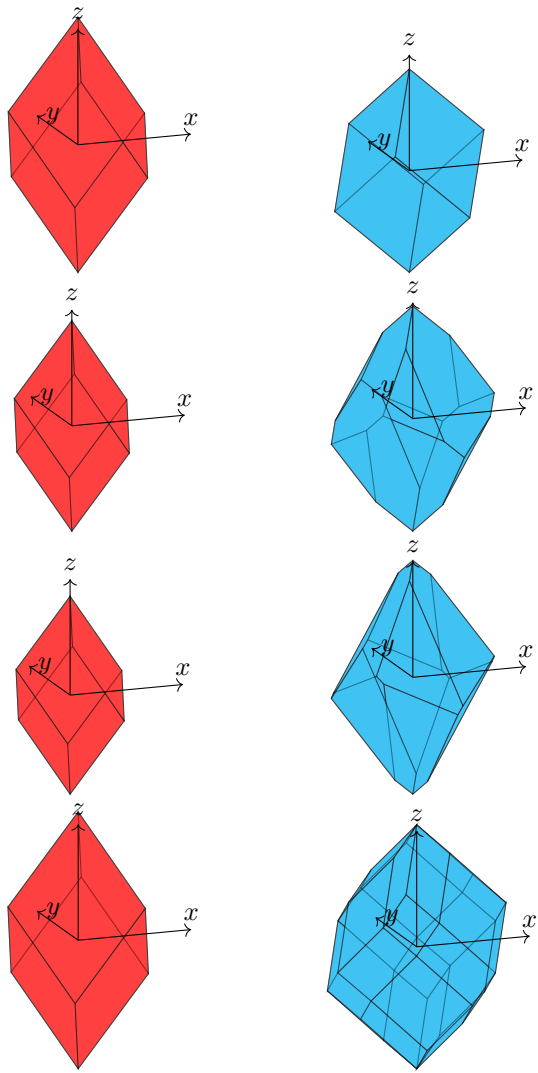
\begin{figure}
  \centering
  \begin{subfigure}{0.48\linewidth}
    \centering
    \tdplotsetmaincoords{75}{-20}
    \begin{tikzpicture}[tdplot_main_coords, scale=0.008]
      \begin{scope}[fill=red]
        \plotCtrlSet[200]{\prAthVertices}{\prAthFaces};
      \end{scope}
    \end{tikzpicture}
  \end{subfigure}
  \begin{subfigure}{0.48\linewidth}
    \centering
    \tdplotsetmaincoords{75}{-20}
    \begin{tikzpicture}[tdplot_main_coords, scale=0.016]
      \begin{scope}[fill=cyan]
        \plotCtrlSet[100]{\prAtoVertices}{\prAtoFaces};
      \end{scope}
    \end{tikzpicture}
  \end{subfigure}
  \begin{subfigure}{0.48\linewidth}
    \centering
    \tdplotsetmaincoords{75}{-20}
    \begin{tikzpicture}[tdplot_main_coords, scale=0.008]
      \begin{scope}[fill=red]
        \plotCtrlSet[200]{\prBthVertices}{\prBthFaces};
      \end{scope}
    \end{tikzpicture}
  \end{subfigure}
  \begin{subfigure}{0.48\linewidth}
    \centering
    \tdplotsetmaincoords{75}{-20}
    \begin{tikzpicture}[tdplot_main_coords, scale=0.016]
      \begin{scope}[fill=cyan]
        \plotCtrlSet[100]{\prBtoVertices}{\prBtoFaces};
      \end{scope}
    \end{tikzpicture}
  \end{subfigure}
  \begin{subfigure}{0.48\linewidth}
    \centering
    \tdplotsetmaincoords{75}{-20}
    \begin{tikzpicture}[tdplot_main_coords, scale=0.008]
      \begin{scope}[fill=red]
        \plotCtrlSet[200]{\prCthVertices}{\prCthFaces};
      \end{scope}
    \end{tikzpicture}
  \end{subfigure}
  \begin{subfigure}{0.48\linewidth}
    \centering
    \tdplotsetmaincoords{75}{-20}
    \begin{tikzpicture}[tdplot_main_coords, scale=0.016]
      \begin{scope}[fill=cyan]
        \plotCtrlSet[100]{\prCtoVertices}{\prCtoFaces};
      \end{scope}
    \end{tikzpicture}
  \end{subfigure}
  \begin{subfigure}{0.48\linewidth}
    \centering
    \tdplotsetmaincoords{75}{-20}
    \begin{tikzpicture}[tdplot_main_coords, scale=0.008]
      \begin{scope}[fill=red]
        \plotCtrlSet[200]{\pMaxthVertices}{\pMaxthFaces};
      \end{scope}
    \end{tikzpicture}
  \end{subfigure}
  \begin{subfigure}{0.48\linewidth}
    \centering
    \tdplotsetmaincoords{75}{-20}
    \begin{tikzpicture}[tdplot_main_coords, scale=0.016]
      \begin{scope}[fill=cyan]
        \plotCtrlSet[100]{\pMaxtoVertices}{\pMaxtoFaces};
      \end{scope}
    \end{tikzpicture}
  \end{subfigure}
  \caption{Top three rows:
    Projections of the sets of reachable actuation vectors as in \cref{def:control:reachable_matrix_thrust_torque} on spaces of thrusts (left) and torques (right), under allocation matrix $\bm{C}_s$, with $s={0.1,1,10}$, from top to bottom.
  Bottom row:
Same projections of the global set of reachable actuation vectors as in \cref{def:control:reachable_thrust_torque}. All sets are based on the same vehicle configuration shown in \cref{fig:control:optimized:1}.\label{fig:control:reachable_set}.
Identical scales are used for each column.}
\end{figure}

The matrix obtained from solving \cref{prog:control:max_avg_power_min} is compared with $\Mtt^+$ and the $\bm{C}_s$ matrices in terms of maximum average power consumption in \cref{fig:control:power_consumption}, and the reachable sets of thrusts and torques in \cref{fig:control:reachable_set}.
It can be seen that differences in maximum power consumption among these matrices remain small.
It is possible that differences are insignificant until a very large number of modules is reached.
However, since \cref{prog:control:max_avg_power_min} is a general convex program and not a conic program, it remains much harder to solve to optimality than \cref{prog:control:alloc_max_auth_3}.
Further analysis therefore remains to be done to determine whether the solution of \cref{prog:control:max_avg_power_min} may prove useful for some rotor configurations.

\begin{figure}
  \centering
  \begin{tikzpicture}
    \ctrlAuthPlot;
  \end{tikzpicture}
  \caption{%
    Control authority along multiple axes for the 12-rotor configuration shown in \cref{fig:control:optimized:1}.
    For each value of $s$, shown on the horizontal axis, $\bm{C}_s$ is computed by solving \cref{prog:control:alloc_max_auth_3} with $\bm{Q}=\bm{Q}_s$.
    The control authority along different subspaces of the actuation vectors is then computed by solving the same \namecref{prog:control:alloc_max_auth_3} with $\bm{C}$ fixed to $\bm{C}_s$ and $\bm{Q}$ changed to the orthogonal projector on the relevant subspace.
    Each plotted line corresponds to a different subspace of the actuation space: three-dimensional thrusts, three-dimensional torques, and one-dimensional thrust and torque along the $x$, $y$, and $z$ axes.\label{fig:control:ctrl_authority}%
  }
\end{figure}
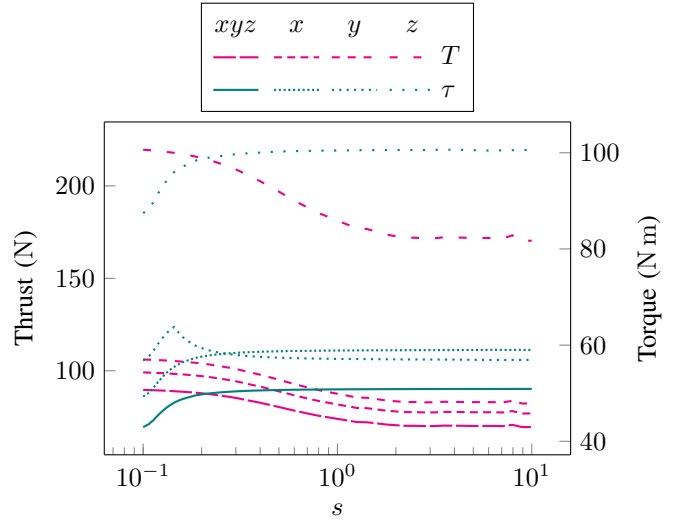

\subsection{Other control allocation methods}

It is worth noting that other techniques exist for the control allocation of multirotor vehicles.
For example, one of them consists of dynamically changing the allocation matrix based on a weighted pseudo-inverse of the actuation matrix~\cite{ducard2011discussion}.
It is also possible to solve the allocation problem in real-time for each issued command actuation vector via a quadratic program~\cite{harkegaard2004dynamic}.
The same could be done to directly minimize the total or maximum power consumption via a general convex program, as in~\cite{dyer2019energy}.
Doing so would ensure that power consumption is minimized for each command vector and not only on average, without reducing the size of the set of achievable actuation vectors as with a fixed allocation matrix.
However, despite the efficiency of modern solvers for some classes of convex optimization problems, it may not be possible to achieve a sufficiently low solving time with configurations of many rotors for the requirements of flight control software, whose inner control loops can run at up to \SI{400}{\Hz}.
For this reason, the focus of this work was placed on matrix-based methods, which need the allocation matrix to be computed only once without any real-time constraint, and which are very efficient to run once this matrix is computed.

\section{Structural aspects of modular configurations}\label{sec:structures}

A main motivation behind the shape and connection system of the Dodecacopter is its ability to be assembled in three dimensions while ensuring that wake interactions between modules at different heights are minimized.
Previously designed modular rotorcraft based on rigid assembly are combined in two dimensions~\cite{oung2011distributed,duffy2015lift,saldana2018modquad} or in three dimensions exclusively by stacking modules on top of each other~\cite{naldi2011class}, thereby reducing their efficiency through wake interactions.
The motivation for the desired three-dimensional assembly ability of the presented system has to do with structural stiffness (or rigidity).
Three-dimensional structures with equal proportions are intuitively stiffer than structures that extend only along two dimensions, since the latter cannot resist efficiently loads applied perpendicularly to the dimensions they extend along.
Stiff structures are, by definition, less subject to deformations than more compliant ones under similar loading conditions.
The advantages of limiting deformations for modular vehicles are obvious:
\begin{itemize}
  \item Deformations may cause a loss of efficiency through a change of the propeller orientations.
  \item They may create vibrations that affect performance.
  \item Large deformations incur nonlinear geometric stiffness changes that can lead to high stresses or buckling.
\end{itemize}

To determine the impact of a vehicle configuration on its stiffness and on its ability to avoid the foregoing negative effects, the standard method of space frame analysis from structural mechanics is employed in this section.
The mechanical connections between modules are assumed stiff enough to be considered perfectly rigid, such that deformations of the frame is due to the flexibility of its members only.
This is a standard assumption of the structural analysis of framed structures that is generally considered a good first-order approximation for small displacements~\cite{kassimali2012matrix}.
In \cref{subsec:structures:structural_characterization}, the structural characterization of modular configurations is formulated in terms of their space frame representation and the elastic stiffness matrix resulting from this representation.
Structural performance indicators of modular configurations based on their space frame representation are introduced in \cref{subsec:structures:structural_performance}.
Finally, \cref{subsec:structures:numerical_comparison} details a numerical comparison of multiple modular configurations based on the introduced performance indicators.

\subsection{Tools of structural analysis}\label{subsec:structures:structural_characterization}

In this subsection, the representation of modular configurations as space frames is formulated.
With this representation, the structural analysis of modular configurations can be performed using linear algebra, through the use of the elastic stiffness matrix.
The general characteristics and significance of this matrix are explained, although its construction, a well-documented process~\cite{kassimali2012matrix}, is not explicitly provided for brevity.

\subsubsection{Frame representation}

Space frames consist of three-dimensional structures made of beams linked at their ends by perfect rigid joints.
Dodecacopter modules, thanks to their dodecahedral shape, fit conveniently to this representation.
The fact that modules connect at vertices of their dodecahedron frame allows vehicles of multiple modules to be represented as frames as well.

\paragraph{Beams}

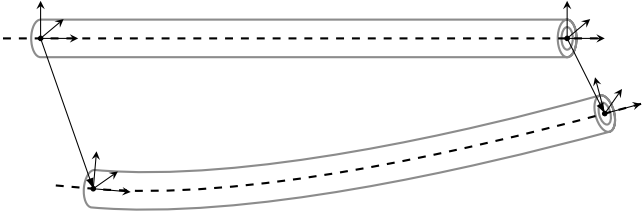
\begin{figure}[h]
  \centering
  \begin{tikzpicture}
    \pic{beam={0}{2}{0}{7}{2}{0}};
    \pic{beam={0.7}{0}{-5}{7.5}{1}{15}};
    \draw[-latex] (0,2) -- (0.7,0);
    \draw[-latex] (7,2) -- (7.5,1);
  \end{tikzpicture}
  \caption{Representation of a beam being deformed under the displacement of the joints at its ends.\label{fig:structures:beam}}
\end{figure}

Beams are slender structural members that can withstand axial and shear forces, torsional and bending moments, and which, for the purpose of this work, are considered made of a single isotropic elastic material.
A beam deformed under external forces is shown in \cref{fig:structures:beam}.
To describe modular configurations as space frames, each edge of the dodecahedra representing the modules in a configuration is assimilated with a single beam.
Since all configurations are made from identical modules and the modules' frames respect the symmetries of the regular dodecahedron, all beam members are identical.
The beams are assumed to be thick-walled cylindrical tubes with a constant cross-section, as shown in \cref{fig:structures:beam}.

\paragraph{Frame topology}

A frame representing a vehicle configuration $\kappa$ is completely characterized by its joint positions and the list of joint pairs linked by members.

\begin{definition}\label{def:structures:conf_frame_joints}
  Given $\kappa = \{(\bm{p}_i,\bm{\eta}_i,\epsilon_i)\ |\ i \in \nset{0}{n-1}\}$, the set of joints in the frame of $\kappa$ is given by
  \begin{equation*}
    \bar{J}_\kappa := \left\{\bm{p}_i + \bm{g}\ |\ i \in \nset{0}{n-1}, \bm{g} \in V^\mathcal{D}\right\}.
  \end{equation*}
  Note that $\njt := \left|\bar{J}_\kappa\right| \leq 20n$, since there are \num{20} joints per module and some joints are shared by multiple modules.
  The $\njt$ joints can then be ordered arbitrarily into a list:
  \begin{equation*}
    J_\kappa = \left[\bm{j}_0,\dots,\bm{j}_{\njt-1}\right],
  \end{equation*}
  such that $\bm{j}_i$ is the element of $\bar{J}_\kappa$ assigned to the index  $i \in \nset{0}{\njt-1}$.
\end{definition}

\begin{definition}\label{def:structures:conf_frame_members}
  Similarly, the members of the frame of $\kappa$ are defined by the list
  \begin{equation}
    E_\kappa := \left[(e_0^-,e_0^+),(e_{1}^-,e_{1}^+),\dots,(e_{\nmb-1}^-,e_{\nmb-1}^+)\right].
  \end{equation}
  $\nmb$ is the number of members in the frame and for $i\in\nset{0}{\nmb-1}$, $(e_{i}^-,e_{i}^+) \in \nset{0}{\njt-1} \times \nset{0}{\njt-1}$ is the pair of joint indices linked by member $i$, where the order of the joints in each pair is arbitrary.
  Here, $\nmb = 30n$ since there are \num{30} members per module and each member belongs to a single module.
\end{definition}


\begin{definition}\label{def:structures:conf_frame}
  The frame of configuration $\kappa$ is defined by
  \begin{equation}
    \mathcal{F}_{\kappa} := \left(J_\kappa,E_\kappa\right),
  \end{equation}
  where $J_\kappa$ and $E_\kappa$ are defined in \cref{def:structures:conf_frame_joints,def:structures:conf_frame_members}.
\end{definition}

\subsubsection{Loads and displacements of frames}

When subject to constant external loads, a frame deforms until reaching equilibrium or failure.
Equilibrium is characterized by the deformation of the members, which, via elastic forces, compensate the external loads, as shown in \cref{fig:structures:deformed_quad:max_compliance}.

Let the external forces and moments applied on frame $\mathcal{F}_\kappa$ be defined by $\left[(\bm{f}_0,\bm{m}_0),\dots,(\bm{f}_{\njt-1},\bm{m}_{\njt-1})\right]$, where $(\bm{f}_i,\bm{m}_i)\in\mathbb{R}^3\times\mathbb{R}^3$ is applied at joint $i\in\nset{0}{\nmb-1}$.
The resulting displacements of the joints at equilibrium are defined by the list $\left[(\bm{s}_0,\bm{r}_0),\dots,(\bm{s}_{\njt-1},\bm{r}_{\njt-1})\right]$, where $\bm{s}_i \in \mathbb{R}^3$ is a translational displacement and $\bm{r}_i \in \mathbb{R}^3$ is the axis-angle representation of a rotation.
These displacements are assumed to be measured relative to the state of the unloaded frame at equilibrium.
With this definition of joint loads and displacements, the aggregated vectors of frame displacements and external loads are given by
\begin{equation}\label{eq:structures:aggregated_vec}
  \bm{v} := \begin{bmatrix}\bm{s}_0 \\ \bm{r}_0 \\ \vdots \\ \bm{s}_{\njt-1} \\ \bm{r}_{\njt-1} \end{bmatrix} \in \mathbb{R}^{6\njt},\ 
  \bm{P} := \begin{bmatrix}\bm{f}_0 \\ \bm{m}_0 \\ \vdots \\ \bm{f}_{\njt-1} \\ \bm{m}_{\njt-1} \end{bmatrix} \in \mathbb{R}^{6\njt}.
\end{equation}



\subsubsection{Elastic stiffness}\label{subsubsec:structures:elastic}

In this work, a first-order approximation of the relationships between $\bm{v}$ and $\bm{P}$ via the \emph{elastic stiffness matrix} $\bm{K}_\kappa \in \mathbb{R}^{6\njt \times 6\njt}$ is used:
\begin{equation}\label{eq:structures:elastic_stiffness}
  \bm{P} = \bm{K}_\kappa \bm{v}.
\end{equation}
This method is appropriate for small displacements and provides a good approximation of a structure's behavior around equilibrium, allowing the comparison of modular configurations from structural performance indicators based on this method.
The linear relationship between displacements and external loads also allows the efficient structural optimization of modular configurations.
The procedure to construct the elastic stiffness matrix of a space frame, which is a symmetric positive semi-definite, can be found in any comprehensive textbook on matrix methods for structural analysis~\cite{weaver1990matrix,kassimali2012matrix}.

Since displacements corresponding to rigid body translations or infinitesimal rotations do not incur any internal loading, the elastic stiffness matrix is not full-rank, but of rank $6\njt-6$.
To fully define the problem of relating displacements with forces and moments, displacements are required to amount to zero rigid body translation and rotation or, equivalently, $\bm{v} \in \nulls{\bm{K}_{\kappa}}^\perp$.
This constraint is represented by
\begin{equation}
  \sum_{i=0}^{\njt-1} \bm{s}_i = \bm{0},\ 
  \sum_{i=0}^{\njt-1} \bm{j}_i \times \bm{s}_i + \bm{r}_i = \bm{0} \label{eq:structures:displacement_constraint}
\end{equation}
and can be simplified to $\bm{H}_{\kappa}\bm{v} = \bm{0}$, where $\bm{v}$ is as defined in \cref{eq:structures:aggregated_vec} and $\bm{H}_{\kappa} \in \mathbb{R}^{6\times 6\njt}$ is the matrix that relates generalized displacements with rigid body translation and rotational displacements.
This matrix can be seen as an orthogonal projection on $\nulls{\bm{K}_{\kappa}}$.
Since $\bm{K}_{\kappa}$ is symmetric, $\nulls{\bm{K}_{\kappa}}^\perp = \range{\bm{K}_{\kappa}}$, which means that every feasible load vector $\bm{P}$ must also verify $\bm{H}_{\kappa} \bm{P} = \bm{0}$.
This feasibility condition can be verified by replacing displacements by loads in \cref{eq:structures:displacement_constraint}, which leads to the natural constraint that sum of forces and moments are zero:
\begin{equation}
  \sum_{i=0}^{\njt-1} \bm{f}_i = \bm{0},\
  \sum_{i=0}^{\njt-1} \bm{j}_i \times \bm{f}_i + \bm{m}_i = \bm{0}. \label{eq:structures:load_constraint}
\end{equation}

These constraints on the loads and displacements ensure that the relationship between them is one-to-one.
In fact, since $\bm{v} \in \range{\bm{K}_{\kappa}} = \range{{\bm{K}_{\kappa}}^+}$, the inverse relationship of \cref{eq:structures:elastic_stiffness} is provided directly by
\begin{equation}\label{eq:structures:elastic_displacements_sol}
  \bm{v} = {\bm{K}_{\kappa}}^+ \bm{K}_{\kappa} \bm{v} = {\bm{K}_{\kappa}}^+ \bm{P}.
\end{equation}

During the assembly of the elastic stiffness matrix, a member stiffness matrix, which gives the relationship between frame displacements and member loads, is also computed.
Computing the loads applied to the individual members is essential to estimate material stresses that could result in local failure.
For a member $i$, $\bm{l}_i \in \mathbb{R}^6$ is defined as the vector containing the axial force, shear forces, torsional moment, and bending moments it is subject to in its local coordinate frame.
The member loads can be aggregated into the vector
\begin{equation}\label{eq:structures:member_loads_vec}
  \bm{F} :=
  \begin{bmatrix}
    \bm{l}_0 \\ \vdots \\ \bm{l}_{\nmb-1}
  \end{bmatrix}.
\end{equation}
A first-order approximation of the linear relationship between external displacements and member loads results in the member stiffness matrix $\bm{K}^{\mathrm{mb}}_{\kappa} \in \mathbb{R}^{6\nmb\times 6\njt}$.
This matrix can be assembled in a process similar to the assembly of $\bm{K}_{\kappa}$~\cite{kassimali2012matrix} that can be performed in parallel.
With this definition, the following holds:
\begin{equation}\label{eq:structures:member_elastic_stiffness}
  \bm{F} = \bm{K}^{\mathrm{mb}}_{\kappa}\bm{v}.
\end{equation}

\subsection{Structural performance}\label{subsec:structures:structural_performance}

Now that the basis for the matrix structural analysis of modular configurations has been laid out, structural performance indicators of arbitrary configurations are defined.
Two of them as specified explicitly in the following.

\subsubsection{Indicators based on stiffness}

\paragraph{Maximum compliance}

For a matrix $\bm{M}$, let $\sigma^{\max}(\bm{M})$ be the maximum singular value of $\bm{M}$.
It is well known that
\begin{equation*}
  \sigma^{\max}(\bm{M}) = \max_{\norm{\bm{x}}_2=1}{\norm{\bm{M}\bm{x}}_2}.
\end{equation*}
Based on this property of singular values and \cref{eq:structures:elastic_displacements_sol}, a possible structural performance indicator of $\kappa$ is $\sigma^{\max}({\bm{K}_{\kappa}}^+)$.
This value represents the maximum norm that a displacement vector resulting from a unit-norm force vector can take.
It is therefore a worst-case scenario that should be sought as small as possible to ensure that a structure is stiff for all arbitrary loading conditions.

A drawback of this performance indicator is that it accounts for load vectors that include both forces and moments, two physical quantities with different units, and has therefore little meaning as such.
The same can be said regarding displacement vectors, which include translations and rotations.
Therefore, a more meaningful performance indicator relies on pure translational displacements that result from external forces only.
Based on \cref{eq:structures:aggregated_vec,eq:structures:elastic_stiffness}, the relationship between translational displacements and external forces is
\begin{equation}\label{eq:structures:elastic_stiffness_tr}
  \begin{bmatrix}
    \bm{s}_0 \\ \vdots \\ \bm{s}_{\njt-1}
  \end{bmatrix}
  =
  {\bm{M}^{\mathrm{tr}}}
  {\bm{K}_{\kappa}}^+
  {\bm{M}^{\mathrm{tr}}}^\top
  \begin{bmatrix}
    \bm{f}_0 \\ \vdots \\ \bm{f}_{\njt-1}
  \end{bmatrix}
  ,
\end{equation}
where
\begin{equation}\label{eq:structures:translation_only}
  \bm{M}^{\mathrm{tr}} =
  \bm{I}_{\njt} \otimes
  {\begin{bmatrix}
      \bm{I}_3 & \bm{0}_{3 \times 3}
  \end{bmatrix}}
  \in \mathbb{R}^{3\njt \times 6\njt}
\end{equation}
and where $\otimes$ is the Kronecker product.
These considerations lead to the performance indicator
\begin{equation}\label{eq:structures:max_compliance_pi}
  \sigma_1 := \sigma^{\max}\left(\bm{M}^{\mathrm{tr}}{\bm{K}_{\kappa}}^+{\bm{M}^{\mathrm{tr}}}^\top\right).
\end{equation}
Since $\sigma^{\max}\left(\bm{M}^{\mathrm{tr}}{\bm{K}_{\kappa}}^+{\bm{M}^{\mathrm{tr}}}^\top\right)$ is a square symmetric matrix, $\sigma_1$ is in fact an eigenvalue and its associated eigenvector of norm one maximizes translational displacements.
Note that three other similar performance measures could be defined, to consider all potential impacts of external forces and moments on translational and rotational displacements.

\paragraph{Maximum member axial load}

Another indicator of performance pertains to the maximum axial load present in a structural frame subject to unit-norm external forces.
The estimation of members' axial loads is crucial, since local buckling, when the compressive axial load in a member exceeds its Euler's critical load, is a common mode of failure of framed structures.
Based on \cref{eq:structures:elastic_displacements_sol,eq:structures:member_loads_vec,eq:structures:member_elastic_stiffness,eq:structures:translation_only}, assuming without loss of generality that each member load vector $\bm{l}_i$ contains the member's axial load as its first component, the axial load $l^{\mathrm{ax}}_i$ of member $i$ is given by
\begin{equation}\label{eq:structures:axial_load}
  l^{\mathrm{ax}}_i := \bm{M}^{\mathrm{ax}}_i\bm{L}
  = \bm{M}^{\mathrm{ax}}_i\bm{K}^{\mathrm{mb}}_{\kappa}{\bm{K}}^+{\bm{M}^{\mathrm{tr}}}^\top\bm{P}^{\mathrm{tr}},
\end{equation}
where
\begin{equation}
  \bm{M}^{\mathrm{ax}}_i :=
  {\bm{e}_{i}^{\nmb}}^\top \otimes \begin{bmatrix}1&0&0&0&0&0\end{bmatrix},
\end{equation}
with $\bm{e}_i^{\nmb}$ being the vector of size $\nmb$ with the $i$-th entry equal to \num{1} and the others equal to \num{0}.
Since $\bm{M}^{\mathrm{ax}}_i\bm{K}^{\mathrm{mb}}_{\kappa}{\bm{K}}^+{\bm{M}^{\mathrm{tr}}}^\top$ is a row matrix, the maximum value reached when it is multiplied on the right by a unit-norm vector is its euclidean norm.
As a consequence, a performance indicator equal to the maximum axial load under unit-norm external forces is defined by
\begin{equation}\label{eq:structures:max_axial_load_pi}
  \sigma_2 :=
  \max_{i \in \nset{0}{\nmb-1}}\norm{\bm{M}^{\mathrm{ax}}_i\bm{K}^{\mathrm{mb}}_{\kappa}{\bm{K}_{\kappa}}^+{\bm{M}^{\mathrm{tr}}}^\top}_2.
\end{equation}

\subsection{Numerical comparison of modular configurations}\label{subsec:structures:numerical_comparison}

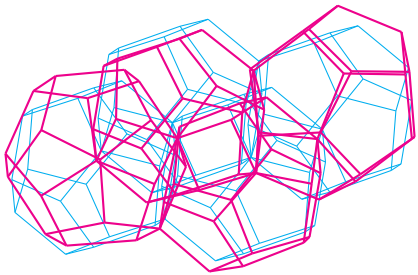
\begin{figure}[h]
  \centering
    \centering
    \tdplotsetmaincoords{60}{65}
    \begin{tikzpicture}[tdplot_main_coords,scale=5,transform shape,thick]
      \quadVertices;
      \quadMaxDisDisplacements;
      \quadMaxDisForces;
      \pic[cyan,thin]{frame={\quadEdges}{}{}};
      \pic[magenta,thick]{frame={\quadEdges}{1}{}};
    \end{tikzpicture}
  \caption{%
    Deformed quadrotor configuration under external forces that maximize the displacement vector, as measured by \cref{eq:structures:max_compliance_pi}, compared with unloaded configuration.
  \label{fig:structures:deformed_quad:max_compliance}}
 \end{figure}

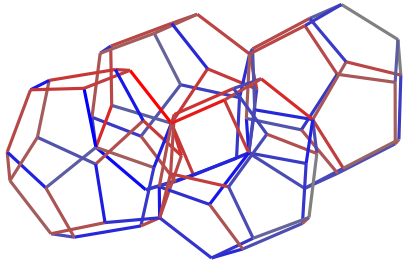
\begin{figure}[h]
  \centering
  \tdplotsetmaincoords{60}{65}
  \begin{tikzpicture}[tdplot_main_coords,scale=5,transform shape,thick]
    \quadVertices;
    \quadMaxAxLDisplacements;
    \quadMaxAxLForces;
    \newcommand\loadcolor[2]{
      \pgfmathparse{floor((#2/100)^(1/3)*100)}
      \ifthenelse{\equal{#1}{t}}{\def\opte{blue!\pgfmathresult!gray}}{\def\opte{red!\pgfmathresult!gray}}
    }
    \pic[magenta,very thick]{frame={\quadEdgesWithMaxAxLoads}{1}{\loadcolor}};
  \end{tikzpicture}
  \caption{
    Deformed configuration under external forces yielding the worst-case axial load as given by \cref{eq:structures:max_axial_load_pi}.
    Red coloring of the members corresponds to compressive loads and blue to tensile loads. Color intensity is proportional to load magnitude.
  \label{fig:structures:deformed_quad:max_ax_load}}
\end{figure}

\begin{figure}
  \centering
  \begin{subfigure}{0.48\linewidth}
    \centering
    \includegraphics[width=\linewidth]{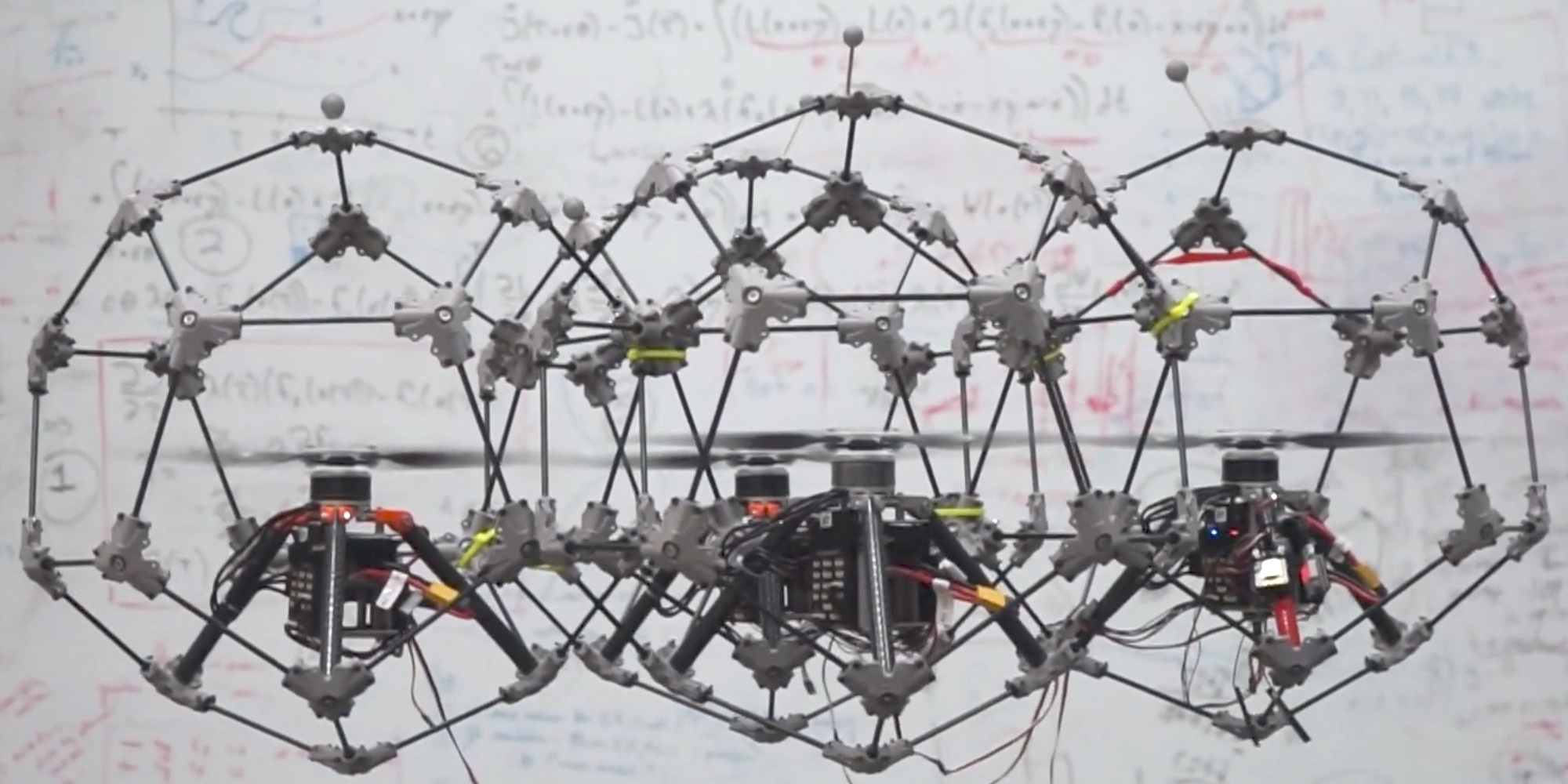}
    \caption{Quadrotor.\label{fig:prototype:confs:quad}}
  \end{subfigure}
  \begin{subfigure}{0.48\linewidth}
    \centering
    \includegraphics[width=\linewidth]{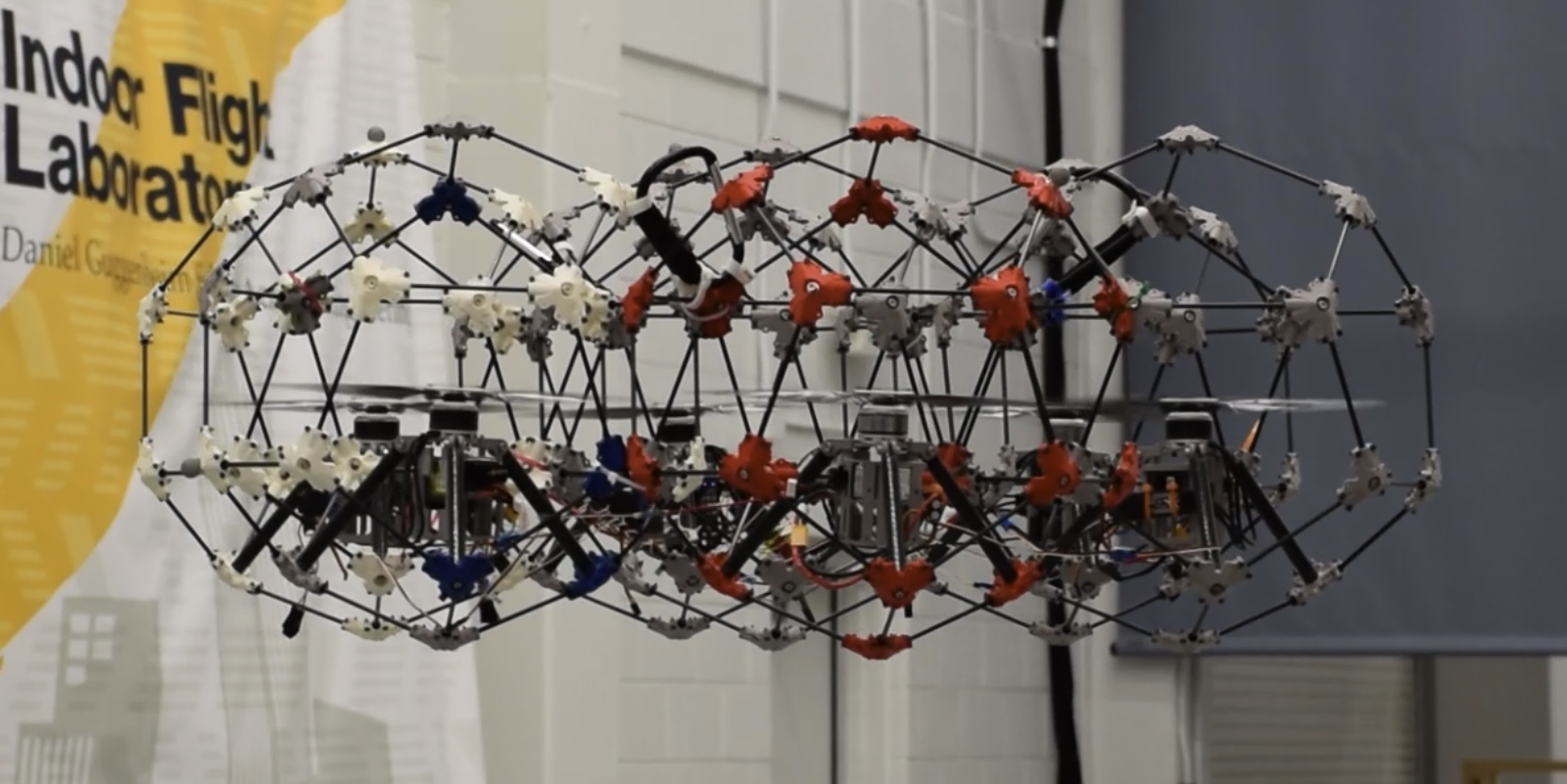}
    \caption{Hexarotor.\label{fig:prototype:confs:hexa}}
  \end{subfigure}
  \begin{subfigure}{0.48\linewidth}
    \centering
    \includegraphics[width=\linewidth]{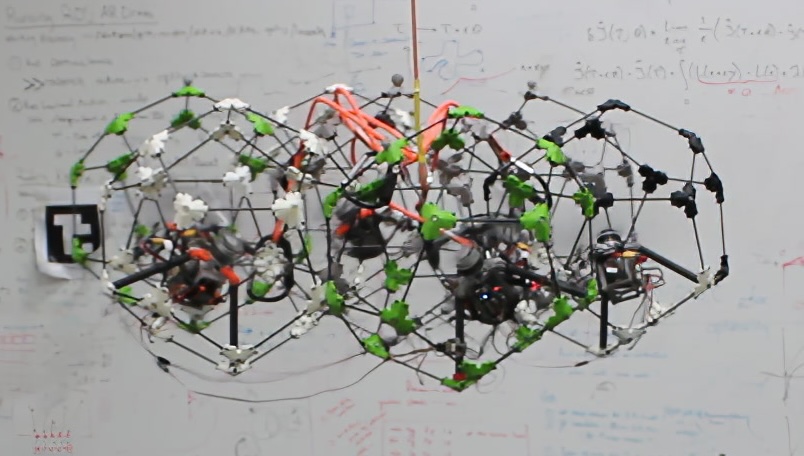}
    \caption{6DOF Hexarotor.\label{fig:prototype:confs:hexa6dofs}}
  \end{subfigure}
  \begin{subfigure}{0.48\linewidth}
    \centering
    \includegraphics[width=\linewidth]{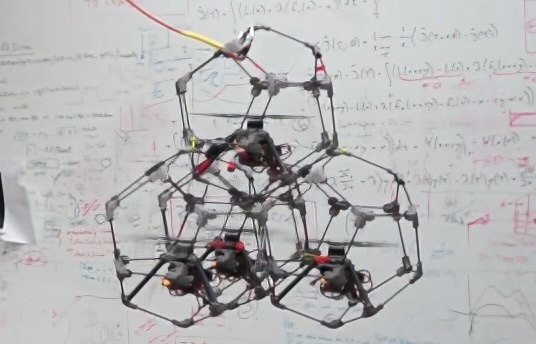}
    \caption{Tetrahedron Quadrotor.\label{fig:prototype:confs:tetra}}
  \end{subfigure}
  \begin{subfigure}{0.48\linewidth}
    \centering
    \includegraphics[width=\linewidth]{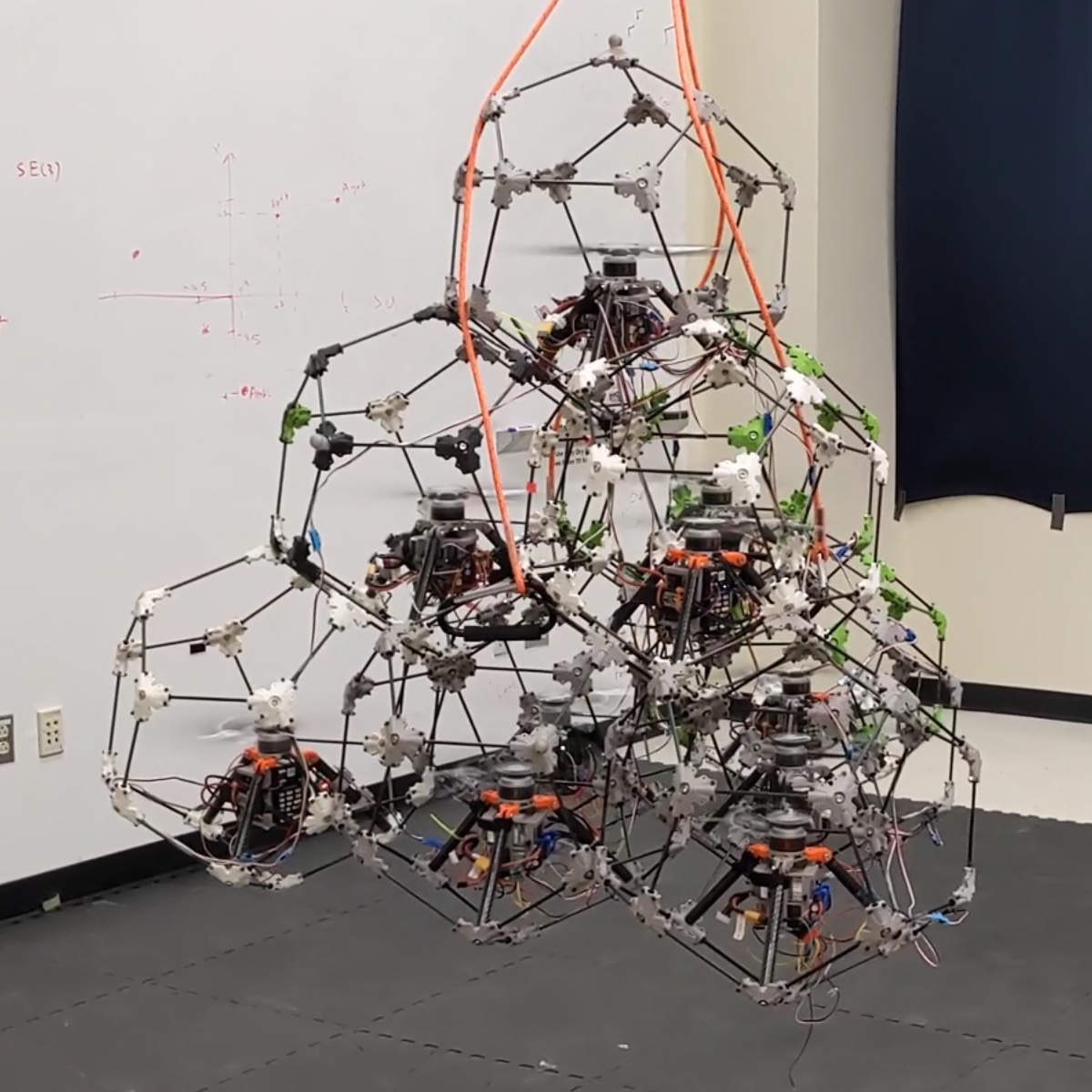}
    \caption{Tetrahedron Decarotor.\label{fig:prototype:confs:decatetra}}
  \end{subfigure}
  \begin{subfigure}{0.48\linewidth}
    \centering
    \includegraphics[width=\linewidth]{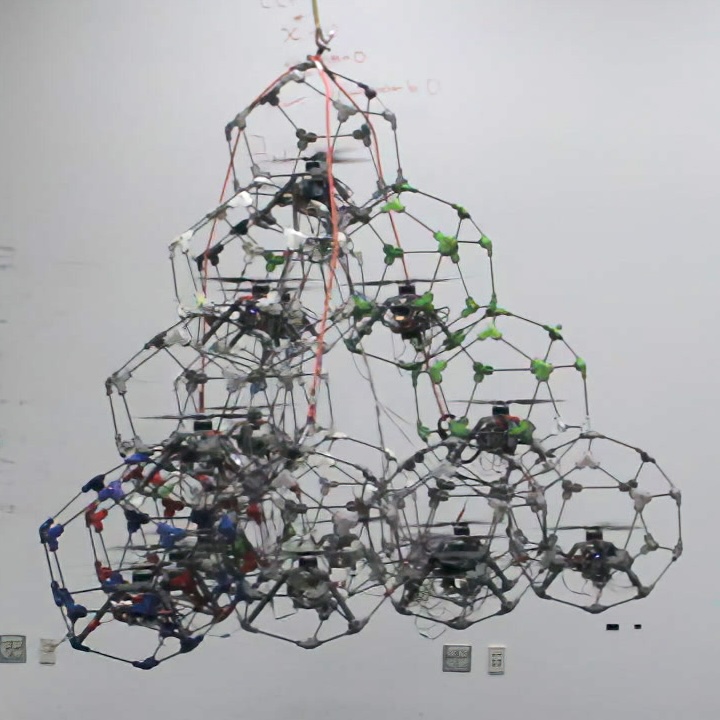}
    \caption{Tetrahedron Hexadecarotor.\label{fig:prototype:confs:tetragen2}}
  \end{subfigure}
  \caption{\label{fig:prototype:confs}}
\end{figure}

The structural performance of six modular configurations was evaluated against the indicators of \cref{subsec:structures:structural_performance}.
The first four analyzed configurations are the standard quad-rotorcraft, a quad-rotorcraft in the shape of a tetrahedron (\cref{fig:prototype:confs:tetra}), a 4-by-4 flat array, and a 16-module tetrahedron formed by combining four tetrahedron-shaped quad-rotorcraft (\cref{fig:prototype:confs:tetragen2}).
These configurations are chosen since they correspond to two pairs of configurations with the same number of modules.
For the sake of completeness, the two other configurations that were flight tested, that is the standard hexarotor (\cref{fig:prototype:confs:hexa}) and a 10-module tetrahedron (\cref{fig:prototype:confs:decatetra}), are also added to the results.
Results are summarized in \cref{table:structures:performance_comparison} and the corresponding external loads for the standard quad-rotorcraft configuration are represented in \cref{fig:structures:deformed_quad:max_compliance,fig:structures:deformed_quad:max_ax_load}.
It can be shown that for an equal number of modules, configurations with more layers have a lower value for the introduced performance indicators, which corresponds to a better performance.

\begin{table}
  \begin{center}
    \begin{tabular}{l c c c c}
      Name & Modules & Layers & $\sigma_1$ & $\sigma_2$ \\
      Quadrotor & 4 & 1 & 1 & 1 \\
      Tetrahedron Quadrotor & 4 & 2 & 0.265 & 0.764 \\
      16-module Array & 16 & 1 & 8.741 & 1.361 \\
      Tetrahedron Hexadecarotor & 16 & 4 & 1.391 & 1.071 \\
      Hexarotor & 6 & 1 & 1.825 & 1.097 \\
      Tetrahedron Decarotor & 10 & 2 & 0.544 & 0.831 \\
    \end{tabular}
  \end{center}
  \caption{%
    Structural performance indicators evaluated against different configurations.
    For the first pairs of configurations, the number of modules is chosen to represent comparable configurations and the number of layers corresponds to the different heights module can be at in a configuration, \num{1} being a flat configuration.
    \label{table:structures:performance_comparison}
  }
\end{table}

\section{Configuration optimization}\label{sec:configuration_optimization}

The programs introduced in \cref{sec:control,sec:structures} are meant to be used on pre-determined modular configurations.
Given the versatility offered by the modular architecture presented in this work and the high number of feasible configurations, it is desirable to design methods to optimize configurations based on their structural and control properties.
Using MIP, it is possible to describe configurations and their properties with binary decision variables, such that the programs already introduced can be extended to account for a parameterized space of possible configurations.
In this section, a MIP formulation of feasible configurations and their relevant properties is provided.
Binary variables indicate whether a module is allocated to a configuration at a given position, with a given orientation, and with a given rotor spinning direction.
The fact that modules must form a connected entity is expressed as a set of linear constraints based on a MIP formulation of contiguity for graphs~\cite{shirabe2005model}.
Constraints to prevent configurations resulting in modules in the wake of other modules are also introduced.
Furthermore, most linear constraints employed in previous programs for control allocation or structural optimization, which become bilinear under the assumption of a parameterized configuration space, are shown to be linearizable with the ``big-M'' method, a standard method of linear programming for linearizing products of variables of which at most one is non-binary~\cite{hillier2015introduction}.

\subsection{Graph of admissible module positions}

The set of positions that modules may take in a configuration and whether a connection exists between modules placed at a pair of positions represents a graph defined in \cref{def:optimal:graph}.
\begin{definition}\label{def:optimal:graph}
  $G_\infty = (V_\infty, E_\infty)$ is the infinite undirected graph (a graph with an infinite number of vertices~\cite{diestel2024graph}) where
  \begin{itemize}
    \item $V_\infty = \mathcal{P}$, as defined in \cref{def:modular:positions_group},
    \item for $\bm{x}$ and $\bm{y} \in \mathcal{P}$, the unordered pair $\{\bm{x},\bm{y}\}$ is in $E_\infty$ if and only if $\bm{x}-\bm{y}$ is in $\bar{\mathcal{C}}^{\mathrm{tr}}$, as defined in \cref{def:modular:cube_connections}.
  \end{itemize}
\end{definition}

With this definition, the fact that $G_\infty$ is undirected is justified by the fact that $\bar{\mathcal{C}}^{\mathrm{tr}}$ is stable by negation.
The positions in a configuration can be described as a subgraph of $G_\infty$ of size $n$, but to perform numerical optimization of configurations, $G_\infty$ needs to be reduced to a finite graph.
Increasing finite subgraphs of $G_\infty$ are defined for this reason in the following.
\begin{definition}\label{def:optimal:finite_graphs}
  The increasing sequence $G_i = (V_i, E_i),\ i \in \mathbb{N}$ is defined by induction to be
  \begin{align*}
    V_0 &= \{ \bm{0}_3 \} \\
    \forall i > 0,\ V_i &= V_{i-1} \cup \{ \bm{y} \in V\ |\ \exists \bm{x} \in V_{i-1}, (\bm{x}, \bm{y}) \in E_\infty\}\\
    \forall i \geq 0,\ E_i &= \left\{\{\bm{x},\bm{y}\} \in E_\infty\ |\ \bm{x} \in V_i, \bm{y} \in V_i \right\}.
  \end{align*}
\end{definition}

\begin{fact}\label{fact:optimal:subgraph_size}
  For a valid configuration $\kappa = \{(\bm{p}_i, \bm{\eta}_i, \epsilon_i),\ i \in \nset{0}{n-1}\}$, there is a subgraph of $G_{\lfloor n/2 \rfloor}$ with vertices $(\bm{x}_0,\dots,\bm{x}_{n-1})$ and a vector $\bm{z} \in \mathbb{R}^3$ such that
  \begin{equation*}
    \forall i, \in \nset{0}{n-1}\ \bm{p}_i = \bm{x}_i+\bm{z}.
  \end{equation*}
\end{fact}
\begin{proof}
  Knowing that all the modules in a valid configuration must be connected, there exists one module from which every other module of the configuration can be reached with at most $\lfloor n/2 \rfloor$ connections.
  Denoting this module's index $j \in \nset{0}{n-1}$, then, by \cref{def:optimal:finite_graphs}, $\{p_i-p_j\ |\ i \in \nset{0}{n-1}\}$ is a subset of $G_{\lfloor n/2 \rfloor}$, such that the resulting subgraph and $\bm{z} = \bm{p}_j$ are as stated.
\end{proof}

\Cref{fact:optimal:subgraph_size} ensures that when searching for an optimal configuration of $n$ modules, using the finite graph $G_{\lfloor n/2 \rfloor}$ instead of $G_\infty$ is not restrictive.
In the rest of this section, $G = (V,E)$ is assumed to be a finite subgraph of $G_\infty$ of $N$ vertices appropriate for the problems formulated.

\subsection{Mixed-integer representation of configurations}

\begin{definition}\label{def:optimal:configuration_representation}
  Let $\bm{z} \in \mathbb{R}^3$ be a variable that represents the offset vector as in \cref{def:modular:configuration}.
  For $\bm{x} \in V$, $\bm{\eta} \in \mathcal{H}$, $\epsilon \in \{0,1\}$, let
  \begin{itemize}
    \item $\bp \in \{0,1\}$ indicate whether a module is placed at position $\bm{z} + \bm{x}$,
    \item $\bpo \in \{0,1\}$ indicate whether a module is placed at position $\bm{z} + \bm{x}$ with orientation $\bm{\eta}$,
    \item $\bpd \in \{0,1\}$ indicate whether a module is placed at position $\bm{z} + \bm{x}$ with rotor spinning direction $\epsilon$.
  \end{itemize}
\end{definition}

The variables $\bm{z}$, $\bp$, $\bpo$, and $\bpd$ completely describe modular configurations.
Based on these variables' definition, the information of allocated positions is redundant, which is a deliberate choice to simplify subsequent definitions.
To ensure that exactly one orientation and one spinning direction is chosen per allocated position, and zero per unallocated position, the introduced binary variables must verify
\begin{align}
  \forall \bm{x} \in V,\ &\sum_{\bm{\eta} \in \mathcal{H}} \bpo = \bp \label{eq:optimal:compatible_orientations} \\
  \forall \bm{x} \in V,\ &\sum_{\epsilon \in \{0,1\}} \bpd = \bp. \label{eq:optimal:compatible_directions}
\end{align}
Moreover, for $\bm{z}$ to satisfy \cref{def:modular:configuration}, the following equality is required
\begin{equation}\label{eq:optimal:offset}
  \bm{z} \sum_{\bm{x} \in V} \bp + \sum_{\bm{x} \in V} \bm{x} \bp = \bm{0}.
\end{equation}

\subsection{Formulation of the connectivity constraint}

A configuration described by a subgraph of $G$ is valid only if that subgraph is connected.
A MIP formulation can be used to define this constraint.
It is based on the work presented in~\cite{shirabe2005model} for allocation of contiguous spatial units.
That work introduces a flow formulation where an ``artificial sink'' is chosen among the possible units and all the allocated units must be able to route their flow to the ``artificial sink'' via other connected allocated units.
Assuming that the number of modules $n \in \mathbb{N}$ for a desired configuration is fixed, the MIP contiguity constraint can be formulated by first introducing the variables:
\begin{itemize}
  \item $\gamma_{\bm{x}}$, a binary variable to indicate whether $\bm{x} \in V$ is the sink,
  \item $\delta_{\bm{x}\bm{y}}$, a nonnegative real variable that represents the flow from $\bm{x}$ to $\bm{y}$ for $\{\bm{x},\bm{y}\} \in E$.
\end{itemize}
Then, set of allocated positions are ensured contiguous with:
\begin{align}
    \forall \bm{x} \in V,\ &\sum_{\{\bm{y}\ |\ (\bm{x},\bm{y})\in E\}} \delta_{\bm{x}\bm{y}} - \delta_{\bm{y}\bm{x}} \leq x_{\bm{x}} - n\gamma_{\bm{x}}
    \label{eq:optimal:contiguous:1} \\
    \forall \bm{x} \in V,\ &\sum_{\{\bm{y}\ |\ (\bm{x},\bm{y})\in E\}} \delta_{\bm{x}\bm{y}} \leq (n-1)x_{\bm{x}}
    \label{eq:optimal:contiguous:2}\\
    &\sum_{\bm{x} \in V} \gamma_{\bm{x}} = 1
    \label{eq:optimal:contiguous:3}
\end{align}

\subsection{Wake interactions}

The search for optimal configurations should avoid returning solutions whose modules are directly in the wake of another module, as wake interactions would reduce the efficiency of these modules.
This requirement can be enforced by adding the constraints
\begin{equation}\label{eq:optimal:wake}
  \bp \leq 1 - \bpo<\bm{y}>
\end{equation}
for every $\bm{x} \in V$, $\bm{y} \in V$, $\bm{\eta} \in \mathcal{H}$ such that a module at position $\bm{x}$ is in the wake of a module at position $\bm{y}$ with orientation $\bm{\eta}$.

\subsection{Control allocation matrix}

To simultaneously optimize configurations and their control allocation matrix, the variable $\bm{C} \in \mathbb{R}^{N\times 6}$ is defined.
Unlike in \cref{sec:control}, $\bm{C}$ has as many rows as vertices in $V$, with each one corresponding to a possible module position.
$V$ is therefore given an arbitrary indexing, so that its vertices are written $\{\bm{x}_i\ |\ i \in \nset{0}{N-1}\}$.

The matrix $\Mtt$ is also modified accordingly, and is now of size $6 \times N$.
Adapting \cref{fact:modular:thrust_torque_matrices,def:control:thrust_torque_matrix} leads to
{%
  \everymath{\displaystyle}
  \begin{align}
    &\forall i \in \nset{0}{N-1},\ 
    \Mtt_{:,i}
    = \nonumber \\
    &\begin{bmatrix}
      k^{\mathrm{th}}
      \sum_{\bm{\eta} \in \mathcal{H}}
      \bpo<\bm{x}_i> \bm{\eta} \\
      k^{\mathrm{to}} \sum_{\bm{\eta} \in \mathcal{H}, \epsilon \in \{0,1\}}
      \bpo<\bm{x}_i> \bpd<\bm{x}_i> \bm{\eta} \epsilon
      +
      k^{\mathrm{th}}
      \sum_{\bm{\eta} \in \mathcal{H}}
      \bpo<\bm{x}_i> (\bm{z} + \bm{x}_i) \times \bm{\eta} \label{eq:optimal:var_thrust_torque}
    \end{bmatrix}.
  \end{align}
}

For $\bm{C}$ to be a proper control allocation matrix, the equalities of \cref{fact:control:allocation_matrix} must be satisfied.
$\bm{C}\bm{H} = \bm{C}$ remains a linear equality in the variables as long as $\bm{H}$ is fixed.
Fixing $\bm{H}$ makes perfect sense because it determines the degrees of freedom of the vehicle to optimize.
Meanwhile, $\Mtt \bm{C} = \bm{H}$ is not linear anymore since $\Mtt$ depends on the parameterized configuration, but it can be linearized via the ``big-M'' method as shown in \cref{fact:optimal:mc_lin}.

\begin{fact}\label{fact:optimal:mc_lin}
  $\Mtt \bm{C} = \bm{H}$ is equivalent to an equality equation written as a sum of products, where each product involves at most one non-binary variable, allowing it to be linearized with the ``big-M'' method.
\end{fact}
\begin{proof}
  From \cref{eq:optimal:var_thrust_torque}, it can be seen that the only non-binary variables in the equation $\Mtt \bm{C} = \bm{H}$ are $\bm{z}$ and $\bm{C}$ itself.
  By multiplying both sides by $\sum_{\bm{x} \in V} \bp$ (which is a positive term), \cref{eq:optimal:offset} can be used to substitute $\bm{z} \sum_{\bm{x} \in V} \bp$ with binary variables only, leaving the coefficients of $\bm{C}$ as the only non-binary variables in the resulting equation.
\end{proof}

Now that the constraints on $\bm{C}$ have been shown to be representable as linear, the same constraints as the ones in \cref{prog:control:alloc_max_auth_2} may be integrated to a configuration optimization problem.
For example, a program can be formulated to maximize the yaw control authority of a 6DOF hexa-rotorcraft.
Control authority can also be used as a constraint, to enforce a minimum authority along particular dimensions, while optimizing a different type of objective, for example a structural one.

\subsection{Structural optimization}

The structural optimization of modular configurations follows the stiffness formulation detailed in \cref{sec:structures}.
The case considered is the one of ensuring robust constraints of the type
\begin{equation}\label{eq:optimal:pb_def}
  \norm{\bm{P}}_2 \leq 1 \implies \bm{c}^\top \bm{v} \leq \lambda,
\end{equation}
where $\bm{c} \in \mathbb{R}^{6\njt}$ is either fixed or a variable, $\bm{v}$ and $\bm{P}$ are valid displacements and loads related by the stiffness equation, and $\lambda \in \mathbb{R}$ is a variable.
This type of constraint can be integrated to MIP programs via a reformulation of the worst-case scenario as second-order conic constraints.

To do so, an extended list of joints from all the positions in $V$ is defined with
\begin{equation}
  \mathcal{J}_V = [\bm{j}_0,\dots,\bm{j}_{\njt-1}],
\end{equation}
where $\njt$ denotes the total number of joints and $\forall i \in \nset{0}{\njt-1},\ \bm{j}_i \in \{\bm{x}_j + \bm{g}\ |\ j \in \nset{0}{N-1}, \bm{g} \in V^{\mathcal{D}}\}$.
Note that the offset vector $\bm{z}$ is not required, since a common translation of a space frame's joints does not change its stiffness.
 The elastic stiffness matrix of the structure $\bm{K} \in \mathbb{R}^{6\njt \times 6\njt}$ can be represented as
\begin{equation}\label{def:optimal:stiffness_matrix}
  \bm{K} = \sum_{\bm{x} \in V} \bp \bm{k}_{\bm{x}},
\end{equation}
where $\bm{k}_{\bm{x}}$ is the contribution to the global stiffness of the members belonging to the module placed at $\bm{x}$.

In general, the rank of $\bm{K}$ may not be equal to $6\njt - 6$ anymore, since the columns and rows of $\bm{K}$ corresponding to joints of unallocated modules are zero.
However, valid displacements and loads can still be defined as the ones that are orthogonal to the null space of $\bm{K}$, similarly to the constraints of \cref{eq:structures:displacement_constraint,eq:structures:load_constraint}, implying that these displacements and loads verify
\begin{align*}
  \bm{K}\bm{v} &= \bm{P} \\
  \bm{K}^+\bm{P} &= \bm{v}.
\end{align*}

To simplify \cref{eq:optimal:pb_def}, the following problem can be solved:
\begin{equation*}
  \maximize{\bm{c}^\top \bm{v}}
  {\norm{\bm{P}}_2 \leq 1}
  {\bm{K}\bm{v} = \bm{P}}
  {\bm{K}^+\bm{P} = \bm{v}}.
\end{equation*}
This problem is a convex quadratic one, whose optimal value is equal to $\norm{\bm{K}^+\bm{c}}_2$ and reached for $\bm{v} = \frac{{(\bm{K}^+)}^2\bm{c}}{\norm{\bm{K}^+\bm{c}}_2}$.
\Cref{eq:optimal:pb_def} is then replaced by
\begin{align}
  \bm{K} \bm{w} &= \bm{c} \label{eq:optimal:struct_w:1} \\
  \norm{\bm{w}}_2 &\leq \lambda, \label{eq:optimal:struct_w:2}
\end{align}
where $\bm{w}$ is a newly introduced displacement variable, which, when verifying \cref{eq:optimal:struct_w:1,eq:optimal:struct_w:2}, always guarantees that $\norm{\bm{K}^+\bm{c}}_2 \leq \lambda$.
When added to a configuration optimization problem, this formulation results in a problem that is at least a MISOCP (MIP second-order cone) problem, owing to \cref{eq:optimal:struct_w:2}.
A possible application of this type of inequality is to ensure that the member axial loads are minimized when subject to unit-norm load vectors.

\subsection{Example}

As an example, a program is formulated to maximize the control authority in terms of three-dimensional torques for an arbitrary a 6DOF configuration of \num{12} modules.
The projection of actuation vectors on the space of torques only is given by the operator
\begin{equation*}
  \bm{S}
  =
  \begin{bmatrix}
    \bm{0}_{3 \times 3} & \bm{0}_{3 \times 3} \\
    \bm{0}_{3 \times 3} & \bm{I}_3
  \end{bmatrix}.
\end{equation*}
Therefore, the optimal configuration is found by solving the following program, which results in the previously studied configuration shown in \cref{fig:control:optimized:1}.
This optimization problem was formulated using the Matlab package YALMIP~\cite{yalmip} and solved with Gurobi~\cite{gurobi}.


\begin{program}\label{program:optimal:example:1}
  With variables $\lambda$, $\bp$, $\bpo$, $\bpd$, $\bm{z}$, $\delta_{\bm{x}}$, $\gamma_{\bm{x}\bm{y}}$, and intermediate variables created from the ``big-M'' method:
  \begin{equation*}
    \minimize{\lambda}
    {}
    {%
      \kern-2em%
      \Mtt \bm{C} = \bm{H}
    }
    {%
      \kern-2em%
      \bm{C} \bm{H} = \bm{C}
    }
    {%
      \kern-2em%
      \norm{\bm{S} \bm{C}^\top \bm{A}_{i,:}^\top}_2 \leq s \bm{b}_i - s \bm{A}_{i,:} \uh,\ \forall i \in \nset{0}{2n-1}
    }
    {%
      \kern-2em%
      \bp,\bpo,\bpd,\bm{z},\gamma_{\bm{x}},\delta_{\bm{x}\bm{y}}\text{\ verify \cref{eq:optimal:compatible_orientations,eq:optimal:compatible_directions,eq:optimal:offset,eq:optimal:contiguous:1,eq:optimal:contiguous:2,eq:optimal:contiguous:3,eq:optimal:wake}}.
    }
  \end{equation*}
\end{program}

%

\begin{figure}
  \centering
  \includegraphics[width=\linewidth]{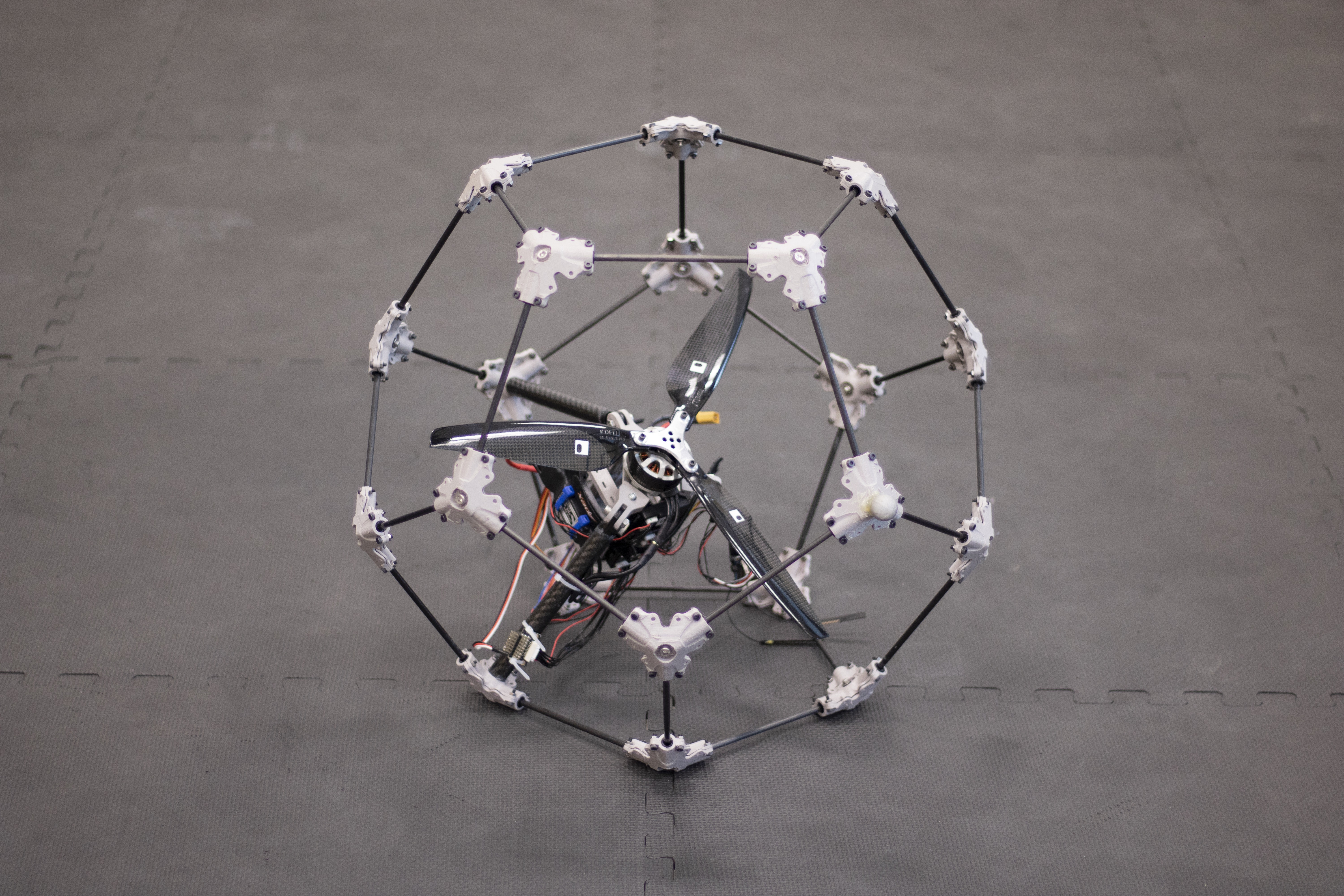}
  \caption{Module prototype.\label{fig:prototype:module}}
\end{figure}

\section{The module prototype}\label{sec:prototype}

A prototype of the Dodecacopter module was built to assess the practical feasibility of the concept presented in this work (\cref{fig:prototype:module,fig:prototype:corner,fig:prototype:confs}).
The components of the prototype are discussed in this section and a few configurations that were flight-tested are presented.
A comparison of the required average motor inputs to hover the flown configurations is performed with the goal of assessing the impact of rotor-rotor interactions on performance for three-dimensional configurations.

\subsection{Prototype description}

\paragraph{Frame and connectors}

\begin{figure}
  \centering
  \includegraphics[width=\linewidth]{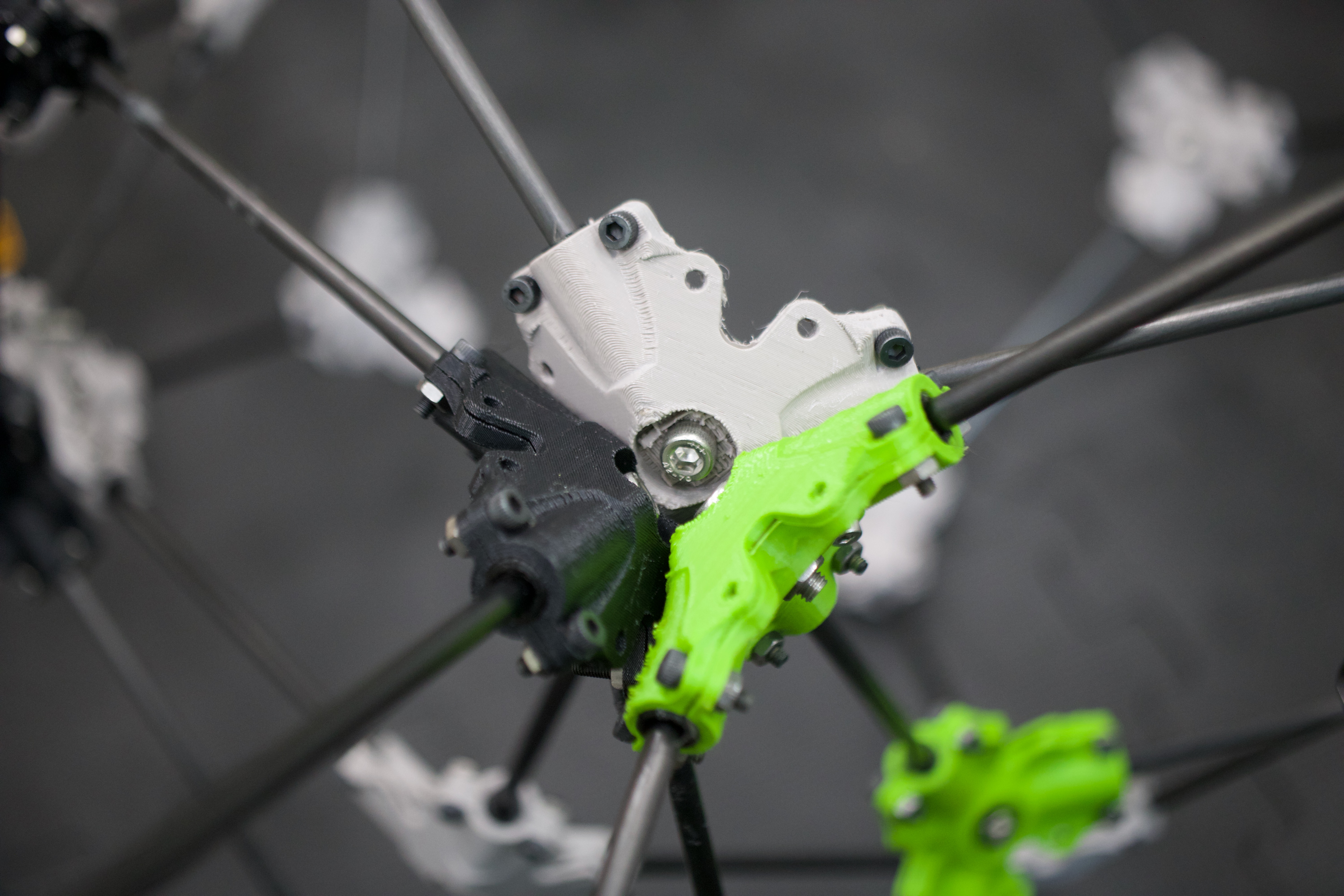}
  \caption{Three corners belonging to different modules forming a connection.\label{fig:prototype:corner}}
\end{figure}

The frame of the prototype is made of carbon fiber reinforced polymer (CRFP) tubes connected together with multiple 3D printed plastic parts.
The first plastic part is a short tube with a closed end made of soft thermal polyurethane (TPU).
These tubes enclose each end of the CRFP tubes and remain in place with a tight fit.
A pair of 3D printed matching parts are used for each corner of the dodecahedron frame.
Each pair is joined together with metal fasteners and clamps three TPU end tubes.
Three assembled corners forming a connection between three modules are shown in \cref{fig:prototype:corner}.

\paragraph{Propulsion}\ 

  For propulsion, each module prototype uses a brushless motor with a rating of \SI{465}{\kilo\volt} controlled by an electronic speed controller, and a three-blade propeller with a diameter of about \SI{39.3}{\centi\meter} that is oriented either CW or CCW.

\paragraph{Flight controller}

Each module can hold a Pixhawk 4 flight controller that sends a PWM signal to the electronic speed controller and may be used as well to control other modules of the assembly via standard servo cable connections.
Only one flight controller per configuration was used to control the whole vehicle for most flights.

\paragraph{Communications}

The prototype includes a \SI{900}{\mega\Hz} radio receiver to receive radio commands from the operator and a \SI{2.4}{\giga\Hz} ESP8266 WiFi module to communicate with the ground station.

\paragraph{Power source}

Two slots are reserved per module to place six-cell lithium-polymer batteries with a capacity of \SI{1350}{\milli\ampere\hour}.

\subsection{Tested configurations}

The configurations that were flown are summarized in \cref{table:prototype:configurations}.
The ArduPilot flight controller software~\cite{ardupilot} was used to control the vehicles.
Each configuration required tuning the gains of the control loops manually, which was particularly time-consuming.
In the future, methods to automatically tune the different control loops based on the assembled configuration would be desirable.
The standard configurations, that is the quadrotor and hexarotor, were the easiest to tune.

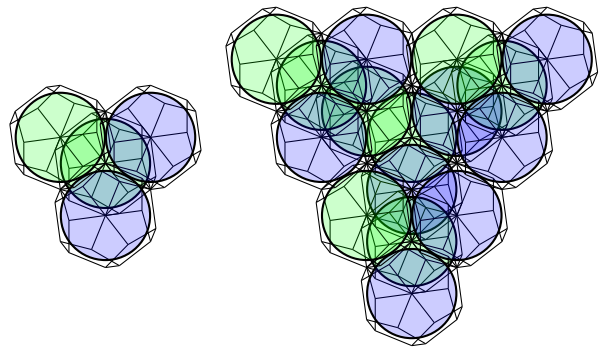
\begin{figure}
  \centering
  \begin{subfigure}[c]{0.32\linewidth}
    \centering
    \tdplotsetmaincoords{0}{0}
    \begin{tikzpicture}[tdplot_main_coords, scale=3]
      \tetraconf
    \end{tikzpicture}
  \end{subfigure}
  \hfill
  \begin{subfigure}[c]{0.66\linewidth}
    \tdplotsetmaincoords{0}{0}
    \begin{tikzpicture}[tdplot_main_coords, scale=3]
      \tetradconf
    \end{tikzpicture}
  \end{subfigure}
  \caption{Top view representation of the tetrahedron quadrotor and hexadecarotor configurations.\label{fig:prototype:tetrahedron_top_views}}
\end{figure}

The tetrahedron configurations were flown successfully as well, but suffered from a slightly reduced power efficiency from wake interactions, since some overlap between rotors at different heights occurs, as shown in \cref{fig:prototype:tetrahedron_top_views}.
The 4-module tetrahedron configuration was also characterized by a relatively low yaw control authority, due to the particular rotor arrangement and the resulting unique control allocation matrix, which implies that the top motor is issued control input changes three times as high as the other motors to control yaw, as seen from the control allocation matrix:
\begin{equation}\label{eq:prototype:tetracopter_allocation_matrix}
  \bm{C}^{\mathrm{tetraquad}}
  \propto
  \begin{bmatrix}
    1 & -1 & 3 & 3 \\
    1 & -1 & -1 & -1 \\
    1 & 1 & -3 & -1 \\
    1 & 1 & 1 & -1
  \end{bmatrix}.
\end{equation}
The effect of the magnitude differences between coefficients assigned to different modules can be seen in \cref{fig:prototype:hover_plots:tetra}.
This configuration had therefore a tendency to start spinning around the yaw axis and show difficulty to recover.
The larger tetrahedron-shaped configurations did not show the same issue, and their overactuated nature ensured that the control allocation matrix could be chosen to yield a good control authority around all moment axes.

The 6DOF hexarotor also posed a challenge from the difficulty in tuning its control loops and from its reduced thrust-to-weight ratio, leading to the motors running close to maximum thrust at hover and leaving little room for control.
At \SI{1.7}{\kilo\gram} per module, the Dodecacopter prototype vehicles are relatively heavy for their size, which clearly limits possibilities for creating 6DOF vehicles capable of agile flights.
This heavy weight is the direct consequence of the complex frame geometry and required number of attachment points which, when manufactured with rapid prototyping methods, need to be somewhat bulky to be structurally sound.
Nonetheless, the successful flights of very diverse configurations of up to sixteen modules show the versatility of the introduced modular system.

\begin{table}
  \begin{center}
    \begin{tabular}{l c c c c c}
      Name & Modules & DOF & Layers & Motor input & Figure \\
      Quadrotor\makecell[l]{\ \\ \ } & 4 & 4 & 1 & 0.30 & \labelcref{fig:prototype:confs:quad} \\
      Hexarotor\makecell[l]{\ \\ \ } & 6 & 4 & 1 & 0.31 &\labelcref{fig:prototype:confs:hexa} \\
      \makecell[l]{6DOF\\\ Hexarotor} & 6 & 6 & 1 & 0.58 & \labelcref{fig:prototype:confs:hexa6dofs} \\
      \makecell[l]{Tetrahedron\\\ Quadrotor} & 4 & 4 & 2 & 0.34 & \labelcref{fig:prototype:confs:tetra} \\
      \makecell[l]{Tetrahedron\\\ Decarotor} & 10 & 4 & 3 & 0.35 & \labelcref{fig:prototype:confs:decatetra} \\
      \makecell[l]{Tetrahedron\\\ Hexadecarotor} & 16 & 4 & 4 & 0.37 & \labelcref{fig:prototype:confs:tetragen2}
    \end{tabular}
  \end{center}
  \caption{%
    List of flown configurations with their number of modules, degrees of freedom, number of layers of modules, and average control input during hover.%
    \label{table:prototype:configurations}
  }
\end{table}

\subsection{Thrust efficiency of three-dimensional configurations}

\begin{figure}
  \centering
  \begin{subfigure}{\linewidth}
    \centering
    \includegraphics[width=\linewidth]{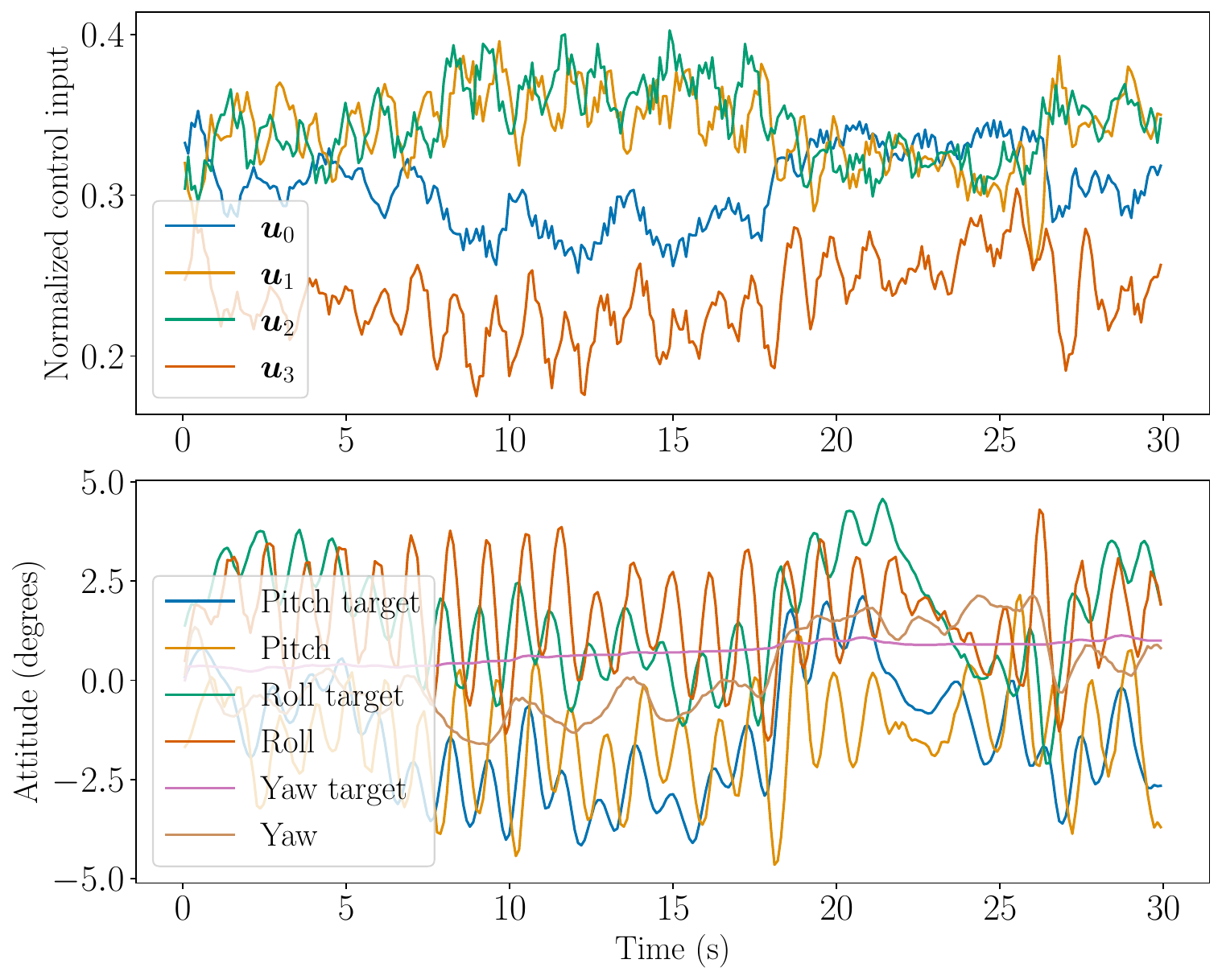}
    \caption{Quadrotor.\label{fig:prototype:hover_plots:quad}}
  \end{subfigure}
  \begin{subfigure}{\linewidth}
    \centering
    \includegraphics[width=\linewidth]{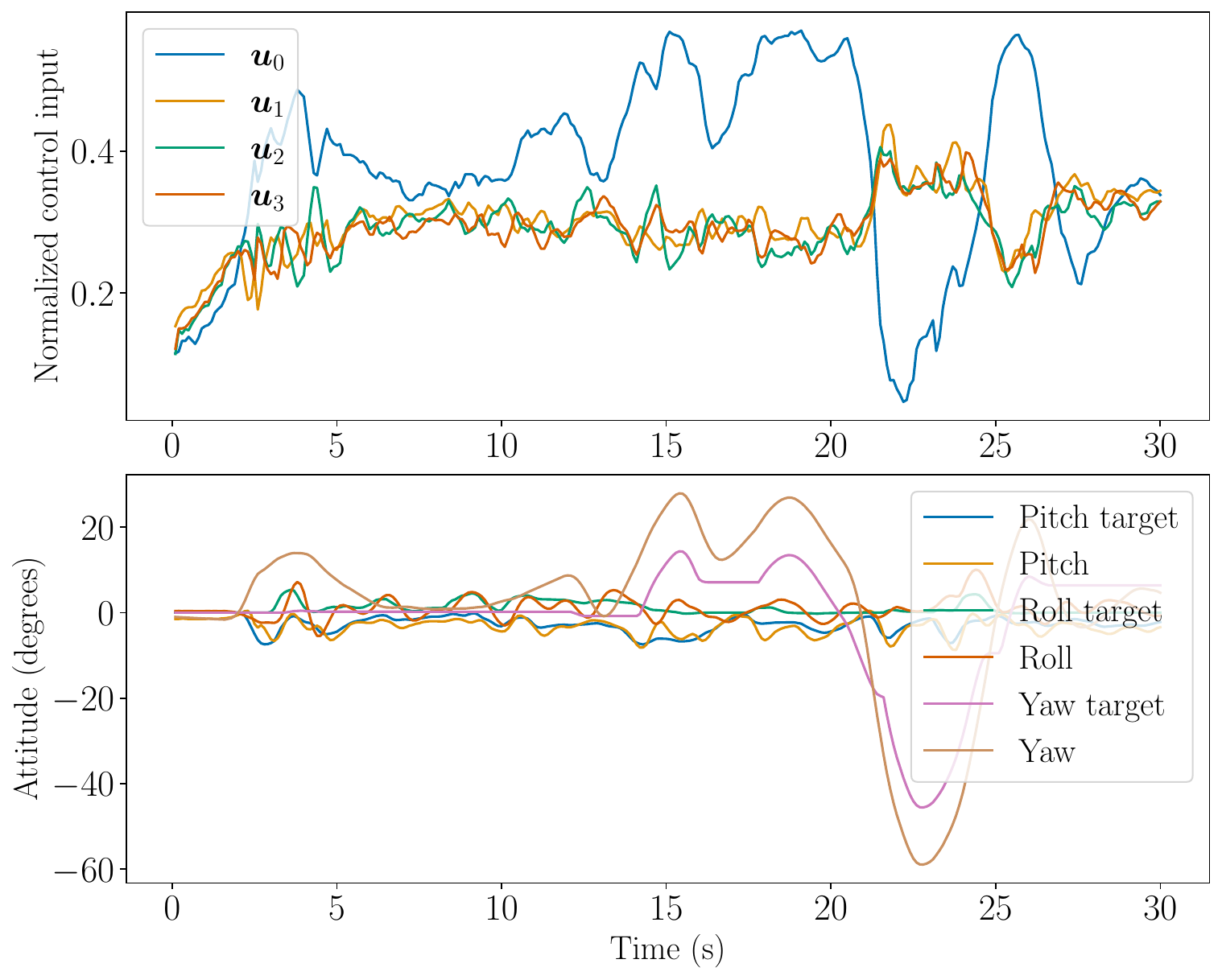}
    \caption{Tetrahedron Quadrotor.\label{fig:prototype:hover_plots:tetra}}
  \end{subfigure}
  \caption{Control inputs and attitude during hover.\label{fig:prototype:hover_plots}}
\end{figure}

To evaluate the impact that stacked layers of rotors have on the total thrust produced by a vehicle, motor inputs required to sustain hover were recorded and compared for the hexarotor, tetrahedron quadrotor, and tetrahedron hexadecarotor.
Vehicles were hovered during at least \SI{30}{\second} each with a motion capture system to provide position feedback.
The average of the normalized motor inputs over all motors and hovering time are shown in \cref{table:prototype:configurations} and samples of flight data are shown in \cref{fig:prototype:hover_plots}.
Vehicles with rotor overlap required a higher average motor input to hover of about \SI{14}{\percent}.
However there was no significant difference between the three tetrahedron configurations despite their different numbers of layers.
Although no definitive conclusion can be made based solely on this observation, it seems to indicate that the loss in efficiency is mostly due to rotor wake interactions between modules at adjacent layers and that interactions between layers further from each other are small.
This suggests that larger three-dimensional configurations with multiple layers and limited propulsive efficiency losses are possible, as long as the positioning of the modules avoids direct rotor overlap.

\section{Conclusion}\label{sec:conclusion}

A new class of modular UAS based on a module shaped as a regular dodecahedron was presented in this paper.
Specifically, this UAS belongs to the category of systems composed of multiple rotorcraft modules which are meant to be assembled into vehicles of varying shapes, sizes, and payload capacity.
The uniqueness of the introduced system resides in the chosen module shape and the associated connection method between modules. 
It was shown that these innovations result in a large variety of achievable geometries of aerial vehicles.
Not only the modules can be assembled to recreate conventional multirotor configurations and flat arrays of rotors, they can also assemble in three dimensions to create stiff structures while limiting wake interactions, or be oriented at different angles to grant six degrees of freedom to an assembled vehicle.
The potential for a large number of rotors in a vehicle implies a large space of possible control allocation matrices.
New methods for the computation of optimal allocation matrices based on control authority or power consumption were introduced and numerically evaluated on a configuration example.
It is worth noting that these methods can be applied to non-modular vehicles with many rotors as well.
To support the claim that three-dimensional module arrangements enable stiffer and stronger configurations, the structure of modular vehicles was represented as a space frame and analyzed using the elastic stiffness method.
It was shown that the intuition behind the better structural properties of three-dimensional vehicles is justified by considering indicators of structural performance on different configurations.
These indicators are based on a robust formulation of the frame displacements and its internal loads when subject to arbitrary bounded loads.
In addition, it was shown that the configuration optimization of modular vehicles can be performed with mixed-integer programming, owing to the discreteness of the set of configuration parameters.
Relevant measures of vehicle performance based on their control allocation and structural properties were reformulated to yield programs with linear or second-order conic constraints, for which efficient solvers exist.
Finally, a module prototype that was designed, built, and flown in six different configurations of up to sixteen modules was described.
This prototype proved the feasibility of the designed modular concept and provided insights on the challenges that some configurations might pose.

Several directions remain open to continue the work on this system, other modular systems, and highly overactuated multirotor vehicles in general.
Regarding the control of modular vehicles, a fully decentralized control system would be desirable to ensure a high level of robustness of the system.
The presented prototype was controlled in a centralized fashion, since the focus was placed on the control allocation problem, but this would become impractical for a larger number of modules than what has been tested so far.
Decentralized control would also be a necessity for the flexible rearrangement of modular vehicles, for example through autonomous de- and self-assembly.
Some of the authors have already started looking at the issue of controlling a different type of modular aerial vehicle in which modules do not communicate and are unaware of their position in the vehicle~\cite{shahab2022controlling}.
The presented control allocation strategies themselves would benefit from a comparison under experimental conditions.
It is expected however that a large number of rotors would be required to see a significant difference in, for example, power consumption.
Then, to enable the control of arbitrary configurations without flight controller tuning, it would be necessary to develop methods to infer optimal control parameters for new configurations.
A possible avenue for doing so would be to perform a system's identification with a few small vehicle configurations, to then use the identified models to derive the models of arbitrary configurations, and finally to deduce optimal control parameters from model-based tuning methods.
However, it remains unclear how some effects might contribute to the dynamics of some specific, large configurations.
For examples, the wake interactions that occur with modules at different heights or different orientations are hard to model and quantify.
They therefore require extensive study to understand how they might affect large vehicle configurations, an other area that the authors have been exploring~\cite{epps2022wakeinteractions}.

\appendices%

\section{Convexity of power consumption-based optimization}\label{appendix:conf_hyper_conv}

\newcommand\hg[3]{\ensuremath{\prescript{}{1}F_1\left(#1,#2,#3\right)}}

Let $f : (0,\infty) \mapsto \mathbb{R}$ be the function defined by
\begin{equation}\label{eq:app:convexity:1}
  \forall x \in (0,\infty)\ f(x) = x^{-\frac{3}{4}} \hg{-\frac{3}{4}}{\frac{1}{2}}{-x},
\end{equation}
where $\prescript{}{1}F_1$ is the hypergeometric function of the first kind.
From~\cite[Eq.~13.2.2]{DLMF}, $\hg{a}{b}{z}$ is given by
\begin{equation}\label{eq:app:convexity:2}
  \hg{a}{b}{z}
  = \sum_{s=0}^{\infty} \frac{(a)_s}{(b)_s s!}z^s
  = 1 + \frac{a}{b}z + \frac{a(a+1)}{b(b+1)2!}z^2 + \dots~.
\end{equation}
from which it can be shown that
\begin{equation}\label{eq:app:convexity:3}
  \frac{\d}{\d z}\hg{a}{b}{z} = \frac{a}{b}\hg{a+1}{b+1}{z}.
\end{equation}
In addition, Kummer's transformation~\cite[Eq.~13.2.39]{DLMF} gives
\begin{equation}\label{eq:app:convexity:4}
  \hg{a}{b}{z}
  =
  \e^z \hg{b-a}{b}{-z}.
\end{equation}
By substituting in \cref{eq:app:convexity:4} the values $a = -\frac{3}{4}$ and $b = \frac{1}{2}$,
\begin{equation}\label{eq:app:convexity:5}
  f(x)
  =
  x^{-\frac{3}{4}}
  \e^{x}
  \hg{\frac{5}{4}}{\frac{1}{2}}{x}.
\end{equation}
Consequently, using \cref{eq:app:convexity:3},
\begin{align*}
  \frac{\d^2}{\d x^2} f(x)
  =
  \bigg[
    &
    \frac{21}{16} x^{-\frac{11}{4}}\e^{-x}\hg{\frac{5}{4}}{\frac{1}{2}}{x}
    \\
    &
    +
    x^{-\frac{3}{4}} \hg{\frac{5}{4}}{\frac{1}{2}}{x}
    \\
    &
    +
    \frac{15}{4} x^{-\frac{3}{4}} \hg{\frac{13}{4}}{\frac{5}{2}}{x}
    \\
    &
    +
    \frac{3}{2} x^{-\frac{7}{4}} \hg{\frac{5}{4}}{\frac{1}{2}}{x}
    \\
    &
    -
    \frac{15}{4} x^{-\frac{7}{4}} \hg{\frac{9}{4}}{\frac{3}{2}}{x}
    \\
    &
    -
    \frac{5} x^{-\frac{3}{4}} \hg{\frac{9}{4}}{\frac{3}{2}}{x}
  \bigg]
  \e^{-x},
\end{align*}
which, after substituting the values of $\prescript{}{1}F_1$ with their power series expansion and grouping the terms of same power together, leads to
\begin{align*}
  &
  \frac{\d^2}{\d x^2} f(x)
  =
  \e^{-x}
  \frac{21}{16} x^{\frac{-11}{4}}
  +
  \e^{-x}
  \frac{15}{4}
  \left(
    1 - \frac{3}{2} + \frac{21}{8}
  \right)
  x^{-\frac{7}{4}}
  \\
  &
  +
  \e^{-x}
  \frac{3}{256} 
  x^{-\frac{3}{4}}
  \sum_{s=0}^{\infty}
  \frac{(\frac{5}{4})_s}{(\frac{1}{2})_s s!}
  x^s
  \dots 
  \\
  &
  \frac{4096s^5 + 46080s^4 + 208384s^3 + 484032s^2 + 592282s + 310875}{8s^5 + 84s^4 + 338s^3 + 651s^2 + 599s + 210}  \\
  &
  \geq 0
\end{align*}

hence $f$ is convex on $(0,\infty)$.





\section*{Acknowledgement}\label{sec:acknowledegment}

The authors would like to thank optimAero LLC for allowing the team to use their facilities during the prototyping of the Dodecacopter as well as the Indoor Flight Laboratory at Georgia Tech.

\printbibliography

\end{document}